\documentclass[[manuscript,12pt]{colt2022} % Include author names

\usepackage[utf8]{inputenc} % allow utf-8 input
\usepackage[T1]{fontenc}    % use 8-bit T1 fonts
\usepackage{hyperref}       % hyperlinks
\usepackage{url}            % simple URL typesetting
\usepackage{booktabs}       % professional-quality tables
\usepackage{amsfonts}       % blackboard math symbols
\usepackage{nicefrac}       % compact symbols for 1/2, etc.
\usepackage{microtype}      % microtypography

\usepackage[toc,page,header]{appendix}
\usepackage{minitoc}
\renewcommand \thepart{}
\renewcommand \partname{}

\usepackage{amsmath}
\usepackage{empheq}

\usepackage{mathtools}
\usepackage{bm}
\usepackage{algorithm}
\usepackage{algorithmic}
\usepackage{wrapfig,soul}
\usepackage{enumitem}
\usepackage{xcolor}
\usepackage{grffile}
\usepackage{thm-restate}
\usepackage{natbib}
\usepackage{amsmath}

\usepackage[most]{tcolorbox}
\usepackage{pgf}
\graphicspath{{./figures/}}
\newcommand{\emphblockoption}{drop shadow,
    colframe=black!60,
    colback=black!10,
    coltitle=white!, 
    left=.2pt,
    right=.2pt,
    boxrule=1pt,
    arc=1pt}
\tcbset{emphblock/.style={code={\pgfkeysalsofrom{\emphblockoption}}}}

\newtheorem{Lemma}{Lemma}
\newtheorem{Proposition}{Proposition}
\newtheorem{Theorem}{Theorem}
\newtheorem{Assumption}{Assumption}
\newtheorem{Remark}{Remark}

\newtheorem{Corollary}{Corollary}

\newcommand{\E}{\mathbb{E}}
\newcommand\numberthis{\addtocounter{equation}{1}\tag{\theequation}}

\makeatletter
\newcommand\footnoteref[1]{\protected@xdef\@thefnmark{\ref{#1}}\@footnotemark}
\makeatother

\title[Towards Understanding A3C: Convergence and Linear Speedup]{Towards Understanding Asynchronous Advantage Actor-critic: Convergence and Linear Speedup}
\usepackage{times}
\coltauthor{%
 \Name{Han Shen} \Email{shenh5@rpi.edu}\\
 \addr Rensselaer Polytechnic Institute
 \AND
 \Name{Kaiqing Zhang} \Email{kaiqing@mit.edu}\\
 \addr Massachusetts Institute of Technology
 \AND
 \Name{Mingyi Hong} \Email{mhong@umn.edu}\\
 \addr University of Minnesota, Minneapolis
 \AND
 \Name{Tianyi Chen} \Email{chent18@rpi.edu}\\
 \addr Rensselaer Polytechnic Institute
}

\begin{document}

\maketitle

\begin{abstract}
	Asynchronous and parallel implementation of standard reinforcement learning (RL) algorithms is a key enabler of the tremendous success of modern RL. 
        Among many asynchronous RL algorithms, arguably the most popular and effective one is the asynchronous advantage actor-critic (A3C) algorithm.  
        Although A3C is becoming the workhorse of RL, its theoretical properties are still not well-understood, including its non-asymptotic analysis and the performance gain of parallelism (a.k.a. linear speedup). 
        This paper revisits the A3C algorithm and establishes its non-asymptotic convergence guarantees.
        Under both i.i.d. and Markovian sampling, we establish the local convergence guarantee for A3C in the general policy approximation case and the global convergence guarantee in softmax policy parameterization. Under i.i.d. sampling, A3C obtains sample complexity of $\mathcal{O}(\epsilon^{-2.5}/N)$ per worker to achieve $\epsilon$ accuracy, where $N$ is the number of workers.  Compared to the best-known sample complexity of $\mathcal{O}(\epsilon^{-2.5})$ for two-timescale AC, A3C achieves \emph{linear speedup}, which justifies the advantage of parallelism and asynchrony in AC algorithms theoretically for the first time.
        Numerical tests on synthetic environment, OpenAI Gym environments and Atari games have been provided to verify our theoretical analysis. 
\end{abstract}

\section{Introduction}

% Reinforcement learning (RL) seeks to maximize the expected reward in a stochastic process which is normally modeled as a Markov decision process (MDP). 
Reinforcement learning (RL) has achieved impressive performance in many domains such as robotics \citep{lillicrap2015,mnih2015nature} and video games \citep{A3C}. However, these empirical successes are often at the expense of 
significant computation. To unlock high computation capabilities, the state-of-the-art RL approaches rely on sampling data from massive parallel simulators on multiple machines \citep{nair2015massively,A3C,assran2019gossip,chen2018comm}. 
Empirically, these approaches can significantly \emph{reduce training time} when implemented in an {\it asynchronous} manner. 
One popular method that achieves the state-of-art performance is the asynchronous variant of the actor-critic (AC) algorithm, referred to as A3C \citep{A3C}.  

A3C builds on the original AC algorithm \citep{KondaAC}. At a high level, AC simultaneously performs policy optimization (a.k.a. the actor step) using the policy gradient (PG) method \citep{SuttonPG} and policy evaluation (a.k.a. the critic step) using the temporal difference learning (TD) algorithm \citep{sutton1988td}. 
% combine the actor method and the critic method. 
% The actor methods use policy function parameterization and seeks to update the policy parameters directly through gradient ascent. A typical method is the policy gradient method (PG). 
% Actor methods typically suffer from high variance and large sampling cost, which leads to slow convergence. Critic method, such as the temporal difference learning algorithms (TD) \citep{sutton1988td}, seeks to learn a critic parameter to approximate the value functions by minimizing the Bellman error. 
% The basic idea of actor critic method is: Given a policy, first use critic method to acquire an approximation of the value function. 
% Use the value function approximation to obtain the policy gradient, and then update the policy parameters with the policy gradient. 
To ensure scalability to large state-action spaces, both actor and critic steps can combine with various function approximation techniques.
To ensure stability, AC is often implemented in a two time-scale fashion, where the actor step runs in the slow timescale and the critic step runs in the fast timescale. 
% In a two time-scale actor critic algorithm, the actor and critic parameters are updated simultaneously. But the actor step size is smaller than that of critic, so that actor parameter changes more slowly than critic parameter.
Similar to other on-policy RL algorithms, AC uses samples generated from the target policy. 
Thus, data sampling is entangled with the learning procedure, which generates significant \emph{overhead}.
To speed up the sampling process of AC, A3C introduces multiple workers with a shared policy, and each worker has its own simulator to perform data sampling. The shared policy can be then updated using samples collected from multiple workers. 

Despite the empirical success achieved by A3C, to the best of our knowledge, its theoretical property is not well-understood. The following \emph{theoretical} questions remain unclear: \textbf{Q1)} Under what assumption does A3C converge? If so, does it converge to the global optimal solution? \textbf{Q2)} What is its convergence rate? \textbf{Q3)} Can A3C obtain benefit (or linear speedup) using parallelism and asynchrony? 
 
For \textbf{Q3}), we are interested in the \emph{training time linear speedup} with $N$ workers, which is the ratio between the training time using a single worker and that using $N$ workers. 
Since asynchronous parallelism mitigates the effect of stragglers and keeps workers busy, the training time speedup can be measured roughly by the sample complexity (i.e., computational) linear speedup \citep{lian2016speedup}:
\begin{align}\label{eq.speedup}
    {\rm Speedup}(N) 
    =\frac{\text{sample complexity with one worker}}{\text{average sample complexity per worker with $N$ workers}}.
\end{align}
If ${\rm Speedup}(N)= \Theta(N)$, the speedup is linear, and the training time roughly reduces linearly as the number of workers increases.
This paper aims to answer this question, towards the goal of providing theoretical justification for the empirical successes of parallel and asynchronous RL.  

% {\red[according to yesterday's discussion, the benefit we can show was not because of asynchrony, but because of parallization, right?]} {\blue[partially also because of asynchrony. in sync case, many workers will be idle for a fraction of time (due to straggler), so we cannot use sample complexity/N to estimate the run time.]}

\subsection{Related works}

\textbf{The PG method and its global convergence.} The global optimality of the stationary points of policy optimization problems has been shown in \citep{bhandari2020global}. Then the finite-time convergence rate for exact PG method with softmax policy was established in \citep{agarwal2019optimality} by utilizing a gradient-dominance type result under relative entropy regularized objective function. Later, \citep{mei2020softmax} extended this result to the entropy regularized setting and established linear convergence rate for exact PG method under softmax parameterization. Later, \citep{bhandari2021linear} has proved linear convergence rate for general class PG methods. In the stochastic setting, \citep{zhang2019global} has established local optimal convergence for stochastic PG with unbiased rollout and increasing step sizes, and \citep{wang2019neural} has established the global convergence of stochastic neural PG with increasing batch of i.i.d. samples. Later, \citep{junzi2021sample} proved that the minibatch version of PG achieves global convergence with the help of relative entropy regularization.
But none of them consider the global convergence of the AC method. On the application side, the PG method has been broadly applied in various settings; see e.g. \citep{chai2019joint,chai2020online,ellaham2021policy,cervino2021multitask}. In \citep{chai2019joint}, the actor critic method was used to jointly optimize the trajectory, transmission and caching content delivery of the unmanned aerial vehicles. In \citep{chai2020online}, the policy gradient method was used to jointly optimize the streaming rate and transmission power. In \citep{ellaham2021policy}, the PG method was used to help the learning of a distribution adaptation strategy. In \citep{cervino2021multitask}, the PG method is used in a multitask learning algorithm which seeks to improve generalization to new tasks. 

\textbf{Analysis of AC algorithm.}
AC method was first proposed by \citep{borkar1997actor,KondaAC}, with asymptotic convergence guarantees provided in \citep{borkar1997actor,KondaAC,Bhatanagar2009NAC}. 
% Off-policy actor-critic can also be shown to   provably converge asymptotically  \citep{degris2012off,zhang2019provably}. 
It was not until recently that the \emph{non-asymptotic} analyses of AC have been established. The finite-sample guarantee for the batch AC algorithm has been established in \citep{yang2018finite,kumar2019sample,fu2020single} with i.i.d. sampling. Later, in \citep{qiu2019finite}, the finite-sample analysis was established for the double-loop nested AC algorithm under the Markovian setting. An improved analysis for the Markovian setting with minibatch updates has been presented in \citep{xu2020improving} for the nested AC method. More recently, \citep{xu2020non,wu2020finite} have provided the first finite-time analyses for the two-timescale AC algorithms under Markov sampling, with both $\tilde{O}(\epsilon^{-2.5})$ sample complexity, which is the best-known sample complexity for two-timescale AC. Through the lens of bi-level optimization, \citep{hong2020ac} has provided finite-sample guarantees for two-timescale AC, when a \emph{natural} policy gradient step is used in the actor. 
Recently, \citep{fu2020single} also analyzed the single-timescale AC algorithm under an exact critic oracle. 
On a less relevant line of research, AC-based multi-agent RL has been studied in \citep{zhang2018fully,christianos2020shared,qu2020scalable}.
However, none of the existing works has analyzed the effect of the asynchronous and parallel updates in AC.

% . Its asymptotic convergence was then proven by \citep{Bhatanagar2009NAC}. A finite-sample convergence rate was later established by \citep{QiuAC}, with the assumption of having access to i.i.d.. samples. Concurrent works on actor critic mainly focus on Markovian sampling and have also shown better convergence rate. \citep{wu2020finite,hong2020ac} analyzed two-timescale actor critic under \textit{average reward} setting, and established a sample complexity of $\mathcal{O}(\epsilon ^ {-2.5})$ which is the best known sample complexity for two time-scale actor critic algorithms.
% % A mini-batch nested-loop actor critic was analyzed by \citep{xu2020improving} under \textit{discounted reward} setting. It achieves a sample complexity of $\mathcal{O}(\epsilon ^ {-2} \log \frac{1}{\epsilon})$ which is considered the best known sample complexity for actor critic algorithms.

\textbf{Parallel and distributed RL methods.} In \citep{A3C}, the original A3C method was proposed and became the workhorse in empirical RL. Later, \citep{babaeizadeh2017a3cgpu} has provided a GPU-version of A3C which significantly decreases training time. Recently, the A3C algorithm is further optimized in modern computers by \citep{stooke2019accelerated}, where a large batch variant of A3C with improved efficiency is also proposed. In \citep{espeholt2018impala}, an importance weighted distributed AC algorithm IMPALA has been developed to solve a collection of problems with one single set of parameters.
A gossip-based distributed AC algorithm has been proposed in \citep{assran2019gossip} which achieves performance competitive to A3C. Additionally, distributed RL is closely related to the multi-agent RL, both of which have a broad range of applications \citep{kar2013collaborative,sadeghi2018optimal,wu2021byzantine}. In \citep{kar2013collaborative}, a distributed algorithm based on Q-learning was proposed and was shown to achieve convergence under a sparse communication network. In \citep{sadeghi2018optimal}, an asynchronous caching approach which utilized PG to find an optimal caching policy was developed. A robust decentralized TD learning method was proposed in \citep{wu2021byzantine} to defend against malicious agents in a multi-agent network.

\textbf{Asynchronous stochastic optimization.}
For solving general optimization problems, asynchronous stochastic methods have received much attention recently. Due to the possible speedup that can be achieved by asynchronous optimization, it has also been extensively applied to various machine learning areas including RL \citep{A3C,sadeghi2018optimal} and distributed learning \citep{tianyu2018decentralized}.
The study of asynchronous stochastic methods can be traced back to 1980s \citep{Bertsekas1989Parallel}. With the batch size $M$, \citep{Agarwal2011Async} analyzed asynchronous SGD (async-SGD) for convex functions, and derived a convergence rate of $\mathcal{O}(K^{-\frac{1}{2}}M^{-\frac{1}{2}})$ if delay $K_0$ is bounded by $\mathcal{O}(K^{\frac{1}{4}}M^{-\frac{3}{4}})$. This result implies linear speedup.
% \citep{Dekel2012Async} proposed an asynchronous mini-batch stochastic gradient method and established a convergence rate of $\mathcal{O}(K^{-\frac{1}{2}} M^{-\frac{1}{2}})$ for convex functions. A linear speedup will be achieved if $K$ is bounded by $\mathcal{O}(K^{\frac{1}{4}} M^{-\frac{3}{4}})$.
\citep{Feyzmahdavian2015Async} extended the analysis of \citep{Agarwal2011Async} to smooth convex with nonsmooth regularization and derived a similar rate. Recent studies by \citep{lian2016speedup} improved upper bound of $K_0$ to $\mathcal{O}(K^{\frac{1}{2}} M^{-\frac{1}{2}})$. However, all these works have focused on the single-timescale SGD with a single variable, which cannot capture the stochastic recursion of the AC and A3C algorithms.
To best of our knowledge, non-asymptotic analysis of asynchronous two-timescale SGD has remained unaddressed, and its speedup analysis is an uncharted territory.

\vspace{-0.2cm}

\subsection{This work}
In this context, we revisit A3C with TD(0) for the critic update. The goal is to provide \emph{non-asymptotic} guarantee and \emph{linear speedup} justification for this popular algorithm.

\textbf{Our contributions.}
Compared to
the existing literature on both the AC algorithms and the async-SGD, our contributions can be summarized as follows.
 
% \begin{itemize}[leftmargin=0.4cm]
% \item 

\textbf{c1)} We revisit two-timescale A3C 
and establish its convergence rates with both i.i.d. and Markovian sampling. We first proves the local convergence rate for A3C in the general function approximation case, and then proves that A3C achieves global convergence for the softmax policy parameterization. To the best of our knowledge, this is the first non-asymptotic convergence result for 
 \emph{asynchronous parallel} AC algorithms.

% \item 
\textbf{c2)}
We characterize the sample complexity of A3C. 
In the i.i.d. setting, A3C achieves a sample complexity of $\mathcal{O}(\epsilon^{-2.5}/N)$ per worker, where $N$ is the number of workers. 
Compared to the best-known complexity of $\mathcal{O}(\epsilon^{-2.5})$ for i.i.d. two-timescale AC \citep{hong2020ac}, A3C achieves \emph{linear speedup}, thanks to the parallelism and asynchrony. In the Markovian setting, if delay is bounded, the sample complexity of A3C matches the order of the non-parallel AC algorithm \citep{wu2020finite}.
 
% \item 
\textbf{c3)}
We test A3C on a synthetic environment to verify our theoretical guarantees with both i.i.d. and Markovian sampling. We also test A3C on the
% compare A3C with the original A3C algorithm \citep{A3C} on 
classic
control tasks and Atari Games. 
% \textbf{Code is available in the supplementary material.} 
% In both settings, A3C has achieved competitive  performance as A3C.
% \end{itemize}

\textbf{Technical challenges.}
Compared to the recent analysis of nonparallel two-timescale AC in \citep{wu2020finite,xu2020non,hong2020ac}, several new challenges arise due to the parallelism and asynchrony.
 
% \textbf{Parameter delay.} In synchronous algorithms, parameters are updated without delay. Parallel asynchrony leaves out the need to synchronize, while also introduces the staleness of parameters. This will introduce extra error in the analysis of both actor and critic convergence. 

\emph{Markovian noise coupled with asynchrony and delay.} The analysis of two-timescale AC algorithm is non-trivial because of the Markovian noise coupled with both the actor and critic steps. 
Different from the nonparallel AC that only involves a single Markov chain, 
A3C introduces multiple Markov chains (one per worker) that mix at different speeds. This is because at a given iteration, workers collect different number of samples and thus their chains mix to different degrees. As we will show later, the worker with the slowest mixing chain will determine the convergence.

% Because of asynchrony, at a given iteration, workers collect different number of samples so their chains mix to different degrees. The worker with the slowest mixing chain will play the critical role in convergence.

\emph{Linear speedup for SGD with two coupled sequences.} Parallel async-SGD has been shown to achieve linear speedup recently \citep{lian2016speedup,sun2017nips}. Different from async-SGD, asynchronous AC is a two-timescale stochastic \emph{semi-gradient} algorithm for solving the more challenging \emph{bilevel} optimization problem (see \citep{hong2020ac}). The errors induced by asynchrony and delay are intertwined with both the actor and critic updates via a nested structure, which makes the sharp analysis more challenging. Our linear speedup analysis should be also distinguished from that of mini-batch async-SGD \citep{lian2017nips}, where the speedup is a result of \emph{variance reduction} thanks to the larger batch size generated by parallel workers.

% {\blue[intuitively, the mechanism for achieving speedup is different between parallel versions of SGD and AC: For SGD, the larger sample size produced by parallelism reduces sample variance hence improves the convergence speed, while for AC, larger sample size helps the mixing of the Chain. ]} This is a bit tricky since in Markovian setting, we didn't achieve linear speedup. 

\section{Preliminaries} 
\subsection{Markov decision process and policy gradient}
A Markov decision process (MDP) can be described by $\mathcal{M}=\{ \mathcal{S}, \mathcal{A}, \mathcal{P}, R, \gamma \}$, where $\mathcal{S}$ is the state space, $\mathcal{A}$ is the action space, $\mathcal{P}(s'|s,a)$ is the probability of transitioning to $s'\in \mathcal{S}$ from state $s \in \mathcal{S}$ and action $a \in \mathcal{A}$, $r(s,a,s')$ is the reward associated with the transition $(s,a,s')$, and $\gamma \in [0,1)$ is a discount factor. Throughout the paper, we assume the reward $r$ is upper-bounded by a constant $r_{\max}$. A policy $\pi: \mathcal{S}\rightarrow \Delta{(\mathcal{A})}$  is defined as a mapping from the state space $\mathcal{S}$ to the probability distribution over the action space $\mathcal{A}$.  
% \kzst{$\pi: \mathcal{S}\times \mathcal{A}\rightarrow [0, 1]$ is defined as a function to map $s$ and $a$ to a probability, such that $\pi(a|s)$ is the probability of choosing action $a$ under state $s$.}

Considering discrete time $t$ in an infinite horizon, a policy $\pi$ can generate a trajectory $(s_0,a_0,\ldots)$ with $a_t\sim \pi(\cdot|s_t)$ and $s_{t+1}\sim \mathcal{P}(\cdot|s_t,a_t)$.
Given a policy $\pi$, we define the state and state action value functions as
% \begin{subequations}
\begin{align}
     V_{\pi}(s) &\coloneqq  \mathbb{E} \left[ \sum_{t=0}^\infty \gamma^t r(s_t, a_t, s_{t+1})  \mid s_0=s\right], \nonumber\\
    Q_{\pi}(s,a) &\coloneqq  \mathbb{E} \left[ \sum_{t=0}^\infty \gamma^t r(s_t, a_t, s_{t+1})\mid s_0=s, a_0 = a \right]
\end{align}
% \end{subequations}
where $\E$ is taken over the trajectory $(s_0,a_0,s_1,a_1,\ldots)$ generated under policy $\pi$. 
With the above definitions, the advantage function is $A_{\pi}(s,a):=Q_{\pi}(s,a)-V_{\pi}(s)$. 
With $\eta$ denoting the initial state distribution, the discounted state visitation measure induced by policy $\pi$ is defined as $d_{\pi}(s) \coloneqq (1-\gamma) \sum_{t=0}^\infty \gamma^t \mathbb{P}(s_t = s\mid s_0\sim \eta, \pi)$. We also overload the notation and define the state-action visitation distribution $d_{\pi}(s,a) = (1-\gamma) \sum_{t=0}^\infty \gamma^t \mathbb{P}(s_t = s\mid s_0\sim \eta, \pi)\pi(a|s)$. 
In the case where $\pi$ is parameterized by $\theta$, we use $d_\theta$ as shorthand notations for $d_{\pi_\theta}$.

% \newpage
% \subsection{Policy gradient theorem}
The goal of RL is to find an optimal policy $\pi^*$ defined as $\pi^* \in \arg\max_{\pi}J(\pi) \coloneqq (1-\gamma)\E_{s \sim \eta} [V_{\pi}(s)]$, with the optimal return defined as $J^*\coloneqq \max_{\pi} J(\pi)$.
When the state and action spaces are large, finding the optimal policy $\pi$ becomes computationally intractable. 
To overcome the inherent difficulty of
learning a function, the policy gradient methods search the best performing policy over a
class of parameterized policies.
We parameterize the policy with parameter $\theta \in \mathbb{R}^{d}$, and solve the optimization problem as
\begin{equation}\label{eq:formulation-raw}
    \max_{\theta \in \mathbb{R}^{d}} J(\theta)~~~{\rm with}~~~J(\theta) \coloneqq (1-\gamma)\E_{s \sim \eta} [V_{\pi_\theta}(s)].
\end{equation}
% \kznote{why have to write it as $\arg\max$? Can't we just write it as $\max$? Also, the maximizer may not be unique...}
To maximize $J(\theta)$ with respect to $ \theta$, one can update $\theta$ using the policy gradient \citep{SuttonPG}
\begin{equation*}\label{eq:policy-gradient}
    \nabla J(\theta) = \E_{s,a \sim d_\theta} \left[ A_{\pi_\theta}(s,a) \psi_\theta(s,a) \right], \numberthis
\end{equation*}
where $\psi_\theta(s,a) \coloneqq \nabla \log\pi_\theta(a|s)$. 
Since computing $\E$ in \eqref{eq:policy-gradient} is expensive if not impossible, popular policy gradient-based algorithms iteratively update $\theta$ using stochastic estimate of \eqref{eq:policy-gradient} such as REINFORCE \citep{Williams1992} and G(PO)MDP \citep{baxter2001jair}. 

 It is also a common practice to adopt regularization and augment the objective function to
\begin{align}\label{definition:jlambda}
    J_\lambda (\theta) \coloneqq J(\theta) - \lambda \E_{s\sim \eta_p}\big[D_{KL}(\pi_p(\cdot|s)|\pi_\theta(\cdot|s))\big] 
\end{align}
with a regularization constant $\lambda \geq 0$. Here $\eta_p$ is a prior distribution of states, $\pi_p$ is a prior policy. The regularization term encourages $\pi_\theta$ to imitate $\pi_p$, incorporating prior knowledge into training process. When $\pi_p$ and $\eta_p$ are set as uniform distributions, the regularization term is reduced to the relative-entropy regularization  widely analyzed in the literature \citep{agarwal2019optimality, bhandari2020global,junzi2021sample}. Moreover, the regularization prevents degenerate solutions that can lead to the pitfall of certain policy parametrization \citep{bhandari2020global}. Given $\pi_p$ and $\eta_p$, we use $R(\theta)$ as a shorthand notation of $-\E_{s\sim \eta_p}\big[D_{KL}(\pi_p(\cdot|s)|\pi_\theta(\cdot|s))\big]$.

\subsection{Actor-critic with value function approximation}
Both REINFORCE and G(PO)MDP-based policy gradient algorithms rely on a Monte-Carlo estimate of the value function $V_{\pi_\theta}(s)$ and thus $\nabla J(\theta)$ by generating a trajectory per iteration. 
However, policy gradient methods based on Monte-Carlo estimate typically suffer from high variance and large sampling cost. 
An alternative way is to recursively refine the estimate of $V_{\pi_\theta}(s)$. 
For a policy $\pi_\theta$, it is known that $V_{\pi_\theta}(s)$ satisfies the Bellman equation \citep{SuttonRL}, that is 
\begin{equation}
V_{\pi_\theta}(s)=   \E_{ a \sim \pi_\theta(\cdot|s),\, s' \sim \mathcal{P}(\cdot|s,a)} \left[ r(s,a,s') + \gamma V_{\pi_\theta}(s')\right].
\end{equation}
In practice, when the state space $\mathcal{S}$ is prohibitively large, one cannot afford the computational and memory complexity of computing $V_{\pi_\theta}(s)$ and $A_{\pi_\theta}(s,a)$. 
To overcome this curse-of-dimensionality, a popular method is to approximate the value function using function approximation techniques. Given the state feature mapping $\phi(\cdot):\mathcal{S}\xrightarrow[]{}\mathbb{R}^{d'}$ for some $d'>0$, we approximate the value function linearly as $V_{\pi_\theta}(s) \approx \hat{V}_\omega(s):=\phi(s)^\top \omega$, where $\omega \in \mathbb{R}^{d'}$ is the critic parameter.   
% We also assume that for all $s \in \mathcal{S}$, the state feature vector $\phi(s)$ is normalized so that $\|\phi(s)\|_2 \leq 1$.
% Then we can write (\ref{eq:formulation-raw}) instead as a bi-level optimization problem:
% \begin{align*}
%     \theta^* = \arg\max_{\theta \in \mathbb{R}^{d}} J(\theta)=\E_{s \sim \eta} [V_{\omega^*_\theta}(s)] subject to \omega^*_\theta \in \arg\min 
% \end{align*}

Given $\pi_\theta$, the task of finding the best $\omega$ such that $V_{\pi_\theta}(s) \approx \hat{V}_\omega(s)$ is usually addressed by TD learning \citep{sutton1988td}.
Given $\pi_\theta$, the task of finding the best $\omega$ such that $V_{\pi_\theta}(s) \approx \hat{V}_\omega(s)$ is usually addressed by TD learning \citep{sutton1988td}. Formally, we first define 
\begin{subequations}\label{def:A-b}
\begin{align}
A_{\theta, \phi} &\coloneqq \E_{ s\sim \mu_{\pi_\theta}, s'\sim \mathcal{P}_{\pi_\theta}}[\phi(s)(\gamma\phi(s')-\phi(s))^\top],\\
b_{\theta, \phi} &\coloneqq \E_{ s\sim \mu_{\pi_\theta}, a \sim \pi_\theta}[r(s,a,s')\phi(s)].
\end{align}
\end{subequations}
where $\mathcal{P}_{\pi_\theta} (s'|s) \coloneqq \sum_{a}\mathcal{P}(s'|s,a)\pi_\theta(a|s)$, and $\mu_{\pi_\theta}$ is the stationary distribution of the Markov chain with transition distribution $\mathcal{P}$ and policy $\pi_\theta$. Then given a policy $\pi_\theta$, the \textit{exact} TD update takes the following form:
\begin{align}\label{eq:TDupdate}
    \omega_{k+1} = \omega_k + \beta \big( A_{\theta, \phi}\omega_k + b_{\theta, \phi} \big).
\end{align}

When analyzing TD, the following standard assumption is often made:
\begin{Assumption}\label{assumption:A}
For all $s \in \mathcal{S}$, the feature vector $\phi(s)$ is normalized so that $\|\phi(s)\|_2 \leq 1$. 
 For all eligible $\theta$, the symmetric part of $A_{\theta,\phi}$, denoted as $(A_{\theta,\phi}+A_{\theta,\phi}^\top)/2$, is negative definite and has a largest eigenvalue upper bounded by $-\lambda$.
\end{Assumption}
Assumption \ref{assumption:A} is common in analyzing TD with linear function approximation; see e.g.,  \citep{JBTD,wu2020finite,wu2021byzantine}. In fact, as shown in \citep{JBTD}, when redundant features are removed such that the feature covariance matrix is full-rank, this assumption is satisfied. With this assumption, $A_{\theta, \phi}$ is full-rank, thus the update in \eqref{eq:TDupdate} admits a unique stationary point $\omega^*(\theta)=-A_{\theta, \phi}^{-1}b_{\theta, \phi}$. Moreover, there exists a constant $R_\omega \coloneqq \frac{r_{\max}}{\sqrt{\lambda}(1-\gamma)^{\frac{3}{2}}}$ such that $\|\omega^*(\theta)\|_2 \leq R_\omega$.

We often use the stochastic approximation of the TD update in \eqref{eq:TDupdate}. With $k$th transition defined as $x_k \coloneqq (s_k,a_k,s_{k+1})$, the corresponding TD target is
\begin{equation}
    \hat{\delta}(x_k,\omega_k) \coloneqq r(s_k,a_k,s_{k+1}) + \gamma \phi(s_{k+1})^\top \omega_k - \phi(s_k)^\top \omega_k
\end{equation}
% $\hat{\delta}(x_k,\omega_k) \coloneqq r(s_k,a_k,s_{k+1}) + \gamma \phi(s_{k+1})^\top \omega_k - \phi(s_k)^\top \omega_k$ 
and the critic gradient $g(x_k,\omega_k) := \hat{\delta}(x_k,\omega_k) \nabla \hat{V}_{\omega_k}(s_k)$.
% $g(x_k,\omega_k) := \hat{\delta}(x_k,\omega_k) \nabla \hat{V}_{\omega_k}(s_k)$, 
% \begin{equation}
%     g(x_k,\omega_k) := \hat{\delta}(x_k,\omega_k) \nabla \hat{V}_{\omega_k}(s_k)
% \end{equation}
We update the parameter $\omega$ via
\begin{equation}\label{eq:td-update}
    \omega_{k+1} = \Pi_{R_\omega}\big(\omega_k + \beta  g(x_k,\omega_k)\big),
\end{equation}
where $\beta$ is the critic step size, and $\Pi_{R_\omega}$ is a projection operator that projects a vector to a $l_2$ norm ball with radius $R_\omega$. 
The projection step is often used to control the norm of  gradient. In AC, it prevents the actor and critic updates from going too far in the ‘wrong’ direction; see e.g., \citep{KondaAC,wu2020finite,xu2020non,zou2019sarsa}.
% {\red[mention projection, or even the projected Bellman equation here?]}

\begin{figure}[t]
\centering
    \includegraphics[width=0.45\textwidth]{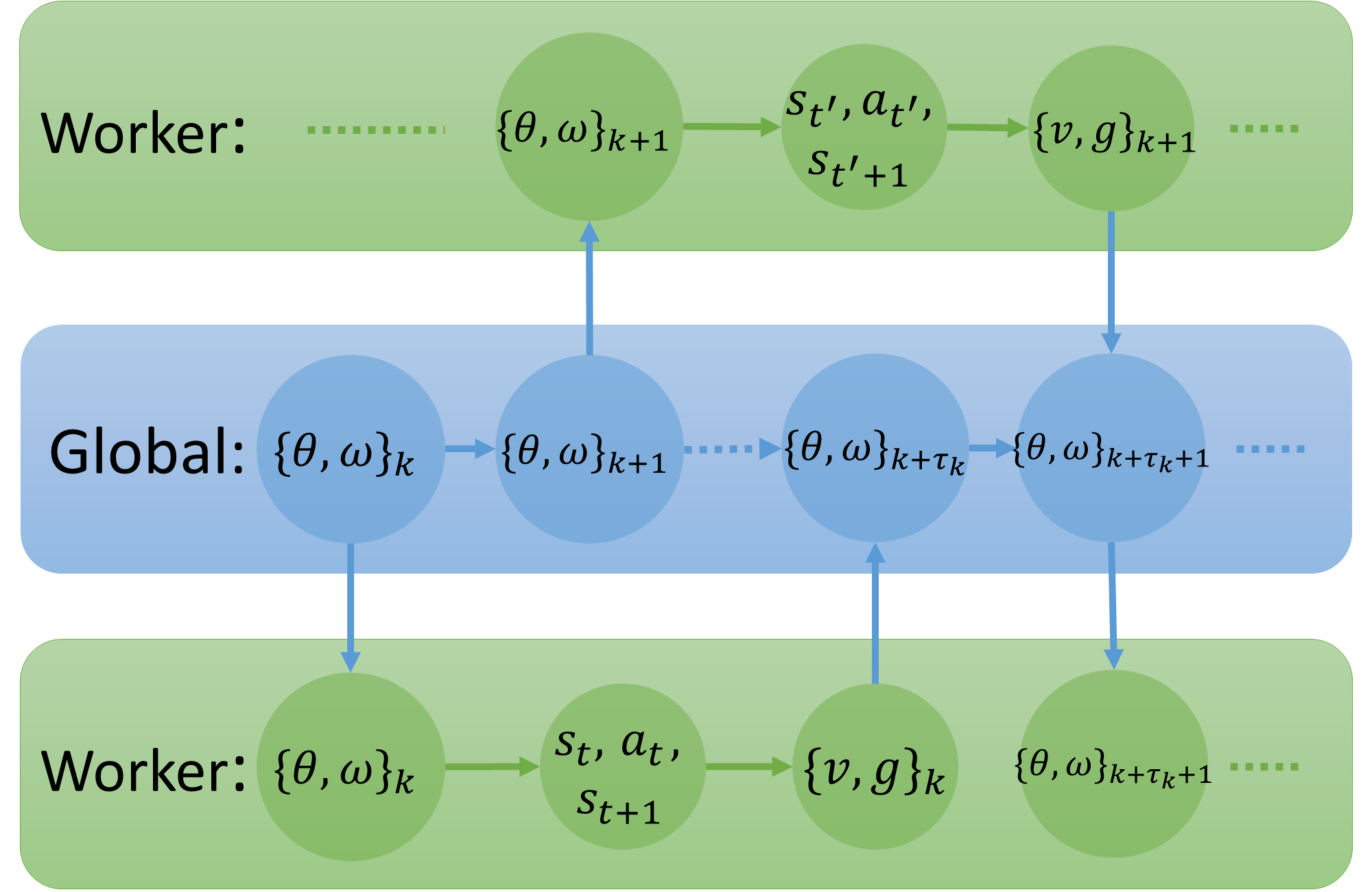}
    \caption{Implementation of A3C with two workers.}
    \label{fig:algorithm}
\end{figure}

Using the definition that $ A_{\pi_\theta}(s,a) = \E_{s' \sim \mathcal{P}}[r(s,a,s') + \gamma V_{\pi_\theta}(s')] - V_{\pi_\theta}(s)$,
% \kznote{this definition of advantage is not correct. The first term is not a Q-function, as you take expectation over $a$..} 
we can also rewrite \eqref{eq:policy-gradient} as $\nabla J(\theta) = \E_{ s,a \sim d_\theta, s' \sim \mathcal{P}} \left[ \left( r(s,a,s') + \gamma V_{\pi_\theta}(s') - V_{\pi_\theta}(s)\right) \psi_\theta(s,a) \right]$.
% \begin{align*}\label{eq:policy-gradient-2}
%     \nabla J(\theta) = \E_{\substack{s,a \sim d_\theta \\ s' \sim \mathcal{P}(\cdot|s,a)}} \left[ \left( r(s,a,s') + \gamma V_{\pi_\theta}(s') - V_{\pi_\theta}(s)\right) \psi_\theta(s,a) \right]. \numberthis
% \end{align*}
Leveraging the value function approximation, we can then approximate the regularized policy gradient as
\begin{equation}\label{eq:pg-adv}
    \widehat{\nabla} J_\lambda(\theta) = \widehat{\nabla} J(\theta) + \lambda \widehat{\nabla} R(\theta) =  \underbrace{\hat{\delta}(x,\omega)\psi_\theta(s,a)}_{v(x,\theta, \omega) } + \lambda \psi_\theta(x^{p}).
\end{equation}
where $v(x,\theta, \omega)$ is an estimator of $\nabla J(\theta)$, and $x^{p}\coloneqq \big(s^p \sim \eta_p,a^p \sim \pi_p(\cdot|s_p)\big)$. Then it is easy to check that $\psi_\theta(x^{p})$ is an unbiased estimator of $\nabla R(\theta)$. This gives rise to the policy update 
\begin{equation}\label{eq:pg-update}
    \theta_{k+1}  = \theta_k + \alpha \big(v(x_k,\theta_k, \omega_k)+\lambda \psi_\theta(x^{p})\big),
\end{equation}
where $\alpha$ is the stepsize for the actor update. 
To ensure convergence when simultaneously performing critic and actor updates, the stepsizes $\alpha$ and $\beta$ often decay at two different rates, which is referred to the two-timescale AC  \citep{KondaAC,wu2020finite}.

% \begin{wrapfigure}{r}{2.8in}
% \vspace{-0.4cm}
% \def\epsfsize#1#2{0.25#1}
% \centerline{\epsffile{images/pics.png}}
%   \caption{Implementation of A3C with two workers.}
% \label{fig:algorithm}
% \vspace{-0.4cm}
% \end{wrapfigure}

\section{A3C Implementation}
% \red{@Han Add a figure here to show how the algorithm is being implemented since it is more complex than traditional async SGD.}

% \begin{figure}[h]
%     \centering
%     \includegraphics[width=.4\textwidth]{images/pics.png}
%     \caption{Implementation of A3C with two workers. }
%     \label{fig:algorithm}
% \end{figure}

To speed up the training process, AC can be implemented over $N$ workers in a shared memory setting {\it without} coordinating among workers \citep{A3C}.
Each worker has its own simulator to perform sampling, and then collaboratively updates the shared policy $\pi_\theta$ using AC updates. 
% Our algorithm runs in an asynchronous manner \citep{A3C}, which typically involves multiple workers and a shared memory where parameters are stored.
% In general, the workers collect samples and compute stochastic gradients locally. Once gradients are obtained, workers can directly access the shared memory and update the parameters. 
As there is no synchronization after each update, the policy used by workers to generate samples may be outdated, which introduces staleness.

\textbf{Notations on samples.} Subscription $t$ in $x_t$ and $x^p_t$ indicates the sample is generated in $t$th local iteration of a worker. When Markovian sampling is used, subscription $t$ in $x_t=(s_t,a_t,s_{t+1})$ also indicates that it is the $t$th transition of the local Markov chain.
We use $k$ to denote the global counter (or iteration), which increases by one whenever a worker finishes the actor and critic updates in the shared memory. 
We use subscription $(k)$ in $(s_{(k)},a_{(k)},s'_{(k)})$ and $(s^p_{(k)},a^p_{(k)})$ to indicate the samples used in the $k$th update.

\begin{algorithm}[t]
\caption{A3C: each worker's view.}\label{algorithm:async-tts-worker}
\begin{algorithmic}[1]
  \STATE \textbf{Global initialize:} Global counter $k\!=\!0$, initial $\theta_0$, $\omega_0$ in the shared memory.
    \STATE\textbf{Worker initialize:} Counter $t\!=\!0$. Sample $s_0 \!\sim\! \eta$, $\hat{s}_0 \!\sim\! \eta$. 
    % Bernoulli variable $b$ with $\mathbb{P}(b\!=\!1)\!=\!\gamma$.
    % Pull $\theta, \omega$ from shared memory.
\FOR{$t=0,1,2,\cdots$}
\STATE\hspace{+0.0cm} Read $\theta, \omega$ in the shared memory.
\STATE\hspace{+0.0cm} \textbf{option 1 (i.i.d. sampling):}\
    \STATE\hspace{+0.3cm} $x_t = \big( s_t \sim \mu_{\pi_{\theta_t}}$, $a_t \sim \pi_{\theta_t}(\cdot|s_t)$, $s'_{t} \sim \mathcal{P}(\cdot|s_t,a_t) \big)$.
    \STATE\hspace{+0.3cm} $\hat{x}_t = \big( \hat{s}_t \sim d_{\pi_{\theta_t}}$, $\hat{a}_t \sim \pi_{\theta_t}(\cdot|\hat{s}_t)$, $\hat{s}'_{t} \sim \mathcal{P}(\cdot|\hat{s}_t,\hat{a}_t) \big)$.
\STATE\hspace{+0.0cm} \textbf{option 2 (Markovian sampling):}
    \STATE\hspace{+0.3cm} $x_t = \big(s_t,a_t \sim \pi_\theta(\cdot|s_t)$, $s_{t+1} \sim \mathcal{P}(\cdot|s_t,a_t) \big)$.
    \STATE\hspace{+0.3cm} $\hat{x}_t = \big(\hat{s}_t,\hat{a}_t \sim \pi_\theta(\cdot|\hat{s}_t)$, $s'_{t+1} \sim \mathcal{P}(\cdot|\hat{s}_t,\hat{a}_t)$\big).
    \STATE\hspace{+0.3cm} With probability $\gamma$: $\hat{s}_{t+1}\!=\!s'_{t+1}$; Otherwise: $\hat{s}_{t+1}\!\sim \!\eta$.
     
    \STATE\hspace{+0.0cm} Compute $g(x_t,\omega) = \hat{\delta}(x_t,\omega) \nabla_\omega \hat{V}_\omega(s_t)$.
    \STATE\hspace{+0.0cm} Compute $v(\hat{x}_t,\theta, \omega)= \hat{\delta}(\hat{x}_t,\omega) \psi_\theta(\hat{s}_t,\hat{a}_t)$.
    \STATE\hspace{+0.0cm} Compute $\psi_\theta (x_t^p)$ with $x_t^p=(s_t^p \sim \eta_p,a_t^p\sim\pi_p (\cdot|s_t^p))$.
    \STATE\hspace{+0.0cm} In the shared memory, perform update \eqref{algorithm:async-tts-update}.
\ENDFOR
\end{algorithmic}
\end{algorithm}
\setlength{\textfloatsep}{10pt}
% \vspace*{-0.1cm}

\textbf{Algorithm flow.} Specifically, we initialize $\theta_0$, $\omega_0$ in the shared memory. Each worker will initialize the simulator with initial state $s_0$.  
Without coordination, workers will load $\theta$, $\omega$ in the shared memory. The worker then generates samples with either i.i.d. or Markovian sampling method. In Markovian sampling case, we maintain separate Markov chains for actor and critic. For critic, we generate samples following the original transition kernel $\mathcal{P}$. While the actor's chain can be viewed as evolving under a transition kernel $\hat{\mathcal{P}}=\gamma \mathcal{P} + (1-\gamma)\eta$. At each iteration, we have a probability of $1-\gamma$ to reset the chain, thus taking the initial state distribution into account. If the actor's chain evolves under $\mathcal{P}$ like critic, asymptotically the initial distribution $\eta$ is forgotten, which will introduce an asymptotic error.
Once samples are obtained, each worker locally computes the gradients, and then updates the parameters in shared memory asynchronously by 
% {\red[mention which step (e.g., the constructing and sampling from $\pi_{\theta}$) is expensive, and where we have allowed parallel sampling.)]}
\begin{subequations}\label{algorithm:async-tts-update}
\begin{align}
    &\omega_{k+1} = \Pi_{R_\omega}\left(\omega_k + \beta g(x_{(k)},\omega_{k-\tau_k})\right) \\
    &\theta_{k+1} \!=\! \theta_k \!+\! \alpha \big( v(\hat{x}_{(k)},\theta_{k-\tau_k}, \omega_{k-\tau_k}) \!+\! \lambda \psi_{\theta_{k-\tau_k}}(x^p_{(k)})\big)
\end{align}
\end{subequations}
 where $\tau_k$ is the delay in the $k$th actor and critic updates. See A3C in Algorithm \ref{algorithm:async-tts-worker} and Figure \ref{fig:algorithm}.

 \textbf{Parallel sampling.}
 The AC update \eqref{eq:td-update} and \eqref{eq:pg-update} uses samples generated ``on-the-fly'' from the target policy $\pi_\theta$, which brings overhead. Compared with \eqref{eq:td-update} and \eqref{eq:pg-update}, the A3C update \eqref{algorithm:async-tts-update} allows parallel sampling from $N$ workers, which is the key to linear speedup. 
We consider the case where only one worker can update parameters in the shared memory at the same time and the update cannot be interrupted. 
In practice, \eqref{algorithm:async-tts-update} can also be performed in a mini-batch fashion. 

 \textbf{Separate sampling protocols.}
In Algorithm \ref{algorithm:async-tts-worker}, we maintain separate sampling protocols for actor and critic. This is due to the mismatch between the actor and critic sampling distribution. As indicated by \eqref{def:A-b}, the desired sampling distribution of critic is $\mu_{\pi_\theta}$.  The policy gradient \eqref{eq:policy-gradient} requires sampling from $d_{\pi_\theta}$. 
However, $d_{\pi_\theta}$ and $\mu_{\pi_\theta}$ are in general different, and the difference is non-diminishing. Therefore, if one uses the same samples for actor and critic, either the actor or the critic update will have a non-diminishing bias.

To mitigate the asymptotic bias, it is just natural to choose different sampling protocols for actor and critic. Our theoretical analysis justifies this choice by proving that such sampling method gives unbiased stochastic gradients asymptotically. We also provide experiments to demonstrate the superiority of the separated sampling methods.
% Nevertheless, practitioners still use the \emph{same} sampling protocol for both actor and critic updates (see e.g., \citep{espeholt2018impala,A3C}). 
% We use the update with a single sample to simplify the notations. 
% In other words, $(s_{(k)},a_{(k)})$ and $(s_{k+1},a_{k+1})$ only forms a Markov chain if they are both sampled by the same worker with Markovian sampling method.

% {\red[briefly discuss the differences and advantages/disadvantages of two sampling methods?]}

% \textbf{Minor differences from A3C \citep{A3C}.}
% The A3C algorithm in Algorithm \ref{algorithm:async-tts-worker} resembles the original A3C method \citep{A3C}.
% Original A3C uses $n$-step TD target while Algorithm \ref{algorithm:async-tts-worker} uses the $1$-step TD target to compute the updates. When $n=1$, A3C reduces to Algorithm \ref{algorithm:async-tts-worker}. When $n> 1$, our result can be extended to the $n$-step TD setting by leveraging the error reduction property of $n$-step TD. As shown in \cite[Eq. (7.3)]{SuttonRL}, the $n$-step TD target is guaranteed to improve over $1$-step target in terms of the expected approximation error. In this paper, we analyze the $1$-step TD target for ease of notations.

\section{Convergence Analysis of A3C}\label{section:convergence}
In this section, we analyze the convergence of A3C in both i.i.d. and Markovian settings. Throughout this section, ${\mathcal{O}}(\cdot)$ contains constants that are independent of $N$ and $K_0$.

To analyze the performance of A3C, we make the following assumptions.
\begin{Assumption}\label{assumption:delay}
There exists $K_0$ such that the delay at each iteration is bounded by $\tau_k \leq K_0, \forall k$.
\end{Assumption}
\vspace*{-0.2cm}
    
Assumption \ref{assumption:delay} ensures the viability of analyzing the asynchronous update; see the same assumption in e.g., \citep{lian2016speedup,assran2019gossip,tianyu2018decentralized}. In practice, the delay usually scales as the number of workers, that is $K_0=\Theta(N)$.

\begin{Assumption}\label{assumption:omega}
For any $\theta, \theta' \in \mathbb{R}^{d}$, $s \in \mathcal{S}$ and$a \in \mathcal{A}$, there exist  constants $C_\psi,L_\psi,L_\pi$ such that: i) $\|\psi_\theta(s,a)\|_2 \leq C_\psi$; ii) $\|\psi_\theta(s,a) - \psi_{\theta'}(s,a)\|_2 \leq L_\psi\|\theta-\theta'\|_2$; iii) $\left|\pi_\theta(a|s)-\pi_{\theta'}(a|s)\right| \leq L_\pi\|\theta-\theta'\|_2$.  
% \begin{enumerate}[leftmargin=0.8cm]
%     \item  $\|\psi_\theta(s,a)\|_2 \leq C_\psi$.
%     \item $\|\psi_\theta(s,a) - \psi_{\theta'}(s,a)\|_2 \leq L_\psi\|\theta-\theta'\|_2$.
%     \item\label{assumption:omega-Lpi} $\left|\pi_\theta(a|s)-\pi_{\theta'}(a|s)\right| \leq L_\pi\|\theta-\theta'\|_2$.
% \end{enumerate}
\end{Assumption}
\vspace*{-0.2cm}

Assumption \ref{assumption:omega} is common in analyzing policy gradient-type algorithms which has also been made by e.g., \citep{agarwal2019optimality,zhang2019global}. 
% It assumes the boundedness and Lipschitiz continuity of score function $\psi_\theta(s,a)$, and the Lipschitz continuity of policy. 
This assumption holds for many policy parameterization methods such as tabular softmax policy \citep{agarwal2019optimality}, Gaussian policy \citep{doya2000reinforcement} and Boltzmann policy \citep{konda1999actor}.

\begin{Assumption}\label{assumption:MDP}
For any $\theta$, assume the Markov chains with transition kernels $\mathcal{P}$ and $\hat{\mathcal{P}}$ are irreducible and aperiodic under policy $\pi_\theta$. Then there exist constants $\kappa > 0$ and $\rho \in (0,1)$ such that
\begin{subequations}
\begin{equation}\label{eq:pmixing}
    \sup_{s \in \mathcal{S}}~~~d_{TV}\left(\mathbb{P}(s_t \in \cdot|s_0=s, \pi_\theta),\mu_{\pi_\theta}\right)
    \leq \kappa \rho^t,
\end{equation}
and
\begin{equation}\label{eq:hatpmixing}
    \sup_{s \in \mathcal{S}}~~~d_{TV}\left(\mathbb{P}(\hat{s}_t \in \cdot|\hat{s}_0=s, \pi_\theta),d_{\pi_\theta}\right)
    \leq \kappa \rho^t.
\end{equation}
\end{subequations}
where $s_t$ is the $t$th state of the Markov chain with transition kernel $\mathcal{P}$, and $\hat{s}_t$ is the $t$th state of the Markov chain with transition kernel $\hat{\mathcal{P}}$.
% starting from $s_0$ and under the policy $\theta$.
\end{Assumption}

Assumption \ref{assumption:MDP} assumes the Markov chain mixes at a geometric rate. This assumption has also been made by other analysis on Markovian sampling; see e.g. \citep{JBTD,wu2020finite}. It is worth noting that the second part of our assumption, that is \eqref{eq:hatpmixing}, holds as long as $\gamma < 1$.

We define the critic approximation error as 
\begin{equation}\label{eq:epsilon_app}
    \epsilon_{\rm app} \coloneqq \max_{\theta \in \mathbb{R}^{d}} \sqrt{\E_{s \sim \mu_{\theta}}|V_{\pi_\theta}(s)-\hat{V}_{\omega^*_{\theta}}(s)|^2}
\end{equation}
where $\mu_{\theta}$ is the stationary distribution under $\pi_\theta$ and $\mathcal{P}$. This error captures the quality of the critic function approximation; see also \citep{qiu2019finite,wu2020finite,wu2021byzantine}. When the MDP is tabular and the feature matrix is full-rank, the value function $V_{\pi_\theta}$ is in the span of the features. In this case, we have $\epsilon_{\rm app}=0$. 

%  Next we will look into another important problem in actor-critic algorithms under discounted MDP. To demonstrate the problem intuitively, we will consider the case where the actor update is slow so that the policy is almost stationary. Then by Assumption \ref{assumption:MDP}, sampling  the marginal state distribution will converge to the stationary distribution $\mu_{\theta}$ which is indeed the distribution required by the critic update as indicated by \ref{def:A-b}. However, the policy gradient in \eqref{eq:policy-gradient} requires sampling from the discounted visitation measure $d^{'}_{\theta}$. The mismatch between

We first give the convergence result of critic update.
\begin{Theorem}[Critic convergence]\label{theorem:async-tts-critic-iid-double}
Suppose Assumptions \ref{assumption:A}--\ref{assumption:MDP} hold. Consider Algorithm \ref{algorithm:async-tts-worker} with i.i.d. sampling and $\hat{V}_\omega(s)=\phi(s)^\top \omega$. Select step size $\alpha = K^{-\frac{3}{5}}$ and $\beta = K^{-\frac{2}{5}}$. Then it holds that
\begin{align}\label{eq:theorem:async-tts-critic-iid-double}
\frac{1}{K}\sum_{k=1}^K \E\left\|\omega_k-\omega_{\theta_k}^*\right\|_2^2
    \!&=\!
 \mathcal{O}\left(\frac{K_0^2}{K^{\frac{4}{5}}}\right)
    \!+\! \mathcal{O}\left(\frac{K_0}{K^{\frac{3}{5}}}\right)\!+\!\mathcal{O}\left(\frac{1}{K^{\frac{2}{5}}}\right).
\end{align}
% where ${\mathcal{O}}(\cdot)$ contains constants that are independent of $N$.
%     \frac{1}{K}\sum_{k=1}^K \E\left\|\omega_k-\omega_{\theta_k}^*\right\|_2^2 \!=\!
%     \mathcal{O}\left(\frac{1}{K^{1-\sigma_2}}\right)
%     \!+\! \mathcal{O}\left(\frac{1}{K^{2(\sigma_1-\sigma_2)}}\right) 
%     \!+\! \mathcal{O}\left(\frac{K_0^2}{K^{2\sigma_2}}\right)
%     \!+\!\mathcal{O}\left(\frac{K_0}{K^{\sigma_1}}\right)
%     \!+\!\mathcal{O}\left(\frac{1}{K^{\sigma_2}}\right).
% \end{align*}
\end{Theorem}

Given the critic convergence, we can present the convergence result of actor update.
\begin{Theorem}[Actor convergence]\label{theorem:async-tts-actor-iid-double}
Under the same assumptions of Theorem \ref{theorem:async-tts-critic-iid-double}, select step size $\alpha = K^{-\frac{3}{5}}$ and $\beta = K^{-\frac{2}{5}}$. Then it holds that
\begin{align}\label{eq:theorem:async-tts-actor-iid-double}
    \frac{1}{K}\sum_{k=1}^K \E\|\nabla J_\lambda(\theta_k)\|_2^2 =\mathcal{O}\Big(\frac{1}{K^{\frac{2}{5}}}\Big)+\mathcal{O}\Big(\frac{K_0^2}{K^{\frac{4}{5}}}\Big)+\mathcal{O}\Big(\frac{K_0}{K^{\frac{3}{5}}}\Big)+\mathcal{O}(\epsilon_{\rm app}).
\end{align}
If $K_0=\Theta(N)=\mathcal{O}(K^{\frac{1}{5}})$, then it holds that
\begin{equation}\label{eq.theorem:async-tts-iid-double}
    \frac{1}{K}\sum_{k=1}^{K}\E\|\nabla J_\lambda(\theta_k)\|_2^2
    =\mathcal{O}\big(K^{-\frac{2}{5}}\big)+\mathcal{O}(\epsilon_{\rm app})
\end{equation}
where ${\mathcal{O}}(\cdot)$ contains constants independent of $N$ and $K_0$.
\end{Theorem}

\begin{Corollary}[Linear speedup]\label{corollary:linear-speedup}
To reach $\epsilon$-accuracy in \eqref{eq.theorem:async-tts-iid-double}, the required number of iterations is $\mathcal{O}(\epsilon^{-2.5})$. Since each iteration of A3C only uses one sample (one transition), the sample complexity is $\mathcal{O}(\epsilon^{-2.5})$, which matches the state-of-the-art sample complexity of two-timescale AC running on one worker. 
Then under A3C, the average sample complexity per worker is $\mathcal{O}({\epsilon^{-2.5}}/{N})$ which indicates linear speedup in \eqref{eq.speedup}. The negative effect of parameter staleness introduced by parallel asynchrony vanishes asymptotically with the step size. Vanished staleness allows for parallel computing from workers to speedup the training process.
\end{Corollary}

% By setting the first term in \eqref{eq.theorem:async-tts-iid} to $\epsilon$, we get the total iteration complexity to reach $\epsilon$-accuracy is $\mathcal{O}(\epsilon^{-2.5})$. Since each iteration only uses one sample (one transition), it also implies a total sample complexity of $\mathcal{O}(\epsilon^{-2.5})$.
% Corollary \ref{corollary:linear-speedup} implies that under the assumption of $K_0 = \mathcal{O}(K^\frac{1}{5})$, A3C achieves convergence rate of $\mathcal{O}(K^{-\frac{2}{5}})$.
% Since the sample complexity does not depend on $K_0=\Theta(N)$, the sample complexity to achieve $\epsilon$-accuracy remains unchanged as $N$ increases. 

\begin{Remark}[Comparison to async-SGD analysis]
Different from async-SGD (e.g., \citep{lian2016speedup}), the optimal critic parameter $\omega_{\theta}^*$ is constantly drifting as $\theta$ changes at each iteration. 
This necessitates setting the actor update to be at a faster time scale than the critic. In this sense, the policy is static relative to the critic asymptotically. 
In actor update, the gradient $v(x,\theta,\omega)$ is biased because of inexact value function. The bias introduced by the critic optimality gap and the function approximation error correspond to the last two terms in \eqref{eq:theorem:async-tts-actor-iid-double}.
\end{Remark}

\subsection{Convergence result with Markovian sampling}

% Now we are ready to first present the critic convergence result.

% Suppose Assumption \ref{assumption:delay} holds. Given the definition of $m_K$, we also define:
% As explained by (\ref{eq:M_K-derivation}) in Lemma \ref{lemma:tts-critic-noise}, $M_K$ is the upper bound to ensure that a worker can collect at least $m_K$ samples up to any given iteration step. $M_K$ is later used in the proof to bound the error term caused by Markovian sampling noise and the drifting of parameters.
\begin{Theorem}[Critic convergence]\label{theorem:async-tts-critic}
Suppose Assumptions \ref{assumption:A}--\ref{assumption:MDP} hold. Consider Algorithm \ref{algorithm:async-tts-worker} with Markovian sampling and $\hat{V}_\omega(s)=\phi(s)^\top \omega$. Select step size $\alpha = K^{-\frac{3}{5}}$ and $\beta = K^{-\frac{2}{5}}$. Then it holds that
\begin{align}\label{eq:theorem:async-tts-critic}
  \frac{1}{K}\sum_{k=1}^K \E\left\|\omega_k-\omega_{k}^*\right\|_2^2 
  = \mathcal{O}\left(\frac{1}{K^{\frac{2}{5}}}\right)
    + \mathcal{O}\left(\frac{K_0^2\log^2 K}{K^{\frac{3}{5}}}\right)
    + \mathcal{O}\left(\frac{K_0 \log K}{K^{\frac{2}{5}}}\right).
\end{align}
\end{Theorem}

The following theorem gives the convergence rate of actor update in Algorithm \ref{algorithm:async-tts-worker}. 
% Note that $\epsilon_{\rm app}$ is defined by (\ref{eq:epsilon_app}). It is considered as a fixed error induced by the critic function approximation method.

\begin{Theorem}[Actor convergence]\label{theorem:async-tts-actor}
Under the same assumptions of Theorem \ref{theorem:async-tts-critic}, select step size $\alpha = K^{-\frac{3}{5}}$ and $\beta = K^{-\frac{2}{5}}$. Then it holds that
\begin{align}\label{eq:theorem:async-tts-actor}
 \frac{1}{K}\sum_{k=1}^{K}\E\|\nabla J_\lambda(\theta_k)\|_2^2
= \mathcal{O}\left(\frac{K_0^2\log^2 K}{K^{\frac{3}{5}}}\right)
    + \mathcal{O}\left(\frac{K_0 \log K}{K^{\frac{2}{5}}}\right)
    + \mathcal{O}\left(\epsilon_{\rm app} \right).
\end{align}
If we further assume $K_0=\Theta(N)=\mathcal{O}(K^{\frac{1}{5}})$. It holds that
\begin{equation}
    \frac{1}{K}\sum_{k=1}^{K}\E\|\nabla J_\lambda(\theta_k)\|_2^2
    =
    \widetilde{\mathcal{O}}\left(K_0 K^{-\frac{2}{5}}\right)
    +\mathcal{O}(\epsilon_{\rm app})
\end{equation}
where $\widetilde{\mathcal{O}}(\cdot)$ hides constants and the logarithmic order of $K$.
\end{Theorem}

% Given Theorem \ref{theorem:async-tts-critic} and Theorem \ref{theorem:async-tts-actor}, choosing $\sigma_1$ and $\sigma_2$ properly gives the following convergence rate.
% \begin{Corollary}\label{corollary:convergence}
% % Given Theorems \ref{theorem:async-tts-critic} and \ref{theorem:async-tts-actor}, 
% Assume $K_0=\mathcal{O}(K^{\frac{1}{5}})$. Select $\sigma_1 = \frac{3}{5}$ and $\sigma_2 = \frac{2}{5}$, then it holds that
% \begin{equation}
% \small
%     \frac{1}{K}\sum_{k=1}^{K}\E\|\nabla J(\theta_k)\|_2^2
%     =
%     \widetilde{\mathcal{O}}\left(K_0 K^{-\frac{2}{5}}\right)
%     +\mathcal{O}(\epsilon_{\rm app}),
% \end{equation}
% where $\widetilde{\mathcal{O}}(\cdot)$ hides constants and the logarithmic order of $K$.
% \end{Corollary}

Different from i.i.d. sampling, the stochastic gradients $g(x,\omega)$ and $v(x,\theta,\omega)$ are biased for Markovian sampling, and the bias decreases as the chain mixes. 
% The workers need to obtain sufficient number of samples before the chain mixes. 
The mixing time corresponds to the logarithmic terms $\log K$ in \eqref{eq:theorem:async-tts-critic} and \eqref{eq:theorem:async-tts-actor}. 
Because of asynchrony, at a given iteration, workers collect different number of samples and their chains mix to different degrees. The worker with the slowest mixing chain will determine the rate of convergence. The product of $K_0$ and $\log K$ in \eqref{eq:theorem:async-tts-critic} and \eqref{eq:theorem:async-tts-actor} appears due to the slowest mixing chain. 
As the last term in \eqref{eq:theorem:async-tts-critic} dominates  other terms asymptotically, the convergence rate degrades as the number of workers increases. 
While the theoretical linear speedup is difficult to establish in the Markovian setting, we will empirically test it in Section \ref{section:nonlinear-experiment}.

\begin{Remark}[Challenges compared to AC analysis] Unlike synchronous AC, A3C introduces asynchrony and delay in both the actor and critic updates. At each iteration $k$, the delayed parameters will introduce extra error in $g(x,\omega_{k-\tau_k})-g(x,\omega_{k})$ and $v(x,\theta_{k-\tau_k},\omega_{k-\tau_k})-v(x,\theta_k,\omega_{k})$. Furthermore, it also causes delays in sampling since samples are drawn from the delayed policy $\pi_{\theta_{k-\tau_k}}$ instead of $\pi_{\theta_k}$. This delay will get amplified 
% when looking into the Markov chain of each worker, 
as every state on the Markov chain is generated by policies with different delays. At local counter $t$ ($t$th transition on local Markov chain), we compare the chain transition in synchronous and asynchronous settings:
% {\small\begin{align*}
%  {\rm\bf sync:}~~~   &s_{t-m} \xrightarrow{\pi_{\theta_{t-m}}} a_{t-m} \xrightarrow{\mathcal{P}} s_{t-m+1}\xrightarrow{\pi_{\theta_{t-m+1}}} a_{t-m+1}\\
%  {\rm\bf async:}~~~    &s_{t-m} \xrightarrow{\pi_{\theta_{k-d_m}}} a_{t-m} \xrightarrow{\mathcal{P}} s_{t-m+1}\xrightarrow{\pi_{\theta_{k-d_{m-1}}}} a_{t-m+1}
% \end{align*}}
% where $k$ is the global counter ($k$th update), $d_m$ is an unknown increasing sequence representing the difference between current counter $k$ and the counter of policy used to generate sample $a_{t-m}$. The parameter delay in each time step makes the Markov chain more difficult to analyze in asynchronous setting.
\begin{align*}
 &{\rm\bf sync:}~~s_{t} \xrightarrow{\theta_{t}} a_{t} \xrightarrow{\mathcal{P}} s_{t+1}\xrightarrow{\theta_{t+1}} a_{t+1}\cdots;\nonumber\\
 &{\rm\bf async:}~~   s_{t} \xrightarrow{\theta_{k-\tau_k}} a_{t} \xrightarrow{\mathcal{P}} s_{t+1}\xrightarrow{\theta_{k+d_t-\tau_{k+d_t}}} a_{t+1}\cdots
\end{align*}
where $k$ is the global counter at which the local Markov chain takes $t$th transition, $\tau_k$ is the delay of policy used to generate $t$th local transition, $d_t$ is the number of global updates between two local transitions. Clearly, the parameter delay makes the Markov chain more difficult to analyze.
\end{Remark}

\subsection{Global convergence under structured problem}
A3C is a gradient ascent type algorithm, thus can only achieve local convergence under a generally non-concave objective function $J_\lambda(\theta)$ w.r.t. $\theta$. However, under some special structured problem, A3C can be shown to achieve global convergence.
In this section, we consider the class of MDP which has finite state space and action space. 
Suppose the policy is parameterized by the softmax function:
\begin{align}
    \pi_\theta (a|s) = \frac{\exp(\theta_{s,a})}{\sum_{s,a}\exp(\theta_{s,a})}
\end{align}
where $\theta \in \mathbb{R}^{|\mathcal{S}||\mathcal{A}|}$ and $\theta_{s,a}$ is the policy parameter corresponds to pair $(s,a)$. The softmax policy class cannot represent deterministic policies with finite $\theta$. To avoid driving $\theta$ to infinity, it is crucial to penalize the deterministic policies with the regularization term introduced in \eqref{definition:jlambda}. To do so, we set the priors $\eta_p$ and $\pi_p$ as uniform distribution on state and action space, then the objective function can be rewritten as
\begin{align}
    J_\lambda (\theta) = J(\theta) + \frac{\lambda}{|\mathcal{S}||\mathcal{A}|}\sum_{s,a}\log \pi_\theta(a|s)+\lambda \log |\mathcal{A}|.
\end{align}
Define the state feature matrix $ \Phi' \coloneqq [\phi(s^{1}),\phi(s^{2}),\dots,\phi(s^{\mathcal{|S|}})]^\top \in \mathbb{R}^{|\mathcal{S}|\times d'}$ of which rows are features. We make the following assumption on $\Phi'$.
\begin{Assumption}\label{assumption:linear approximable}
  For any eligible $\theta$, there exists $\omega_\theta \in \mathbb{R}^{d'}$ such that $\Phi' \omega_\theta = V_{\pi_\theta}$.
\end{Assumption}
This assumption assumes that the value function $V_{\pi_\theta}$ can be accurately approximated by linear functions. For the assumption to hold, it suffices to select a squared full-rank feature matrix $\Phi'$. It is worth noting that when this assumption does not hold, our result in Theorem \ref{theorem:global} holds with an extra error term, which is the function approximation error $\epsilon_{\rm app}$.

\begin{figure*}[t]
    \hspace{-0.3cm}
    % \includegraphics[width=0.25\textwidth]{./images/critic-sample-workers-markov.pdf}
    %     \hspace{-0.1cm}
    \includegraphics[width=0.33\textwidth]{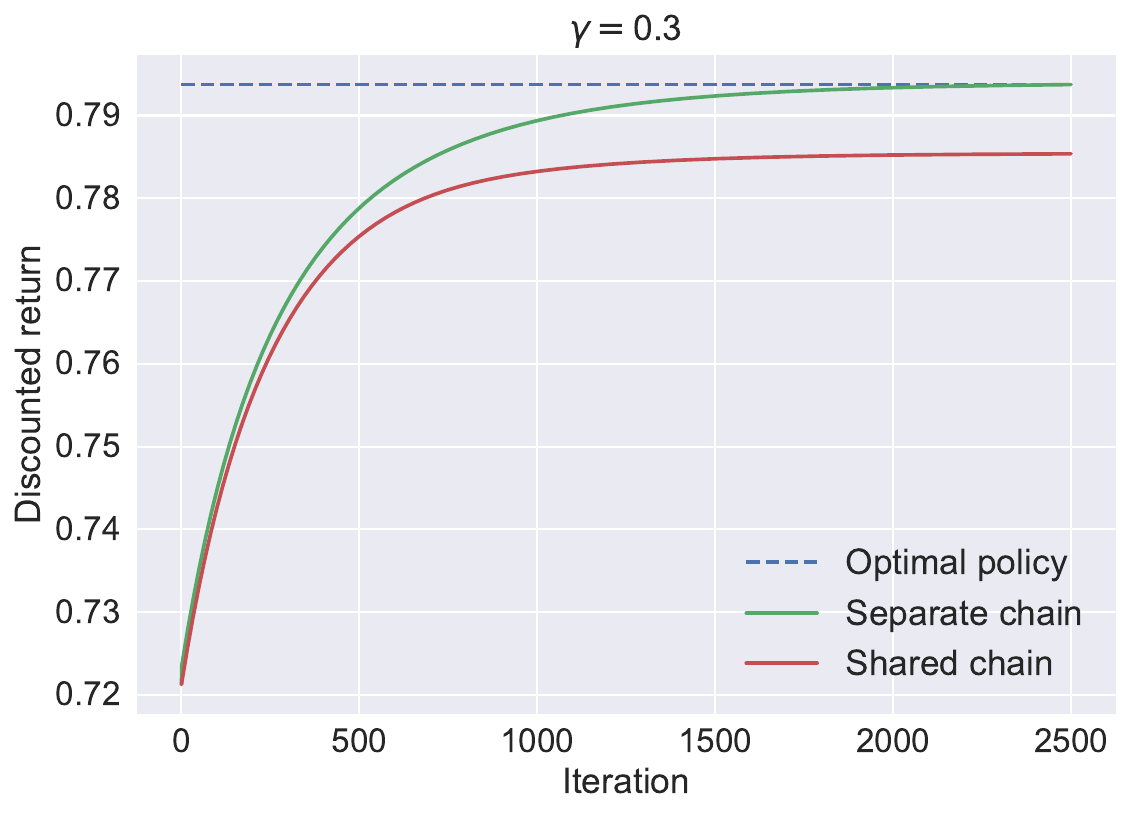}
        \hspace{-0.1cm}
    \includegraphics[width=0.33\textwidth]{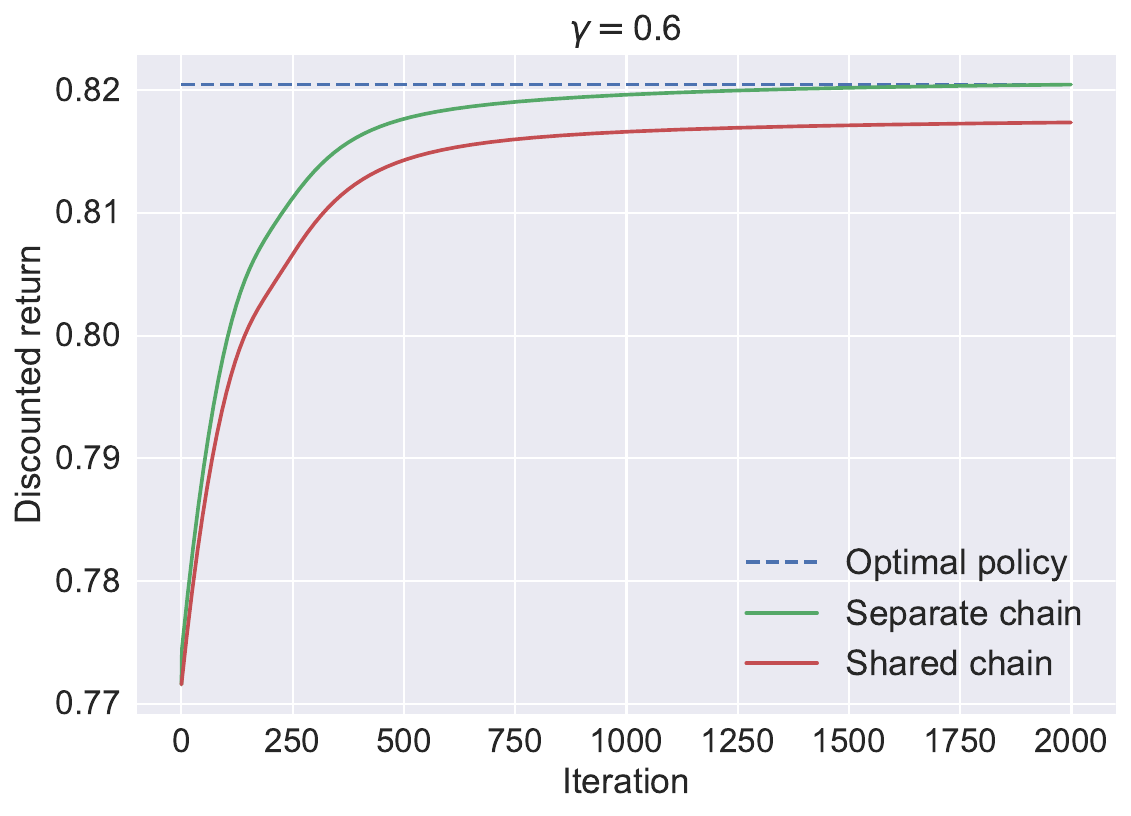}
        \hspace{-0.1cm}
    \includegraphics[width=0.33\textwidth]{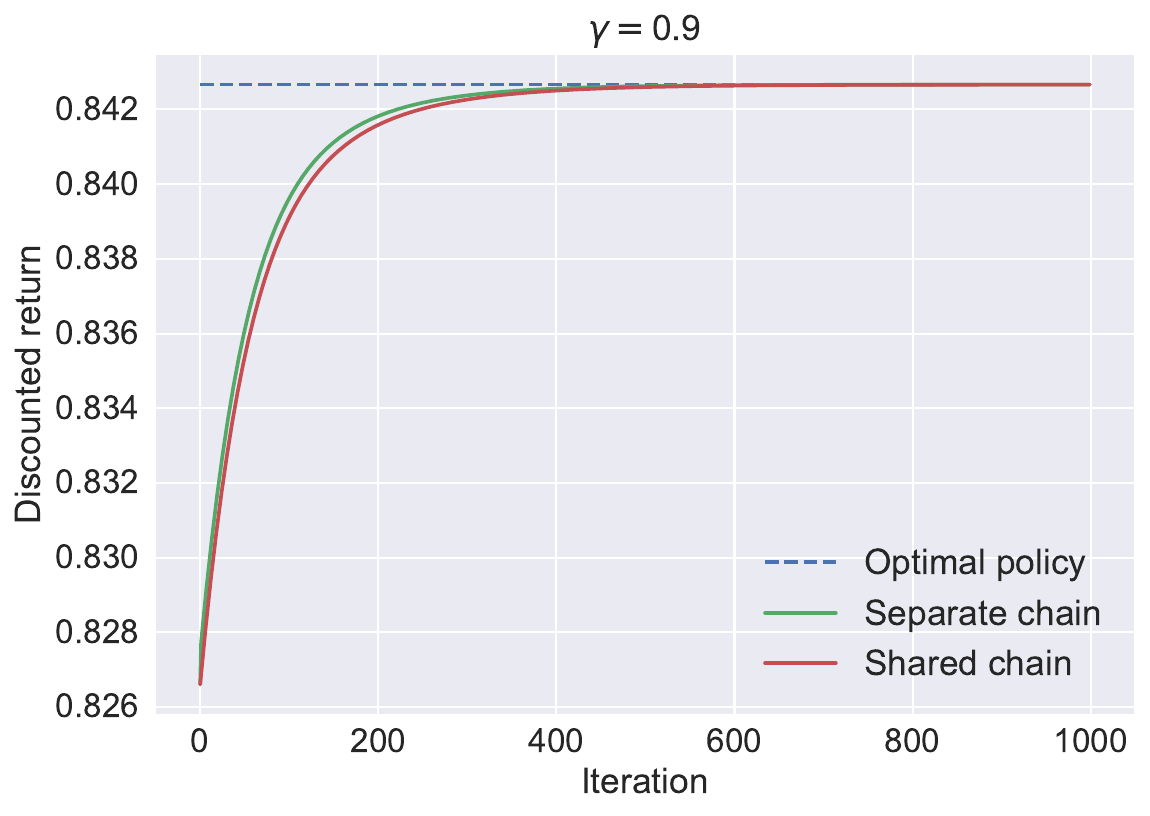}
        \hspace{-0.1cm}
    \vspace{-0.4cm}
    \caption{Algorithm \ref{algorithm:async-tts-worker} with separate chain sampling (option 2) vs shared chain sampling (setting $\hat{x}_t=x_t$ in the algorithm). The asymptotic error roughly scales propotionally to $1-\gamma$. With a smaller $\gamma$, the objective function $J$ becomes more shortsighted, and thus initial state distribution (restarting the chain) plays a more important role. If the actor shares the sample with critic, then a lack of chain restarting will introduce an unavoidable asymptotic error that grows larger as $\gamma$ becomes smaller. Separate chain sampling works thanks to the random restarting with a probability scaling with $\gamma$.}
    \label{figchain}
    \vspace*{-0.15cm}
\end{figure*}

 \begin{figure*}[t]
        % \vspace*{-0.2cm}
    \hspace{-0.3cm}
    % \includegraphics[width=0.25\textwidth]{./images/critic-sample-workers-iid.pdf}
    %     \hspace{-0.1cm}
    \includegraphics[width=0.248\textwidth]{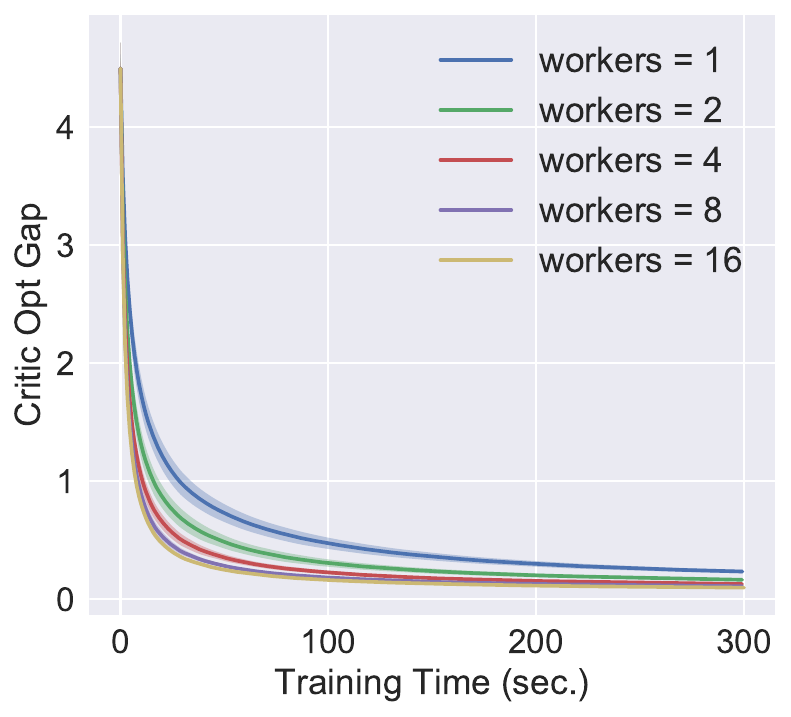}
        \hspace{-0.1cm}
    \includegraphics[width=0.254\textwidth]{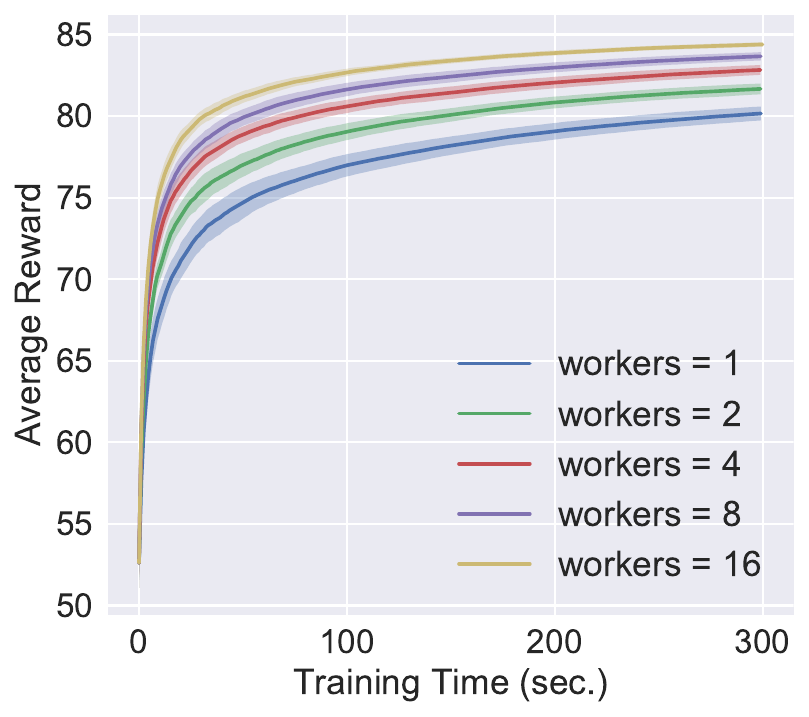}
        \hspace{-0.1cm}
    \includegraphics[width=0.254\textwidth]{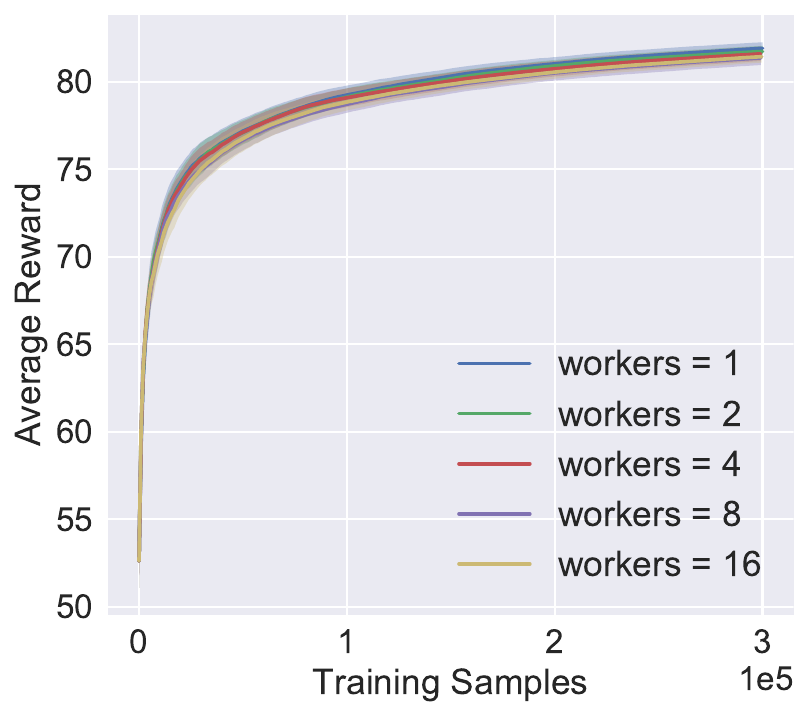}
        \hspace{-0.1cm}
    \includegraphics[width=0.244\textwidth]{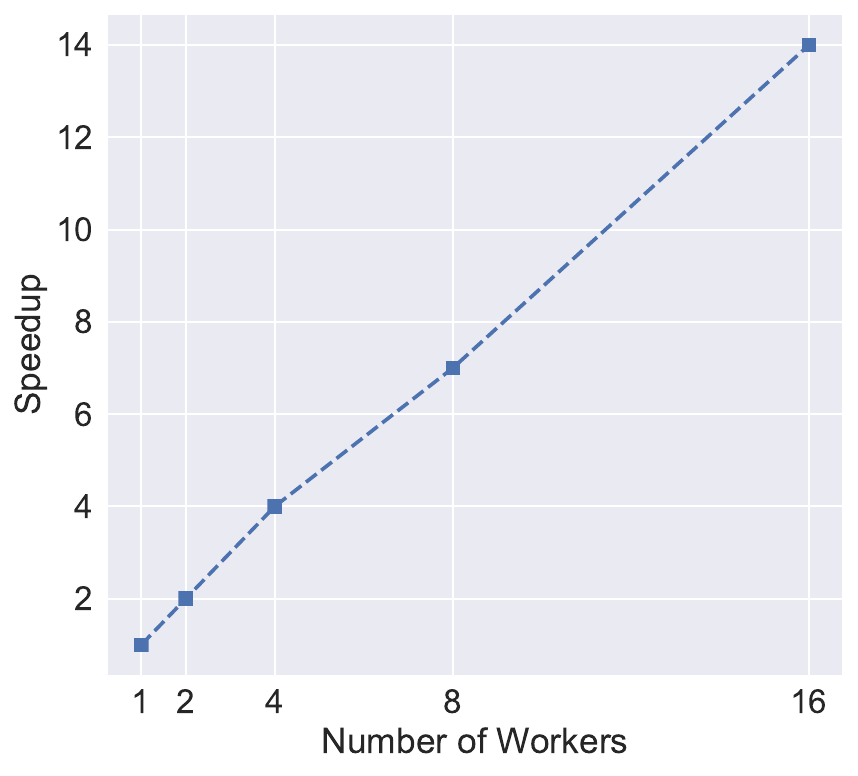}
    \vspace*{-0.6cm}
    \caption{Convergence results of A3C with i.i.d. sampling in synthetic environment.}
    \label{fig1}
        \vspace*{-0.0cm}
\end{figure*}

To establish global convergence, a gradient-dominance type condition was proven in \citep{agarwal2019optimality}:
\begin{Lemma}\label{lemma:gradientdominance}
With softmax policy parameterization and uniform priors, if $\|\nabla J_\lambda (\theta)\|_2 \leq \frac{\lambda}{2|\mathcal{S||\mathcal{A}|}}$, then $J^* - J(\theta) \leq  \epsilon_\lambda\coloneqq\frac{2\lambda}{1-\gamma} \big\|\frac{d_{\pi^*}}{\eta} \big\|_{\infty}$.
\end{Lemma}
For an arbitrary accuracy $\epsilon$, if we set $\lambda=\frac{(1-\gamma)\epsilon}{2\|\frac{d_{\pi^*}}{\eta} \|_{\infty}}$, then we have $\epsilon_\lambda=\epsilon$. Note that in order for $\|\frac{d_{\pi^*}}{\eta} \|_{\infty}$ to be finite, we need $\eta(s)\!>\!0$ for any $s\in\mathcal{S}$, which can be assumed without loss of generality. In the case where $\eta\!>\!0$ does not hold, one can start with an exploratory initial state distribution $\eta'\!>\!0$ like in \citep{agarwal2019optimality}, and our result still holds. This lemma allows us to establish connection between the gradient norm and optimality gap, giving rise to the following theorem.
\begin{Theorem}\label{theorem:global}
Suppose Assumptions \ref{assumption:A},\ref{assumption:delay} and \ref{assumption:MDP}--\ref{assumption:linear approximable} hold. Consider Algorithm \ref{algorithm:async-tts-worker} with softmax policy and linear critic function $\hat{V}_\omega(s)=\phi(s)^\top \omega$. Select step size $\alpha = K^{-\frac{3}{5}}$ ,$\beta = K^{-\frac{2}{5}}$ and let $K_0=\Theta(N)=\mathcal{O}({K}^{\frac{1}{5}})$, then it holds 
\begin{subequations}
for i.i.d. sampling
    \begin{equation}
     J^*-\frac{1}{K}\sum_{k=1}^{K}\E\big[J(\theta_k)\big]
    =
    \mathcal{O}\left(\lambda^{-2} K^{-\frac{2}{5}}\right)
    +\epsilon_\lambda,
    \end{equation}
    and for Markovian sampling
    \begin{equation}
     J^*-\frac{1}{K}\sum_{k=1}^{K}\E\big[J(\theta_k)\big]
    =
    \widetilde{\mathcal{O}}\left( \lambda^{-2} K_0 K^{-\frac{2}{5}}\right)+\epsilon_\lambda.
    \end{equation}
\end{subequations}
\end{Theorem}

\section{Numerical Experiments}\label{section:nonlinear-experiment}
We test the impact of separate sampling and the speedup property of A3C in both synthetically generated and Gym environments. 
The tests on synthetic environment were performed in a 16-core CPU computer, and those on Atari games were run in a 4 GPU computer. 

\vspace{-0.1cm}
\subsection{Separate sampling protocol}\label{section:separatemarkovchain}

\begin{figure*}[t]
    \hspace{-0.3cm}
    % \includegraphics[width=0.25\textwidth]{./images/critic-sample-workers-markov.pdf}
    %     \hspace{-0.1cm}
    \includegraphics[width=0.248\textwidth]{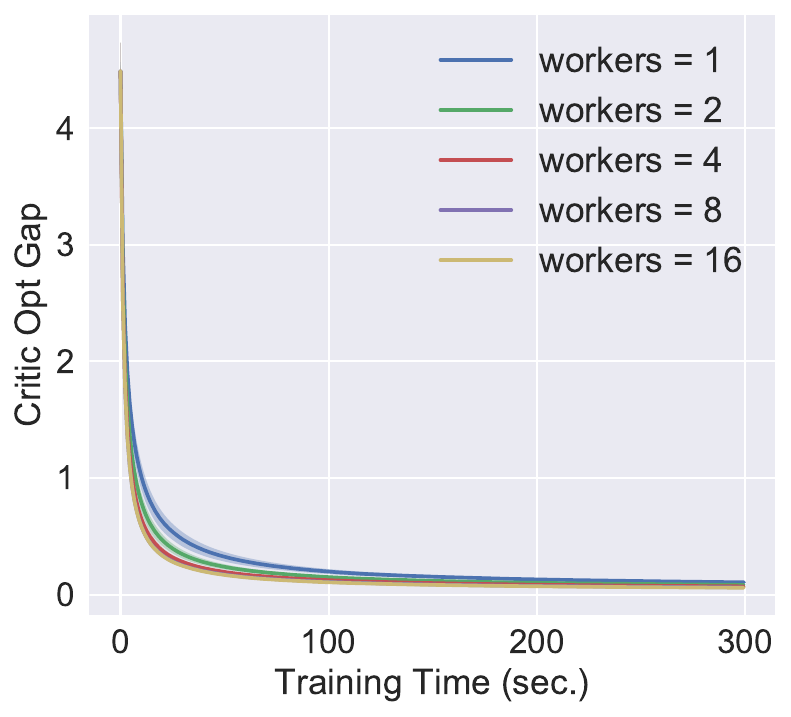}
        \hspace{-0.1cm}
    \includegraphics[width=0.254\textwidth]{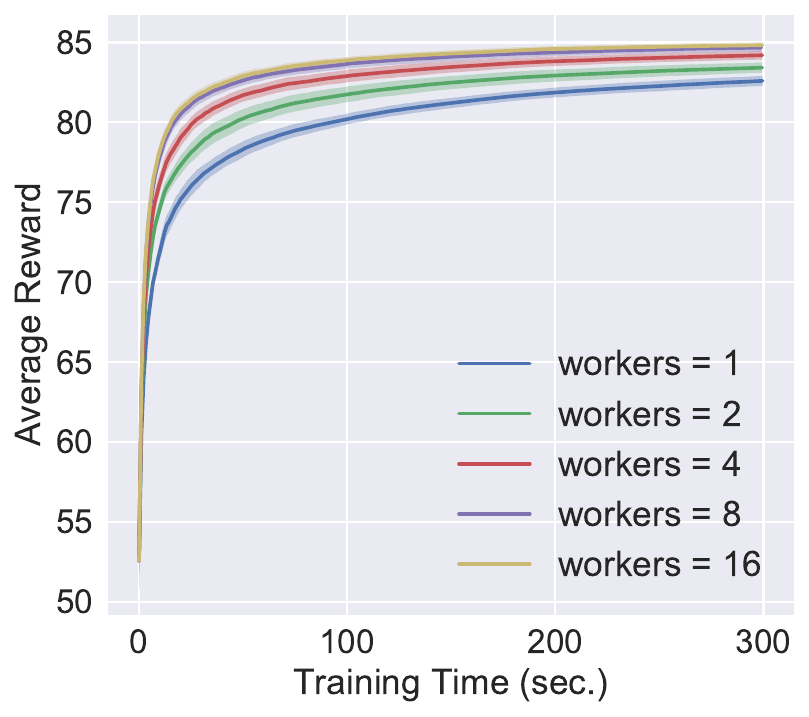}
        \hspace{-0.1cm}
    \includegraphics[width=0.254\textwidth]{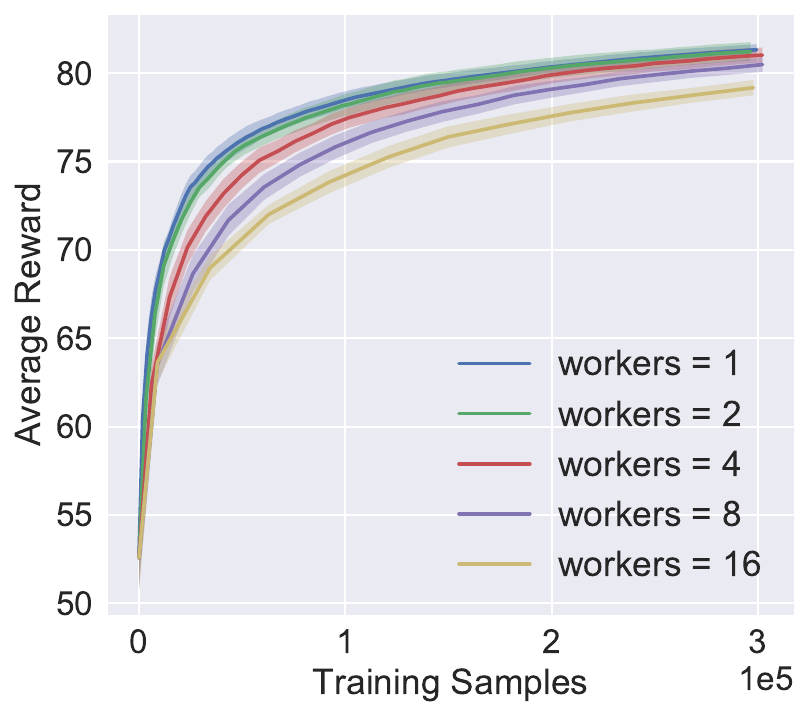}
        \hspace{-0.1cm}
    \includegraphics[width=0.244\textwidth]{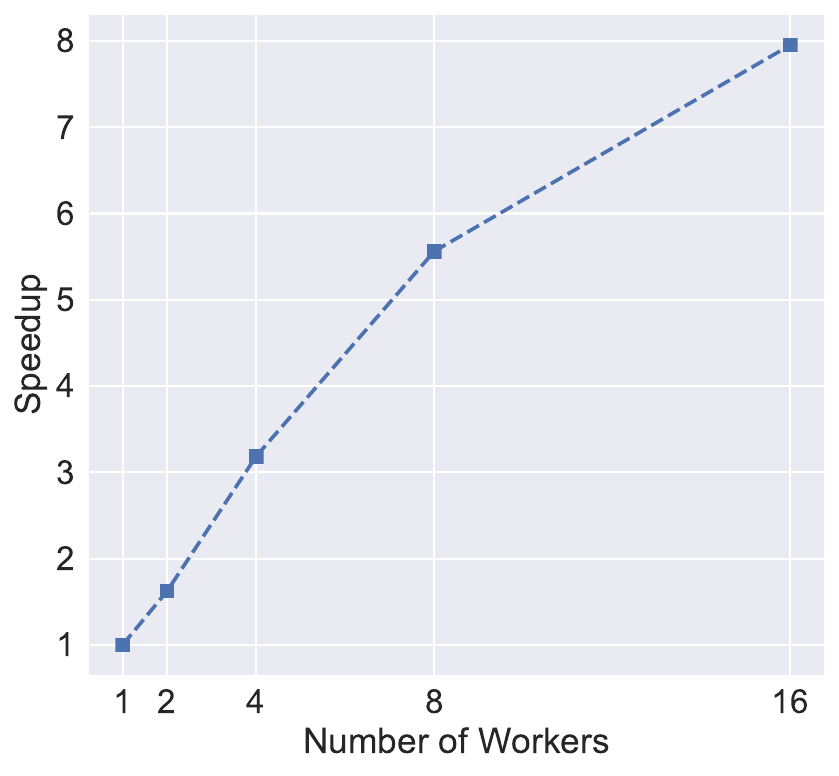}
    \vspace*{-0.6cm}
    \caption{Convergence results of A3C with Markovian sampling in synthetic environment.}
    \label{fig2}
    \vspace*{-0.1cm}
\end{figure*}

\begin{figure*}[t]
        % \vspace*{-0.4cm}
    \hspace{-0.3cm}
    \includegraphics[width=0.254\textwidth]{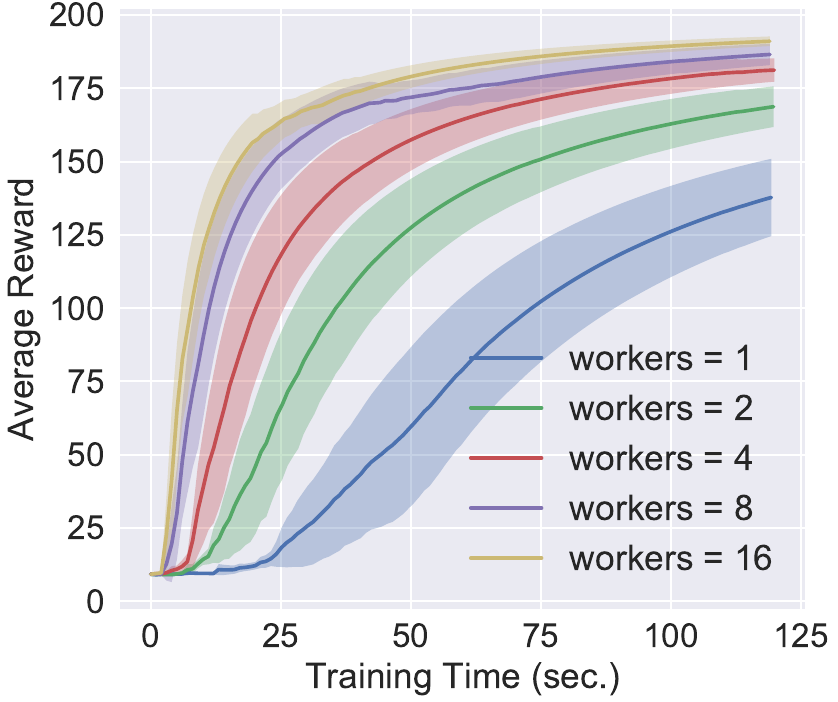}
      \hspace{-0.1cm}
    \includegraphics[width=0.25\textwidth]{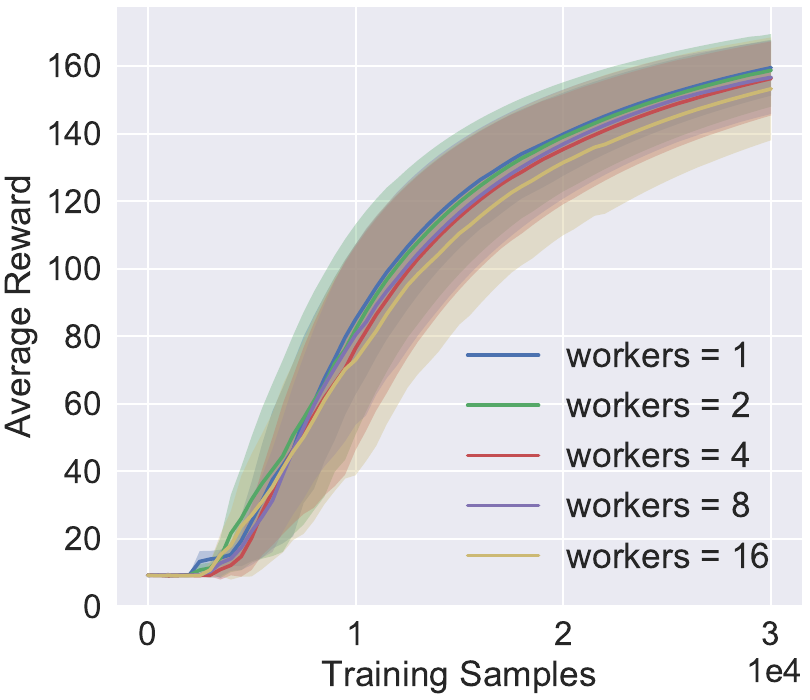}
      \hspace{-0.1cm}
    \includegraphics[width=0.249\textwidth]{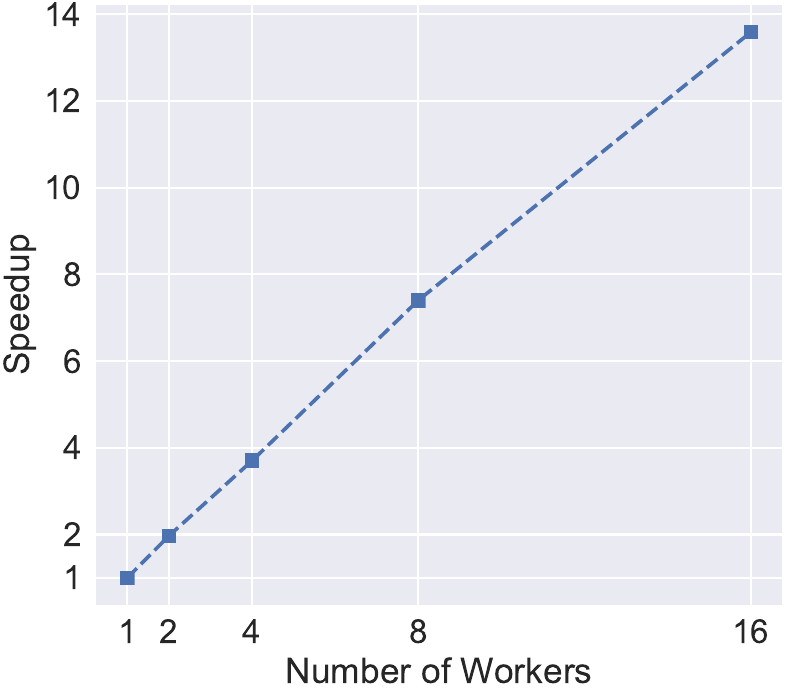}
      \hspace{-0.1cm}
        \includegraphics[width=0.248\textwidth]{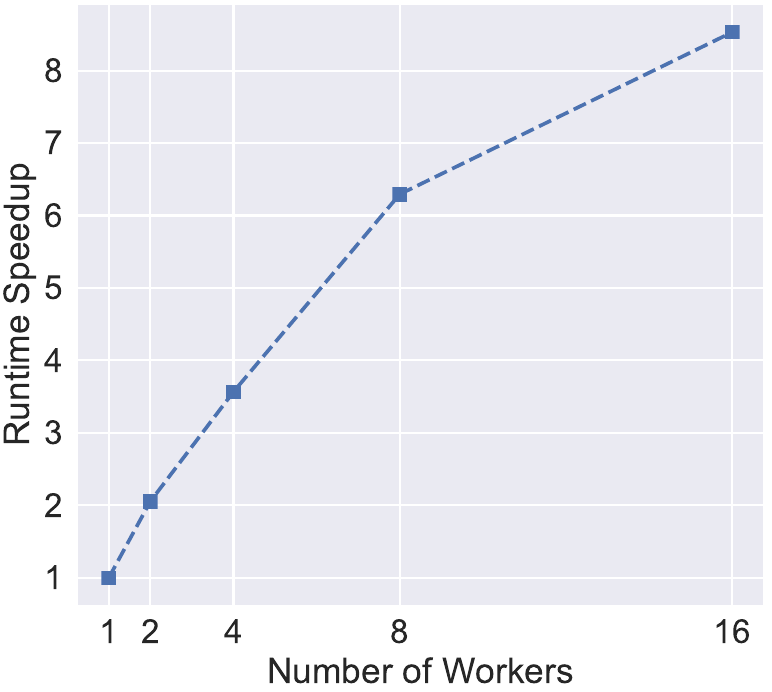}
    \vspace*{-0.6cm}
    \caption{Speedup of A3C in OpenAI gym classic control task (Carpole).}
    \label{fig3}
        \vspace*{-0.0cm}
\end{figure*}

% \begin{figure*}[t]
%         \vspace*{-0.2cm}
%     \hspace{-0.3cm}
%     \includegraphics[width=0.25\textwidth]{./images/rew-time-workers-SeaquestNoFrameskip-v4.pdf}
%       \hspace{-0.1cm}
%     \includegraphics[width=0.25\textwidth]{./images/rew-sample-workers-SeaquestNoFrameskip-v4.pdf}
%       \hspace{-0.1cm}
%     \includegraphics[width=0.245\textwidth]{./images/rew-speedup-sample-workers-SeaquestNoFrameskip-v4.pdf}
%       \hspace{-0.1cm}
%     \includegraphics[width=0.245\textwidth]{./images/rew-speedup-time-workers-SeaquestNoFrameskip-v4.pdf}
%     \vspace*{-0.2cm}
%     \caption{Speedup of A3C in OpenAI Gym Atari game (Seaquest).}
%     \label{fig4}
%         \vspace*{-0.2cm}
% \end{figure*}

\begin{figure*}[t]
        \vspace*{-0.2cm}
    \hspace{-0.3cm}
    \includegraphics[width=0.25\textwidth]{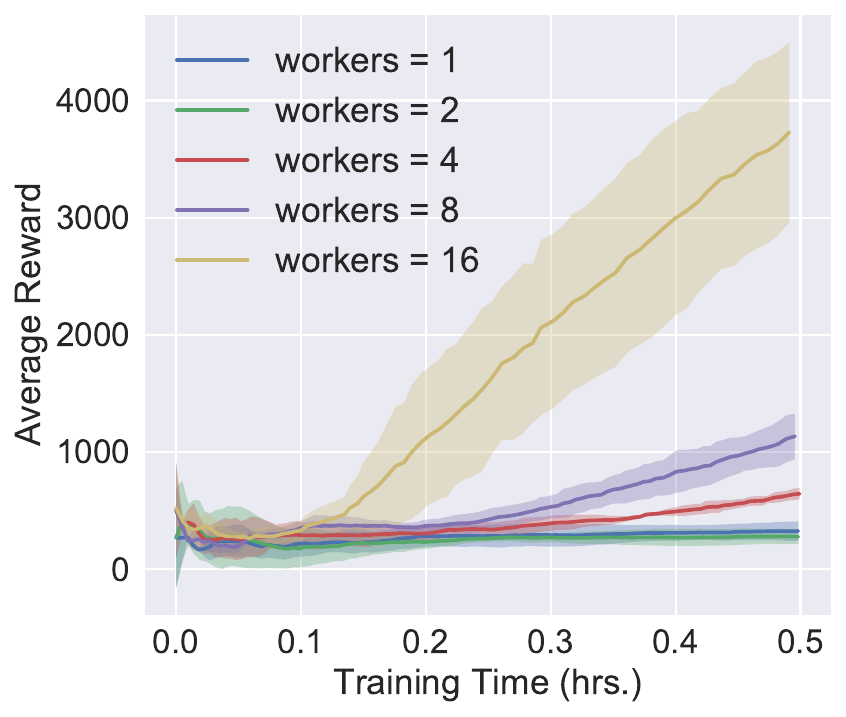}
      \hspace{-0.1cm}
    \includegraphics[width=0.25\textwidth]{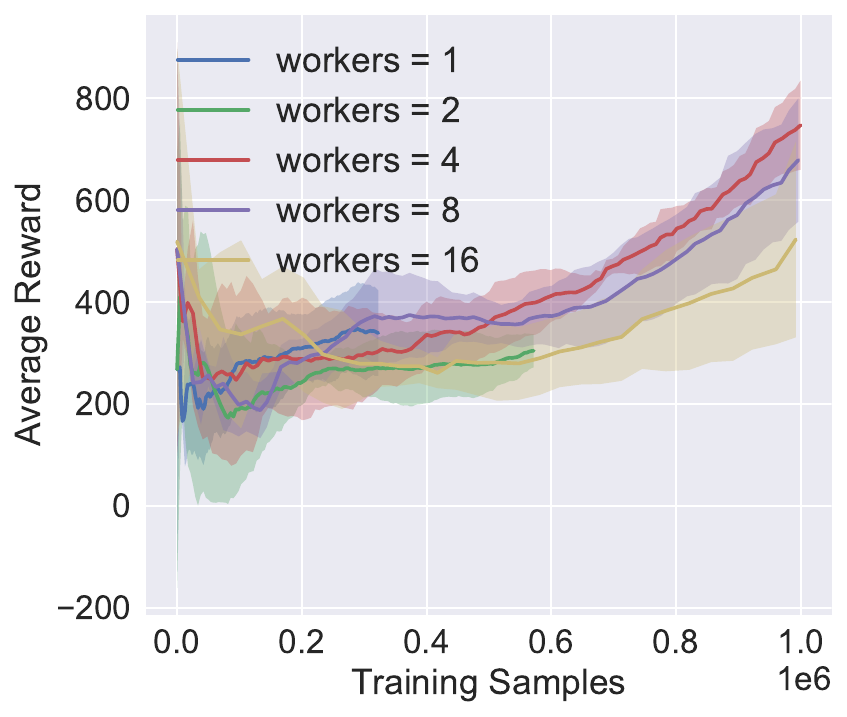}
      \hspace{-0.1cm}
    \includegraphics[width=0.245\textwidth]{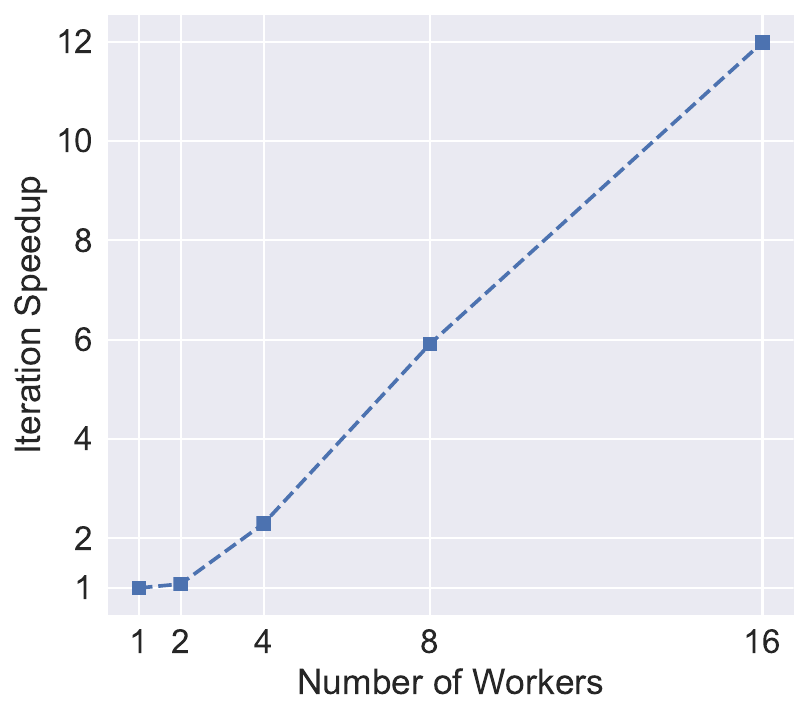}
      \hspace{-0.1cm}
    \includegraphics[width=0.245\textwidth]{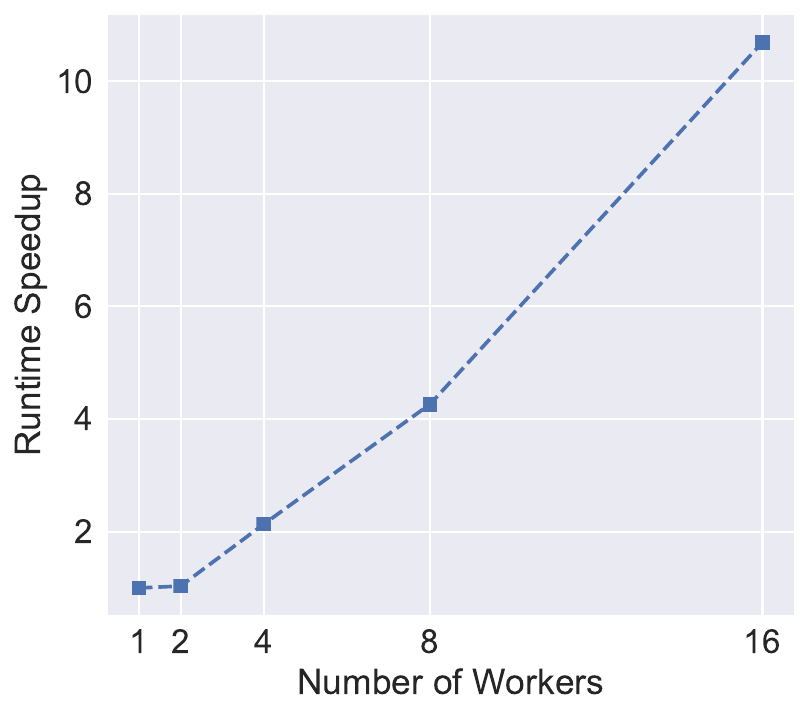}
    \vspace*{-0.2cm}
    \caption{Speedup of A3C in OpenAI Gym Atari game (Beamrider).}
    \label{fig6}
        \vspace*{-0.2cm}
\end{figure*}

\begin{figure*}[t]
        % \vspace*{-0.2cm}
    \hspace{-0.3cm}
    \includegraphics[width=0.25\textwidth]{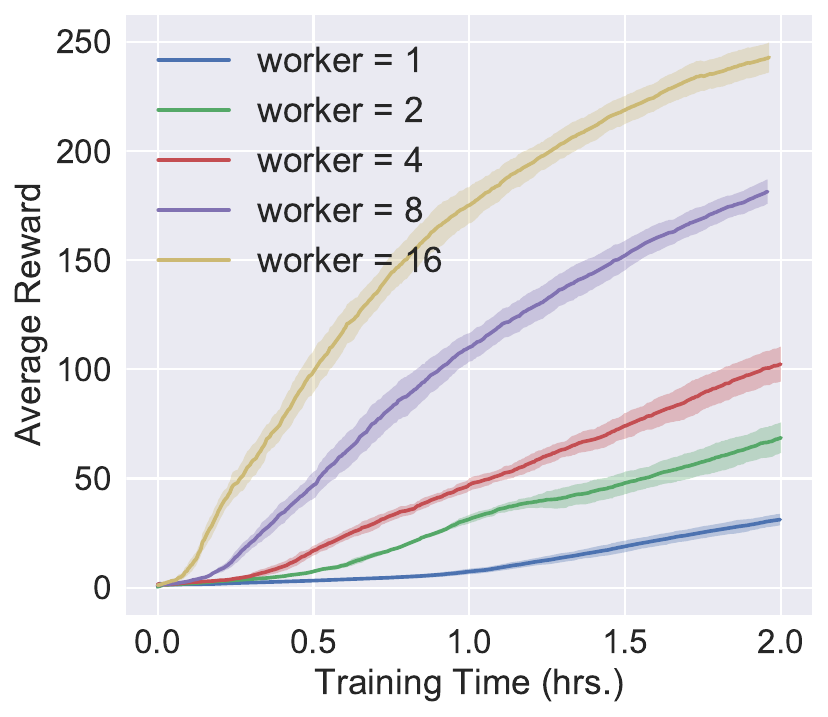}
      \hspace{-0.1cm}
    \includegraphics[width=0.25\textwidth]{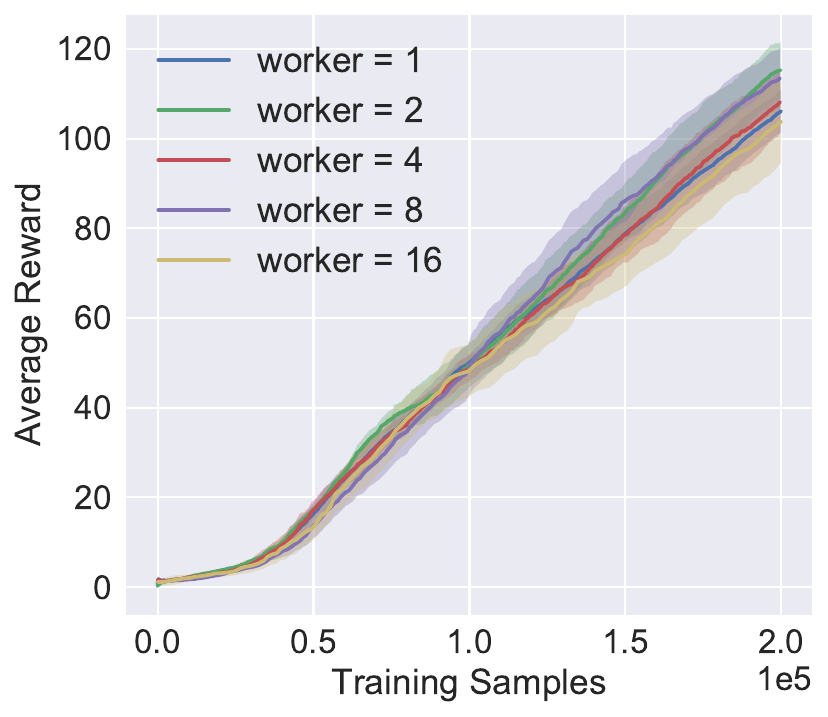}
      \hspace{-0.1cm}
    \includegraphics[width=0.25\textwidth]{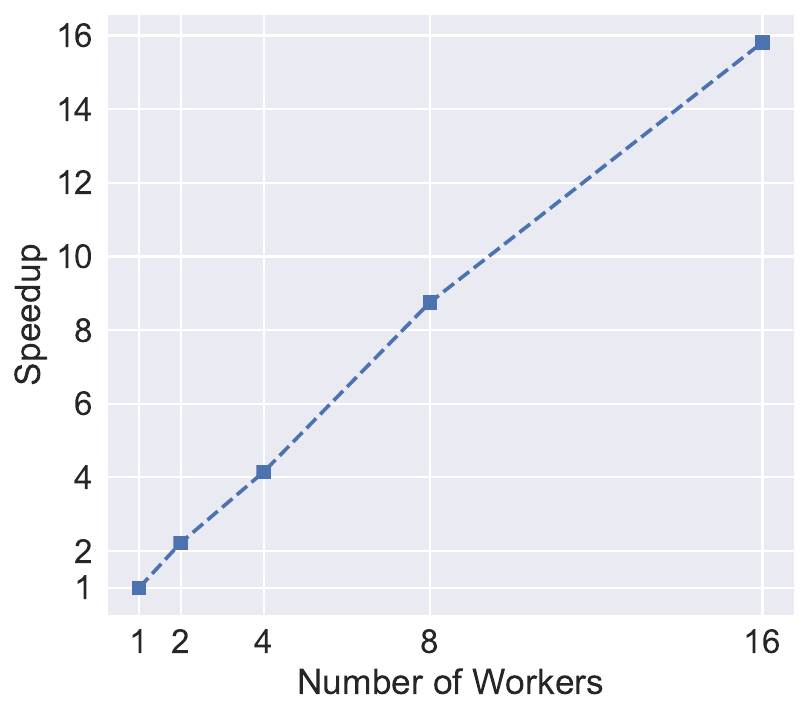}
      \hspace{-0.1cm}
    \includegraphics[width=0.25\textwidth]{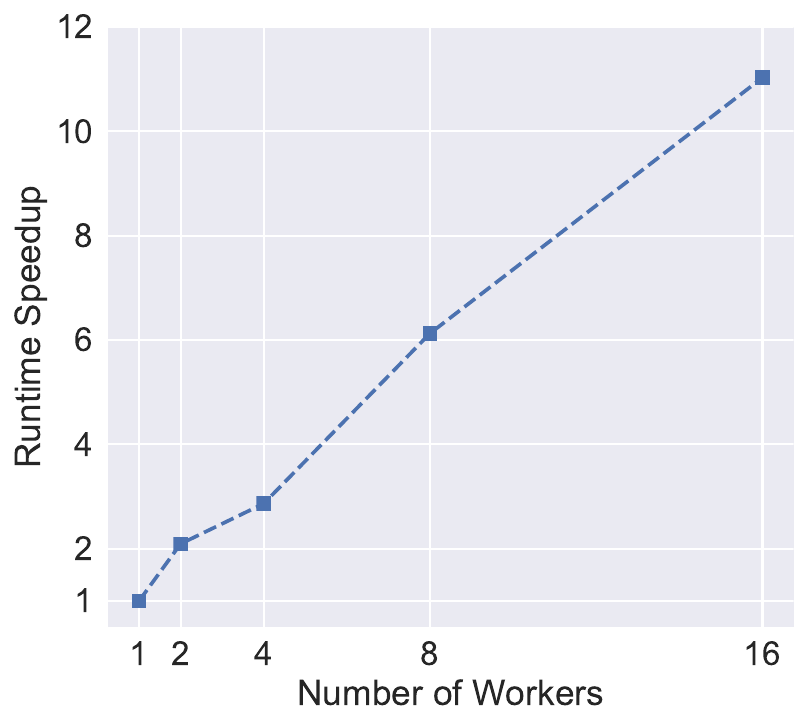}
    \vspace*{-0.6cm}
    \caption{Speedup of A3C in OpenAI Gym Atari game (Breakout).}
    \label{fig5}
        \vspace*{-0.0cm}
\end{figure*}

\begin{figure*}[t]
        % \vspace*{-0.2cm}
    \hspace{-0.3cm}
    \includegraphics[width=0.25\textwidth]{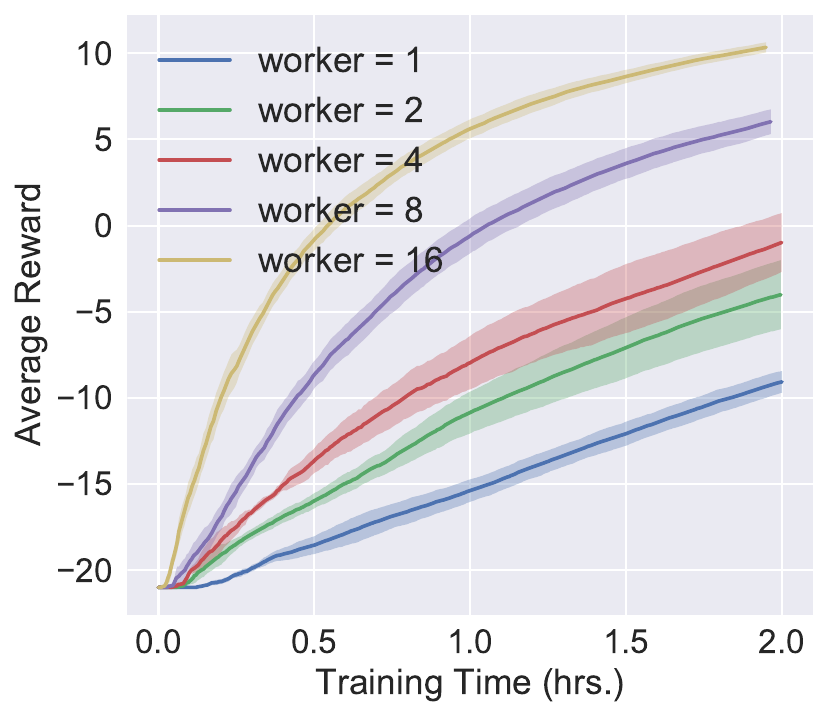}
      \hspace{-0.1cm}
    \includegraphics[width=0.25\textwidth]{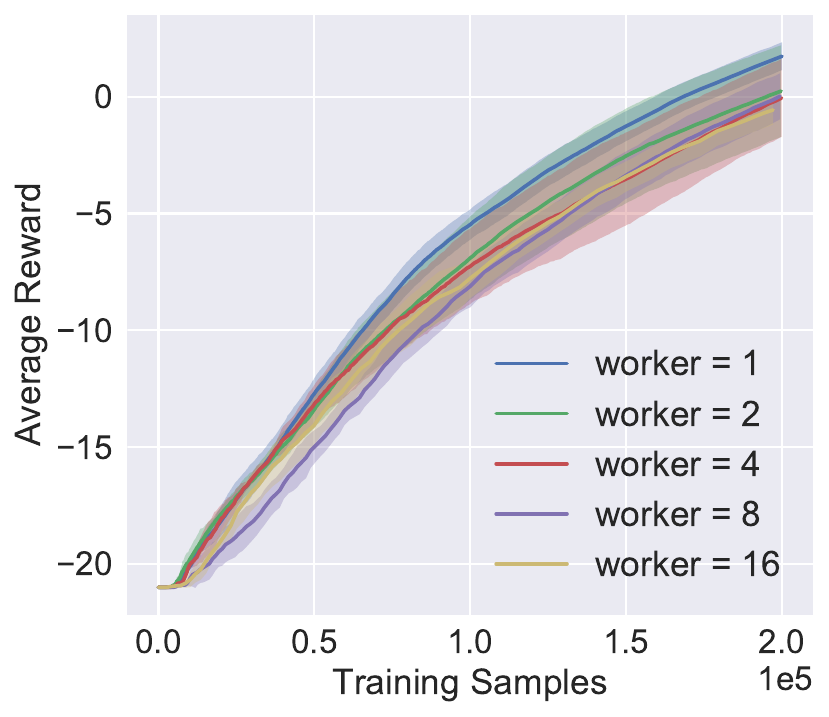}
      \hspace{-0.1cm}
    \includegraphics[width=0.25\textwidth]{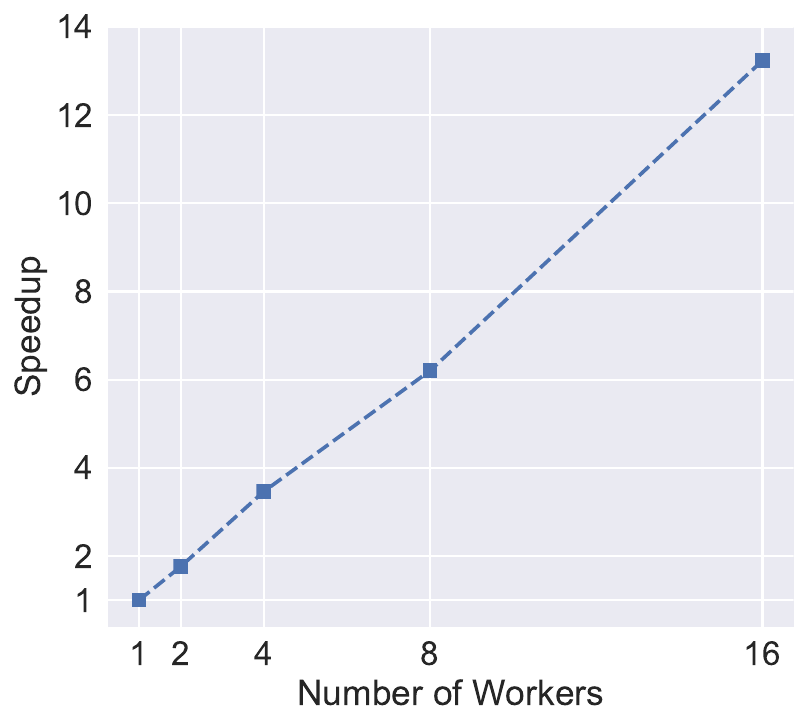}
      \hspace{-0.1cm}
    \includegraphics[width=0.25\textwidth]{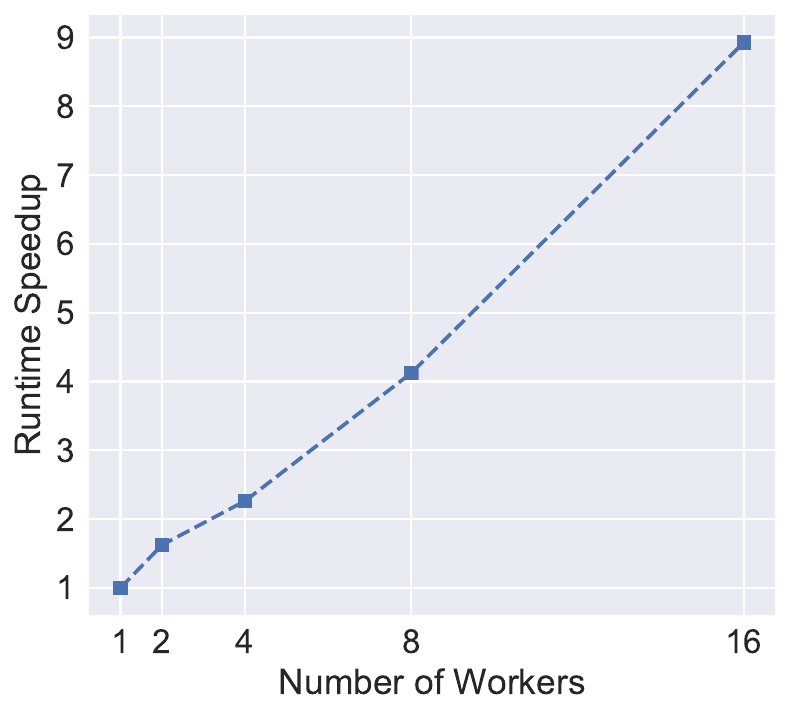}
    \vspace*{-0.6cm}
    \caption{Speedup of A3C in OpenAI Gym Atari game (Pong).}
    \label{fig7}
\end{figure*}

We compare the separate chain sampling method in Algorithm \ref{algorithm:async-tts-worker} with the shared chain sampling method. The shared chain method is simply using the same sample for both actor and critic, i.e., setting $\hat{x}_t=x_t$ in Algorithm \ref{algorithm:async-tts-worker}.

% \textbf{Synthetic environment.}
To clearly demonstrate the impact of sampling, we mitigate the impact from other sources such as delay and MDP non-ergodicity by considering a synthetic environment with $1$ worker. In this test, we use the tabular softmax policy parameterization. The synthetic MDP has a state space $|\mathcal{S}|=10$, an discrete action space of $|\mathcal{A}|=4$. State features each has a dimension of 10. Elements of the transition matrix, the reward and the state features are randomly sampled from a uniform distribution over $(0,1)$.

It can be clearly observed from Figure \ref{figchain} that using the same sample for actor and critic leads to an asymptotic error scaled with choice of $\gamma$. The intuitive explanation is in the caption of Figure \ref{figchain}. Although when $\gamma \xrightarrow[]{}1$, the error is small (but still exists), we want our algorithm design to not restrict the choice of $\gamma$, and thus adopt the separate chain sampling method.

\vspace{-0.1cm}
\subsection{Linear speedup}

\begin{table}[t]
\begin{center}
\begin{tabular}{ |c|c| }
 \hline
 Hyper-parameters & Value \\
 \hline
 Number of workers & 1,2,4,8,16 \\
 Optimizer & Adam \\
 Step size & 0.00015 \\
 Batch size & 20 \\
 Discount factor & 0.99 \\
 Entropy coefficient & 0.01 \\
 Frame size & 80 $\times$ 80 \\
 Frame skip rate & 4\\ 
 Grayscaling & Yes \\
 Training reward clipping & [-1,1] \\
%  End of episode when life lost & Yes \\
%  Max number of no-ops after environment reset & 30 \\ 
 \hline
\end{tabular}
\end{center}
\caption{\label{tab:hyperparameters}Hyper-parameters of A3C in the Atari games.}
\vspace{-0.0cm}
\end{table}

% \textbf{Hardware device.} The tests on synthetic environment and CartPole was performed in a 16-core CPU computer. The tests on Atari games was run in a 4 GPU computer. 

\textbf{Experiment settings.}
For the synthetic environment, we used linear value function approximation and tabular softmax policy \citep{agarwal2019optimality}. For CartPole, we used a 3-layer MLP with 128 neurons and sigmoid activation function in each layer. The first two layers are shared for both actor and critic network. For the Atari games, we used a convolution-LSTM network. For network details, see \citep{a3g}.

% \textbf{Hyper-parameters.} 
For the separate sampling protocol test, we have $\alpha = 0.6$ and critic step size $\beta = 0.7$, along with $\lambda=0.3$. For the speedup tests in synthetic envrionment, we set actor step size $\alpha_k = \frac{0.05}{(1+k)^{0.6}}$ and critic step size $\beta_k = \frac{0.05}{(1+k)^{0.4}}$. In tests of CartPole, we run Algorithm \ref{algorithm:async-tts-worker} with a minibatch of 20 samples. We update the actor network with a step size of $\alpha_k = \frac{0.01}{(1+k)^{0.6}}$ and critic network with a step size of $\beta_k = \frac{0.01}{(1+k)^{0.4}}$. See Table \ref{tab:hyperparameters} for hyper-parameters in Atari game tests.

\textbf{Synthetic environment.}
We first test the speedup property of A3C in a synthetic environment with $|\mathcal{S}|=100$, $|\mathcal{A}|=5$ and state feature with dimension $10$. The reward and transition matrix of the MDP are randomly generated in the same way as that in section \ref{section:separatemarkovchain}.
% constructed in the same way as that in section \ref{section:separatemarkovchain}
We evaluate the convergence of actor in terms of the average reward and the critic in terms of the gap $\|\omega_k - \omega_{\theta_k}^*\|_2$. 
% The critic optimality gap is given by $\|\omega - \omega_\theta^*\|_2$, where $\omega_\theta^*$ is computed with $-A_{\theta, \phi}^{-1}b_{\theta, \phi}$. Test reward is computed by sampling a trajectory with the current policy and then summing the rewards. 

% Select initial step size $c_1=c_2=0.05$, and time scales $\sigma_1=0.6$, $\sigma_2=0.4$. 
Figures \ref{fig1} and \ref{fig2} show the training time and sample complexity of running A3C with i.i.d. sampling and Markovian sampling respectively.
The speedup plot is measured by the number of samples needed to achieve a target running average reward under different number of workers.
All the results are average over 10 Monte-Carlo runs.
Figure \ref{fig1} shows that the sample complexity of A3C stays the same with different number of workers under i.i.d. sampling. Also, it can be observed from the speedup plot of Figure \ref{fig1} that the A3C achieves roughly linear speedup, which is consistent with Corollary \ref{corollary:linear-speedup}.
The speedup of A3C with Markovian sampling shown in Figure \ref{fig2} is roughly linear when number of workers is small.
% {\red[is the speedup computed by using (1)? or by directly computing the training time?]} 
% {\red[only have one algorithm, change ``Algorithm 1" to A3C?]}
 
\textbf{OpenAI Gym environments.}
We also test the speedup property of A3C with neural network parametrization in the classic control (Carpole) and the Atari (Breakout and Pong) environments.  
In Figures \ref{fig3}-\ref{fig7}, each curve was averaged over 5 Monte-Carlo runs with $95\%$ confidence interval. 
% {\red[mention specifics is in the end of supplementary material?]}
Figures \ref{fig3}--\ref{fig7} show the speedup of A3C under different number of workers, where the average reward is computed by taking the running average of test rewards. 
The speedup is respectively measured by the number of samples and training time needed per worker to achieve a target average reward. 
Although not justified theoretically, Figures \ref{fig3}--\ref{fig7} suggest that the sample complexity speedup is roughly linear, and the runtime speedup slightly degrades when the number of workers increases. This is partially due to hardware limit. Similar observation was obtained in async-SGD \cite[Fig. 4]{lian2016speedup}.

\section{Conclusions}
This paper revisits the A3C algorithm. 
With linear value function approximation, the convergence of the A3C algorithm has been established under both i.i.d. and Markovian sampling settings. Under i.i.d. sampling, A3C achieves linear speedup compared to the best-known sample complexity of AC, theoretically justifying the benefit of parallelism and asynchrony for the first time. Under Markov sampling, such a linear speedup can be observed in most benchmark tasks. 
One limitation of this paper is that theoretical linear speedup cannot be established in the Markovian setting. This motivates two interesting directions: i) developing new tools of analyzing two-timescale SGD with Markov sampling; and, ii) designing better algorithms than A3C to achieve better speedup.

\vspace{-0.2cm}
\bibliography{abrv, async-ac}

\newpage
\appendix
\begin{center}
	\Large \textbf{Supplementary Material} \\
\end{center}

\section{Preliminary Lemmas}
\subsection{Geometric mixing}
The operation $p \otimes q$ denotes the product between two distributions $p(x)$ and $q(y)$, i.e. $(p \otimes q)(x,y) = p(x) \cdot q(y)$.
\begin{Lemma}\label{lemma:geom-s-a-s'}
Suppose Assumption \ref{assumption:MDP} holds. For any $\theta \in \mathbb{R}^{d}$, we have
\begin{subequations}
    \begin{equation}
        \label{eq:geom-s-a-s'}
    \sup_{s_0 \in \mathcal{S}} d_{TV}\left(\mathbb{P}((s_t,a_t,s_{t+1})\in\cdot | s_0, \pi_\theta) , \mu_{\theta} \otimes \pi_\theta \otimes \mathcal{P} \right) \leq \kappa \rho^t.
    \end{equation}
    and
    \begin{equation}
        \label{eq:geom-s-a-s'hat}
    \sup_{s_0 \in \mathcal{S}} d_{TV}\left(\mathbb{P}((\hat{s}_t,\hat{a}_t,s'_{t+1})\in\cdot | s_0, \pi_\theta) , d_{\theta} \otimes \pi_\theta \otimes \mathcal{P} \right) \leq \kappa \rho^t.
    \end{equation}
\end{subequations}
where $(s_t,a_t,s_{t+1})$ is the $t$th transition on the Makov chain with transition kernel $\mathcal{P}$. $(\hat{s}_t,\hat{a}_t)$ is the $t$th state-action pair on Markov chain with transition kernel $\hat{\mathcal{P}}$, and $s'_{t+1}\sim \mathcal{P}(\cdot | \hat{s}_t,\hat{a}_t)$.
\end{Lemma}
\begin{proof}
We start with
\begin{align*}
    &\sup_{s_0 \in \mathcal{S}} d_{TV}\left(\mathbb{P}((s_t,a_t,s_{t+1})=\cdot | s_0, \pi_\theta) , \mu_{\theta} \otimes \pi_\theta \otimes \mathcal{P} \right) \\
    &= \sup_{s_0 \in \mathcal{S}} d_{TV}\left(\mathbb{P}(s_t=\cdot | s_0, \pi_\theta) \otimes \pi_\theta \otimes  \mathcal{P}, \mu_\theta \otimes \pi_\theta \otimes  \mathcal{P} \right) \\
    &= \sup_{s_0 \in \mathcal{S}} \frac{1}{2} \int_{s \in \mathcal{S}}\sum_{a \in \mathcal{A}} \int_{s' \in \mathcal{S}} \left.| \mathbb{P}(s_t=ds|s_0, \pi_\theta)\pi_\theta(a|s)\mathcal{P}(ds'|s,a)- \mu_\theta(ds)\pi_\theta(a|s)\mathcal{P}(ds'|s,a) \right|\\
    &= \sup_{s_0 \in \mathcal{S}} \frac{1}{2} \int_{s \in \mathcal{S}}\left.| \mathbb{P}(s_t=ds|s_0, \pi_\theta)- \mu_\theta(ds) \right|\sum_{a \in \mathcal{A}}\pi_\theta(a|s)\int_{s' \in \mathcal{S}}\mathcal{P}(ds'|s,a) \\
    &= \sup_{s_0 \in \mathcal{S}} d_{TV}\left(\mathbb{P}(s_t\in\cdot | s_0, \pi_\theta) , \mu_\theta \right) \\
    &\leq \kappa \rho^t, \numberthis
\end{align*}
Inequality \eqref{eq:geom-s-a-s'} along with the fact that the stationary distribution of the Markov chain with transition probability $\hat{\mathcal{P}}$ and policy $\pi_\theta$ is simply $d_\theta$ immediately implies \eqref{eq:geom-s-a-s'hat}.
This completes the proof.
\end{proof}

For the use in the later proof, given $K > 0$, we first define $m_K$ as:
\begin{align*}\label{eq:m_K-definition}
    m_K \coloneqq \min\left\{ \hspace{0.06cm} m \in \mathbb{N}^+ \hspace{0.06cm}| \hspace{0.06cm} \kappa \rho^{m-1} \leq \min\{\alpha,\beta\}\right\}, \numberthis
\end{align*}
where $\kappa$ and $\rho$ are constants defined in \eqref{assumption:MDP}. $m_K$ is the minimum number of samples needed for the Markov chain to approach the stationary distribution so that the bias incurred by the Markovian sampling is small enough. 

\subsection{Auxiliary Markov chain}
The auxiliary Markov chain is a virtual Markov chain with no policy drifting --- a technique developed in \citep{zou2019sarsa} to analyze stochastic approximation algorithms in non-stationary settings. We provide an analysis here for completeness.
\begin{Lemma}\label{lemma:auxiliary-chain}
Under Assumption \ref{assumption:delay} and Assumption \ref{assumption:omega}, consider the update \eqref{algorithm:async-tts-update} in Algorithm \ref{algorithm:async-tts-worker} with Markovian sampling. For a given number of samples $m$, consider the Markov chain of the worker that contributes to the $k$th update:
\begin{align*}
    s_{t-m} \xrightarrow{\theta_{k-d_m}} a_{t-m} \xrightarrow{\mathcal{P}} s_{t-m+1} \xrightarrow{\theta_{k-d_{m-1}}} a_{t-m+1} \cdots s_{t-1} \xrightarrow{\theta_{k-d_1}} a_{t-1}  \xrightarrow{\mathcal{P}} s_{t} \xrightarrow{\theta_{k-d_0}} a_{t} \xrightarrow{\mathcal{P}} s_{t+1},
\end{align*}
where $(s_t,a_t,s_{t+1})=(s_{(k)},a_{(k)},s'_{(k)})$, and $\{d_j\}_{j=0}^{m}$ is some increasing sequence with $d_0 \coloneqq \tau_k$.

Given $(s_{t-m}, a_{t-m},s_{t-m+1})$ and $\theta_{k-d_m}$, we construct its auxiliary Markov chain by repeatedly applying $\pi_{\theta_{k-d_m}}$:
\begin{align*}
    s_{t-m} \xrightarrow{\theta_{k-d_m}} a_{t-m} \xrightarrow{\mathcal{P}} s_{t-m+1} \xrightarrow{\theta_{k-d_m}} \widetilde{a}_{t-m+1} \cdots \widetilde{s}_{t-1} \xrightarrow{\theta_{k-d_m}} \widetilde{a}_{t-1}\xrightarrow{\mathcal{P}} \widetilde{s}_t \xrightarrow{\theta_{k-d_m}} \widetilde{a}_{t} \xrightarrow{\mathcal{P}} \widetilde{s}_{t+1}.
\end{align*}
Then we have:
\begin{align*}    \label{eq:auxiliary-markov-chain-4}
    &d_{TV}\left(\mathbb{P}\big((s_t,a_t)\in \cdot | \theta_{k-d_m},s_{t-m+1}\big),\mathbb{P}\big((\widetilde{s}_t,\widetilde{a}_t) \in \cdot | \theta_{k-d_m},s_{t-m+1}\big)\right) \\
    &\leq \frac{1}{2}|\mathcal{A}|L_\pi\sum_{i=\tau_k}^{d_m}\E\left[\|\theta_{k-i}-\theta_{k-d_m}\|_2 | \theta_{k-d_m},s_{t-m+1}\right]. \numberthis
\end{align*}
\end{Lemma}
\begin{proof}
% Equations (\ref{eq:auxiliary-markov-chain-1}) - (\ref{eq:auxiliary-markov-chain-3}) have been shown in Lemma A.2. of \citep{wu2020finite}. For (\ref{eq:auxiliary-markov-chain-4}), we have
Throughout the lemma, all expectations and probabilities are conditioned on $\theta_{k-d_m}$ and $s_{t-m+1}$. We omit this condition for convenience. 

With $\widetilde{x}_t \coloneqq (\widetilde{s}_t,\widetilde{a}_t,\widetilde{s}_{t+1})$, first we have
\begin{align*} \label{eq:lemma-auxiliary-0}
    &d_{TV}\left(\mathbb{P}(s_{t+1}\in \cdot),\mathbb{P}(\widetilde{s}_{t+1} \in \cdot)\right) \\
    &= \frac{1}{2} \int_{s' \in \mathcal{S}}\left| \mathbb{P}(s_{t+1}=ds')-\mathbb{P}(\widetilde{s}_{t+1}=ds') \right| \\
    &= \frac{1}{2} \int_{s' \in \mathcal{S}}\left| \int_{s\in \mathcal{S}}\sum_{a \in \mathcal{A}}\mathbb{P}(s_{t}=ds,a_{t}=a,s_{t+1}=ds')-\mathbb{P}(\widetilde{s}_t=ds,\widetilde{a}_t=a,\widetilde{s}_{t+1}=ds') \right| \\
    &\leq \frac{1}{2} \int_{s' \in \mathcal{S}}\int_{s\in \mathcal{S}}\sum_{a \in \mathcal{A}}\left| \mathbb{P}(s_t=ds,a_t=a,s_{t+1}=ds')-\mathbb{P}(\widetilde{s}_t=ds,\widetilde{a}_t=a,\widetilde{s}_{t+1}=ds') \right| \\
    &= \frac{1}{2} \int_{s \in \mathcal{S}}\sum_{a \in \mathcal{A}}\int_{s'\in \mathcal{S}}\left| \mathbb{P}(s_t=ds,a_t=a,s_{t+1}=ds')-\mathbb{P}(\widetilde{s}_t=ds,\widetilde{a}_t=a,\widetilde{s}_{t+1}=ds') \right| \\
    &= d_{TV}\left( \mathbb{P}(x_{t} \in \cdot),\mathbb{P}(\widetilde{x}_{t} \in \cdot) \right), \numberthis
\end{align*}
where the second last equality is due to Tonelli's theorem.
Next we have
\begin{align*} \label{eq:lemma-auxiliary-1}
    &d_{TV}\left(\mathbb{P}(x_t \in \cdot),\mathbb{P}(\widetilde{x}_t \in \cdot) \right) \\
    &= \frac{1}{2}\int_{s \in \mathcal{S}}\sum_{a \in \mathcal{A}}\int_{s' \in \mathcal{S}}\left|\mathbb{P}(s_t=ds,a_t=a,s_{t+1}=ds')-\mathbb{P}(\widetilde{s}_t=ds,\widetilde{a}_t=a,\widetilde{s}_{t+1}=ds') \right| \\
    &= \frac{1}{2}\int_{s \in \mathcal{S}}\sum_{a \in \mathcal{A}}\left|\mathbb{P}(s_t=ds,a_t=a)-\mathbb{P}(\widetilde{s}_t=ds,\widetilde{a}_t=a) \right|\int_{s' \in \mathcal{S}} \mathcal{P}(s_{t+1}=ds'|s_t=ds,a_t=a) \\
    &= \frac{1}{2}\int_{s \in \mathcal{S}}\sum_{a \in \mathcal{A}}\left|\mathbb{P}(s_t=ds,a_t=a)-\mathbb{P}(\widetilde{s}_t=ds,\widetilde{a}_t=a) \right| \\
    &= d_{TV}\left(\mathbb{P}\left((s_t,a_t) \in \cdot \right),\mathbb{P}\left((\widetilde{s}_t,\widetilde{a}_t) \in \cdot \right) \right). \numberthis
\end{align*}
Due to the fact that $\theta_{k-\tau_k}$ is dependent on $s_t$, we need to write $\mathbb{P}(s_t,a_t)$ as
\begin{align*}
    \mathbb{P}(s_t,a_t)
    &= \int_{\theta_{k-\tau_k} \in \mathbb{R}^d}\mathbb{P}(s_t,\theta_{k-\tau_k},a_t) \\
    &= \int_{\theta \in \mathbb{R}^d} \mathbb{P}(s_t) \mathbb{P}(\theta_{k-\tau_k}=d\theta|s_t) \pi_{\theta_{k-\tau_k}}(a_t|s_t) \\
    % &=  \mathbb{P}(s_t) \int_{\theta \in \mathbb{R}^d}\mathbb{P}(\theta_{k-\tau_k}=d\theta|s_t) \pi_{\theta_{k-\tau_k}}(a_t|s_t) \\
    &= \mathbb{P}(s_t) \E[\pi_{\theta_{k-\tau_k}}(a_t|s_t)|s_t]. \numberthis
\end{align*}
Then we have
\begin{align*}\label{eq:lemma-auxiliary-2-1}
    &d_{TV}\left(\mathbb{P}\left((s_t,a_t) \in \cdot \right),\mathbb{P}\left((\widetilde{s}_t,\widetilde{a}_t) \in \cdot \right) \right) \\
    &= \frac{1}{2}\int_{s \in \mathcal{S}}\sum_{a \in \mathcal{A}} \left|  \mathbb{P}(s_t=ds) \E[\pi_{\theta_{k-\tau_k}}(a_t=a|s_t=ds)|s_t=ds]-\mathbb{P}(\widetilde{s}_t=ds) \pi_{\theta_{k-d_m}}(\widetilde{a}_t=a|\widetilde{s}_t=ds)\right| \\
    &\leq \frac{1}{2}\int_{s \in \mathcal{S}}\sum_{a \in \mathcal{A}} \left| \mathbb{P}(s_t=ds) \E[\pi_{\theta_{k-\tau_k}}(a_t=a|s_t=ds)|s_t=ds]-\mathbb{P}(s_t=ds) \pi_{\theta_{k-d_m}}(a_t=a|s_t=ds)\right|\\
    &~~~~+ \frac{1}{2}\int_{s \in \mathcal{S}}\sum_{a \in \mathcal{A}} \left| \mathbb{P}(s_t=ds) \pi_{\theta_{k-d_m}}(\widetilde{a}_t=a|\widetilde{s}_t=ds)-\mathbb{P}(\widetilde{s}_t=ds) \pi_{\theta_{k-d_m}}(\widetilde{a}_t=a|\widetilde{s}_t=ds)\right| \\
    &= \frac{1}{2}\int_{s \in \mathcal{S}}\mathbb{P}(s_t=ds) \sum_{a \in \mathcal{A}} \left|  \E[\pi_{\theta_{k-\tau_k}}(a_t=a|s_t=ds)|s_t=ds] - \pi_{\theta_{k-d_m}}(a_t=a|s_t=ds)\right|\\
    &~~~~+ \frac{1}{2}\int_{s \in \mathcal{S}}\left| \mathbb{P}(s_t=ds) -\mathbb{P}(\widetilde{s}_t=ds)\right|.\numberthis
\end{align*}
Using Jensen's inequality, we have
\begin{align*}\label{eq:lemma-auxiliary-2}
    &d_{TV}\left(\mathbb{P}\left((s_t,a_t) \in \cdot \right),\mathbb{P}\left((\widetilde{s}_t,\widetilde{a}_t) \in \cdot \right) \right) \\
    &\leq \frac{1}{2}\int_{s \in \mathcal{S}}\mathbb{P}(s_t=ds) \sum_{a \in \mathcal{A}} \E\left.\left[\left|  \pi_{\theta_{k-\tau_k}}(a_t=a|s_t=ds) - \pi_{\theta_{k-d_m}}(a_t=a|s_t=ds)\right|\right|s_t=ds\right]\\
    &~~~~+ \frac{1}{2}\int_{s \in \mathcal{S}}\left| \mathbb{P}(s_t=ds) -\mathbb{P}(\widetilde{s}_t=ds)\right| \\
    &\leq \frac{1}{2}\int_{s \in \mathcal{S}}\mathbb{P}(s_t=ds) \sum_{a \in \mathcal{A}} \E\left.\left[\|\theta_{k-\tau_k}-\theta_{k-d_m}\|_2\right|s_t=ds\right] + \frac{1}{2}\int_{s \in \mathcal{S}}\left| \mathbb{P}(s_t=ds) -\mathbb{P}(\widetilde{s}_t=ds)\right|\\
    &= \frac{1}{2} |\mathcal{A}| L_\pi \E\|\theta_{k-\tau_k}-\theta_{k-d_m}\|_2 + d_{TV}\left( \mathbb{P}(s_t \in \cdot),\mathbb{P}(\widetilde{s}_t \in \cdot) \right)  \numberthis 
\end{align*}
where the last inequality follows Assumption \ref{assumption:omega}. 

Now we start to prove \eqref{eq:auxiliary-markov-chain-4}. First we have
\begin{align*}\label{eq:lemma-auxiliary-markov-chain-tmp0}
    &d_{TV}\left(\mathbb{P}\big((s_t,a_t)\in \cdot \big),\mathbb{P}\big((\widetilde{s}_t,\widetilde{a}_t)\in \cdot \big)\right) \\
    &\overset{\eqref{eq:lemma-auxiliary-2}}{\leq} d_{TV}\left(\mathbb{P}(s_t\in \cdot),\mathbb{P}(\widetilde{s}_t \in \cdot)\right)+\frac{1}{2}|\mathcal{A}|L_\pi\E\|\theta_{k-\tau_k}-\theta_{k-d_m}\|_2  \\
     &\overset{\eqref{eq:lemma-auxiliary-0}}{\leq} d_{TV}\left(\mathbb{P}(x_{t-1}\in \cdot),\mathbb{P}(\widetilde{x}_{t-1} \in \cdot)\right) +\frac{1}{2}|\mathcal{A}|L_\pi\E\|\theta_{k-\tau_k}-\theta_{k-d_m}\|_2 \\
    &\overset{\eqref{eq:lemma-auxiliary-1}}{=} 
    d_{TV}\left(\mathbb{P}\big((s_{t-1},a_{t-1})\in \cdot\big),\mathbb{P}\big((\widetilde{s}_{t-1},\widetilde{a}_{t-1})\in \cdot \big)\right). \numberthis 
\end{align*}
% Now we have
% \begin{align*}\label{eq:lemma-auxiliary-markov-chain-tmp0}
%     d_{TV}\left(\mathbb{P}(x_t\in \cdot),\mathbb{P}(\widetilde{x}_t \in \cdot)\right)
%     \leq d_{TV}\left(\mathbb{P}(x_{t-1}\in \cdot),\mathbb{P}(\widetilde{x}_{t-1} \in \cdot)\right) +\frac{1}{2}|\mathcal{A}|L_\pi\E\|\theta_{k-\tau_k}-\theta_{k-d_m}\|_2. \numberthis
% \end{align*}
Since $ d_{TV}\left(\mathbb{P}\big((s_{t-m},a_{t-m})\in \cdot\big),\mathbb{P}\big((\widetilde{s}_{t-m},\widetilde{a}_{t-m})\in \cdot \big)\right)=0$, recursively applying \eqref{eq:lemma-auxiliary-markov-chain-tmp0} for $\{t-1,...,t-m\}$ gives
\begin{align*}
d_{TV}\left(\mathbb{P}\big((s_t,a_t)\in \cdot \big),\mathbb{P}\big((\widetilde{s}_t,\widetilde{a}_t)\in \cdot \big)\right)
    &\leq \frac{1}{2}|\mathcal{A}|L_\pi\sum_{j=0}^{m}\E\|\theta_{k-d_j}-\theta_{k-d_m}\|_2 \\
    &\leq \frac{1}{2}|\mathcal{A}|L_\pi\sum_{i=\tau_k}^{d_m}\E\|\theta_{k-i}-\theta_{k-d_m}\|_2, \numberthis
\end{align*}
which completes the proof.
\end{proof}

\subsection{Lipschitz continuity of actor and critic}
We first give a proposition regarding the $L_{\lambda}$-Lipschitz continuity of the regularized policy gradient under proper assumptions, which has been shown by \citep{zhang2019global,agarwal2019optimality}.
\begin{Proposition}\label{prop:Lj-lip}
Suppose Assumption \ref{assumption:omega} hold. For any $\theta, \theta' \in \mathbb{R}^{d}$, we have $\|\nabla J_\lambda(\theta) - \nabla J_\lambda(\theta')\|_2 \leq L_{\lambda} \|\theta - \theta'\|_2$, where $L_{\lambda}$ is a positive constant.
\end{Proposition}

We provide a justification for Lipschitz continuity of $\omega_\theta^*$ in the next proposition.
\begin{Proposition}\label{proposition:omega-lipschitz}
Suppose Assumption \ref{assumption:A}, \ref{assumption:omega} and \ref{assumption:MDP} hold. For any $\theta_1, \theta_2 \in \mathbb{R}^{d}$, we have
\begin{align*}
    \|\omega_{\theta_1}^* - \omega_{\theta_2}^*\|_2 \leq L_{\omega} \|\theta_1-\theta_2\|_2,
\end{align*}
where $L_{\omega} \coloneqq 2r_{\max}|\mathcal{A}|L_\pi (\lambda^{-1}+\lambda^{-2}(1+\gamma))(1+\log_\rho \kappa^{-1} + (1-\rho)^{-1})$.
\end{Proposition}
\begin{proof}
We use $A_1$, $A_2$, $b_1$ and $b_2$ as shorthand notations of $A_{\pi_{\theta_1}}$, $A_{\pi_{\theta_2}}$, $b_{\pi_{\theta_1}}$ and $b_{\pi_{\theta_2}}$ respectively. By Assumption \ref{assumption:A}, $A_{\theta, \phi}$ is invertible for any $\theta \in \mathbb{R}^d$, so we can write $\omega_{\theta}^* = -A_{\theta, \phi}^{-1} b_{\theta, \phi}$. Then we have
\begin{align*}\label{eq:w1*-w2*}
    \|\omega^*_1 - \omega^*_2 \|_2
    &=\| -A_1^{-1}b_1 +  A_2^{-1}b_2\|_2 \\
    &=\| -A_1^{-1}b_1 - A_1^{-1}b_2 + A_1^{-1}b_2 + A_2^{-1}b_2 \|_2 \\
    &=\| -A_1^{-1}(b_1 - b_2) - (A_1^{-1} - A_2^{-1})b_2 \|_2 \\
    &\leq \|A_1^{-1}(b_1 - b_2)\|_2 + \| (A_1^{-1} -  A_2^{-1})b_2\|_2 \\
    &\leq \|A_1^{-1}\|_2\|b_1 - b_2 \|_2 + \| A_1^{-1} -  A_2^{-1} \|_2\|b_2\|_2 \\
    &= \|A_1^{-1}\|_2\|b_1 - b_2 \|_2 + \| A_1^{-1}(A_2-A_1)A_2^{-1} \|_2\|b_2\|_2 \\
    &\leq \|A_1^{-1}\|_2\|b_1 - b_2 \|_2 + \| A_1^{-1} \|_2 \|A_2^{-1} \|_2 \|b_2\|_2 \|(A_2-A_1)\|_2  \\
    &\leq \lambda^{-1} \left\|b_1 - b_2 \right\|_2 + \lambda^{-2}r_{\max} \left\|A_1 - A_2 \right\|_2, \numberthis
\end{align*}
where the last inequality follows Assumption \ref{assumption:A}, and the fact that
\begin{align*}
    \|b_2\|_2 = \left\|\E[r(s,a,s')\phi(s)]\right\|_2 \leq \E\left\|r(s,a,s')\phi(s)\right\|_2 \leq \E\left[|r(s,a,s')|\|\phi(s)\|_2\right] \leq r_{\max}.
\end{align*}
Denote $(s^1,a^1,s'^1)$ and $(s^2,a^2,s'^2)$ as samples drawn with $\theta_1$ and $\theta_2$ respectively, i.e. $s^1 \sim \mu_{\theta_1}$, $a^1 \sim \pi_{\theta_1}$, $s'^1 \sim \mathcal{P}$ and  $s^2 \sim \mu_{\theta_2}$, $a^2 \sim \pi_{\theta_2}$, $s'^2 \sim \mathcal{P}$. Then we have
\begin{align*}\label{eq:b1-b2}
    \left\|b_1 - b_2 \right\|_2
    &= \left\| \E\left[r(s^1,a^1,s'^1)\phi(s^1) \right] -  \E\left[r(s^2,a^2,s'^2)\phi(s^2) \right]\right\|_2 \\
    &\leq \sup_{s,a,s'}\|r(s,a,s')\phi(s)\|_2 \|\mathbb{P}((s^1,a^1,s'^1)\in\cdot)-\mathbb{P}((s^2,a^2,s'^2)\in\cdot)\|_{TV}\\
    &\leq r_{\max} \|\mathbb{P}((s^1,a^1,s'^1)\in\cdot)-\mathbb{P}((s^2,a^2,s'^2)\in\cdot)\|_{TV}\\
    &= 2r_{\max} d_{TV} \left( \mu_{\theta_1}\otimes\pi_{\theta_1}\otimes\mathcal{P}, \mu_{\theta_2}\otimes\pi_{\theta_2}\otimes\mathcal{P} \right) \\
    &\leq 2r_{\max} |\mathcal{A}| L_{\pi} (1+\log_\rho \kappa^{-1} + (1-\rho)^{-1})\|\theta_1-\theta_2\|_2, \numberthis
\end{align*}
where the first inequality follows the definition of total variation (TV) norm, and the last inequality follows Lemma A.1. in \citep{wu2020finite}. Similarly we have:
\begin{align*}\label{eq:a1-a2}
    \left\|A_1 - A_2 \right\|_2
    &\leq 2(1+\gamma) d_{TV} \left( \mu_{\theta_1}\otimes\pi_{\theta_1}, \mu_{\theta_2}\otimes\pi_{\theta_2} \right) \\
    &= (1+\gamma) |\mathcal{A}| L_{\pi} (1+\log_\rho \kappa^{-1} + (1-\rho)^{-1})\|\theta_1-\theta_2\|_2. \numberthis
\end{align*}
Substituting \eqref{eq:b1-b2} and \eqref{eq:a1-a2} into \eqref{eq:w1*-w2*} completes the proof.
\end{proof}

\section{Proof of Main Theorems}

\subsection{Proof of Theorem \ref{theorem:async-tts-critic-iid-double}}\label{section:async-tts-critic-proof-double}

We first define the exact TD update as:
\begin{align*}
    \overline{g}(x,\omega)&\coloneqq \E_{\substack{s \sim \mu_{\theta}, a \sim \pi_\theta, s' \sim \mathcal{P}}} \left[g(x,\omega)\right]. \numberthis
    % &=\E_{\substack{s \sim \mu_{\theta},a \sim \pi_\theta \\ s' \sim \mathcal{P}}} \left[\left(r(s,a,s') + \gamma \phi(s')^\top \omega - \phi(s)^\top \omega\right)\phi(s) \right],
\end{align*}
We also define constant $C_{\delta} \coloneqq r_{\max} + (1+\gamma)\max\{\frac{r_{\max}}{1-\gamma}, R_\omega\}$, and we immediately have
\begin{align}\label{eq:C_delta}
    \|g(x,\omega)\|_2\leq |r(x)+\gamma \phi(s')^\top \omega - \phi(s)^\top \omega|  \leq r_{\max} + (1+\gamma)R_\omega \leq C_\delta 
\end{align}
and likewise, we have $\|\overline{g}(x,\omega)\|_2  \leq   C_\delta$. 

The critic update in Algorithm \ref{algorithm:async-tts-worker} can be written as:
\begin{equation}\label{update:async-tts-critic-update}
    \omega_{k+1} = \Pi_{R_\omega}\left(\omega_k + \beta g(x_{(k)},\omega_{k-\tau_k})\right),
\end{equation}
where $\tau_k$ is the delay of the parameters used in evaluating the $k$th stochastic gradient, and $x_{(k)} \coloneqq (s_{(k)},a_{(k)},s_{(k)}')$ is the sample used to evaluate the stochastic gradient at $k$th update. 
% Now we are ready to give the convergence proof.

\begin{proof}
Using $\omega_k^*$ as shorthand notation of $\omega_{\theta_k}^*$, we start with the optimality gap
\begin{align*}\label{eq:tmp1}
    &\|\omega_{k+1}-\omega^*_{k+1}\|_2^2 \\
    &= \|\Pi_{R_\omega}\left(\omega_k + \beta g(x_{(k)},\omega_{k-\tau_k})\right) - \omega^*_{k+1}\|_2^2 \\
    &\leq \|\omega_k + \beta g(x_{(k)},\omega_{k-\tau_k}) - \omega^*_{k+1}\|_2^2 \\
    &= \left\|\omega_k - \omega^*_k \right\|_2^2 + 2 \beta \left\langle \omega_k-\omega^*_k, g(x_{(k)},\omega_{k-\tau_k})\right\rangle + 2 \left\langle \omega_k-\omega^*_k, \omega_k^*-\omega_{k+1}^*\right\rangle\\
    &\quad + \left\| \omega_k^*-\omega_{k+1}^* + \beta g(x_{(k)},\omega_{k-\tau_k}) \right\|_2^2\\
    &= \left\|\omega_k - \omega^*_k \right\|_2^2 + 2 \beta \left\langle \omega_k-\omega^*_k, g(x_{(k)},\omega_{k-\tau_k})\right\rangle \\&\quad+ 2 \left\langle \omega_k-\omega^*_k, \omega_k^*-\omega_{k+1}^*\right\rangle+2\left\| \omega_k^*-\omega_{k+1}^*\right\|_2^2 + 2C_\delta^2\beta^2. \numberthis
\end{align*}
The second term in \eqref{eq:tmp1} can be decomposed as
\begin{align}\label{eq:omegak-omegak*g/iid}
     \left\langle \omega_k-\omega^*_k, g(x_{(k)},\omega_{k-\tau_k})\right\rangle
     &= \left\langle \omega_k-\omega^*_k, \overline{g}(\theta_k,\omega_k)\right\rangle
     + \left\langle \omega_k-\omega^*_k, g(x_{(k)},\omega_{k-\tau_k})-g(x_{(k)},\omega_{k})\right\rangle \nonumber\\
     &\quad + \left\langle \omega_k-\omega^*_k, g(x_{(k)},\omega_{k})-\overline{g}(\theta_k,\omega_k)\right\rangle.
\end{align}
We first bound $\left\langle \omega_k-\omega^*_k, \overline{g}(\theta_k,\omega_k)\right\rangle$ in \eqref{eq:omegak-omegak*g/iid} as
\begin{align*}\label{eq:tmp2}
    \left\langle \omega_k-\omega^*_k, \overline{g}(\theta_k,\omega_k)\right\rangle
    &= \left\langle \omega_k-\omega^*_k, \overline{g}(\theta_k,\omega_k) - \overline{g}(\theta_k,\omega_k^*)\right\rangle \\
    &= \left\langle \omega_k-\omega^*_k, \E\left[\left(\gamma \phi(s') - \phi(s) \right)^\top (\omega_k-\omega^*_k) \phi(s)\right] \right\rangle \\
    &= \left\langle \omega_k-\omega^*_k, \E\left[\phi(s)\left(\gamma \phi(s') - \phi(s) \right)^\top\right] (\omega_k-\omega^*_k)  \right\rangle \\
    &= \left\langle \omega_k-\omega^*_k, A_{\pi_{\theta_k}} (\omega_k-\omega^*_k) \right\rangle \\
    &\leq  -\lambda \|\omega_k-\omega^*_k\|_2^2, \numberthis
\end{align*}
where the first equality is due to $\overline{g}(\theta,\omega^*_\theta)=A_{\theta, \phi}\omega_\theta^* + b=0 $, and the last inequality follows Assumption \ref{assumption:A}.

We then bound the term $\left\langle \omega_k-\omega^*_k, g(x_{(k)},\omega_{k-\tau_k}) - g(x_{(k)},\omega_k)\right\rangle$ in \eqref{eq:omegak-omegak*g/iid} as
\begin{align*}\label{eq:tmp5}
    \left\langle \omega_k-\omega^*_k, g(x_{(k)},\omega_{k-\tau_k}) - g(x_{(k)},\omega_k)\right\rangle 
    &= \left\langle \omega_k-\omega^*_k, \left(\gamma \phi(s_{(k)}') - \phi(s_{(k)}) \right)^\top (\omega_{k-\tau_k}-\omega_k) \phi(s_{(k)})\right\rangle \\
    &\leq (1+\gamma)\|\omega_k-\omega^*_k\|_2\|\omega_{k-\tau_k}-\omega_k\|_2 \\
    &\leq (1+\gamma)\|\omega_k-\omega^*_k\|_2 \left\|\sum_{i=k-\tau_k}^{k-1}(\omega_{i+1}-\omega_i )\right\|_2 \\
    &\leq (1+\gamma)\|\omega_k-\omega^*_k\|_2 \sum_{i=k-\tau_k}^{k-1}\beta \|g(x_i,\omega_{i-\tau_i})\|_2 \\
    &\leq 2 C_\delta  K_0 \beta \|\omega_k-\omega^*_k\|_2,  \numberthis
\end{align*}
where the last inequality follows the definition of $C_\delta$ in \eqref{eq:C_delta}.
Substituting \eqref{eq:tmp5} and \eqref{eq:tmp2} into \eqref{eq:omegak-omegak*g/iid} gives
\begin{align}\label{eq:omegak-omegak*g2}
     \left\langle \omega_k-\omega^*_k, g(x_{(k)},\omega_{k-\tau_k})\right\rangle
     &\leq  -\lambda \|\omega_k-\omega^*_k\|_2^2 + 2 C_\delta K_0 \beta \|\omega_k-\omega^*_k\|_2 \nonumber\\
     &\quad + \left\langle \omega_k-\omega^*_k, g(x_{(k)},\omega_{k})-\overline{g}(\theta_k,\omega_k)\right\rangle.
\end{align}

Next we jointly bound the third and fourth term in \eqref{eq:tmp1} as
\begin{align*}\label{eq:tmp8}
    &\left\langle \omega_k-\omega^*_k, \omega_k^*-\omega_{k+1}^*\right\rangle + \left\| \omega_k^*-\omega_{k+1}^*\right\|_2^2 \\
    &\leq \left\| \omega_k-\omega^*_k\right\|_2 \left\|\omega_k^*-\omega_{k+1}^*\right\|_2 + \left\| \omega_k^*-\omega_{k+1}^*\right\|_2^2\\
    &\leq 2L_{\omega} \left\| \omega_k-\omega^*_k\right\|_2 \left\|\theta_k-\theta_{k+1}\right\|_2 + 2L_{\omega}^2\left\| \theta_k-\theta_{k+1}\right\|_2^2\\
    &\leq 2L_{\omega} C_p \alpha  \left\| \omega_k-\omega^*_k\right\|_2 + 2L_{\omega}^2 C_p^2 \alpha^2, \numberthis
\end{align*}
where constant $C_p \coloneqq C_\delta C_\psi + \lambda C_\psi$. The second inequality is due to the $L_\omega$-Lipschitz continuity of $\omega_\theta^*$ shown in Proposition \ref{proposition:omega-lipschitz}, and the last inequality follows the fact that
\begin{align*}\label{eq:C_p}
\|\theta_k-\theta_{k+1}\|_2 = \alpha \|\hat{\delta}(x_{(k)}, \omega_{k-\tau_k})\psi_{\theta_{k-\tau_k}}(s_{(k)},a_{(k)}) + \lambda \psi_{\theta_{k-\tau_k}}(x_{(k)}^p)\|_2 \leq
\alpha C_p. \numberthis
\end{align*}
Substituting \eqref{eq:omegak-omegak*g2} and \eqref{eq:tmp8} into \eqref{eq:tmp1}, and taking expectation on both sides yield
\begin{align*} \label{eq:async-tts-tmp0}
  \E\|\omega_{k+1}-\omega^*_{k+1}\|_2^2
    &\leq (1-2\lambda \beta)\E\left\|\omega_k - \omega^*_k \right\|_2^2 + 2\beta(C_1 \frac{\alpha}{\beta} + C_2 K_0\beta)\E\left\| \omega_k-\omega^*_k\right\|_2\\&~~~~+ 2 \beta \E \left\langle \omega_k-\omega^*_k, g(x_{(k)},\omega_k) - \overline{g}(\theta_k,\omega_k)\right\rangle + C_q\beta^2, \numberthis
\end{align*}
where $C_1 \coloneqq L_{\omega} C_p $, $C_2 \coloneqq 2 C_\delta$ and $C_q \coloneqq 2C_\delta^2 + 2L_{\omega}^2 C_p^2 \frac{\alpha^2}{\beta^2}$. 

For brevity, we use $x \sim \theta$ to denote $s \sim \mu_\theta$, $a \sim \pi_\theta$ and $s' \sim \mathcal{P}$ in this proof. Consider the third term in \eqref{eq:async-tts-tmp0} conditioned on $\theta_k,\omega_k,\theta_{k-\tau_k}$. We bound it as
\begin{align*}\label{eq:async-tts-tmp1-iid}
    &\E \left[\left\langle \omega_k-\omega^*_k, g(x_{(k)},\omega_k) - \overline{g}(\theta_k,\omega_k)\right\rangle| \theta_k, \omega_k, \theta_{k-\tau_k}\right] \\
    &= \left\langle \omega_k-\omega^*_k, \E_{x_{(k)}\sim\theta_{k-\tau_k}}\left[g(x_{(k)},\omega_k)|\omega_k\right] - \overline{g}(\theta_k,\omega_k)\right\rangle  \\
    % &= \bigg\langle \omega_k-\omega^*_k, \E_{\substack{s_{(k)}\sim \mu_{\theta_{k-\tau_k}}\\ a_{(k)} \sim \pi_{\theta_{k-\tau_k}} \\ s'_{(k)} \sim \mathcal{P}}}\left[g(x_{(k)},\omega_k)|\omega_k\right] - \overline{g}(\theta_k,\omega_k)\bigg\rangle  \\
    &= \bigg\langle \omega_k-\omega^*_k, \overline{g}(\theta_{k-\tau_k},\omega_k) - \overline{g}(\theta_k,\omega_k)\bigg\rangle \\
    &\leq  \|\omega_k-\omega^*_k\|_2 \|\overline{g}(\theta_{k-\tau_k},\omega_k) - \overline{g}(\theta_k,\omega_k)\|_2 \\
    % &\leq \left(\|\omega_k\|_2 + \|\omega^*_k\|_2 \right)\left\|\overline{g}(\theta_{k-\tau_k},\omega_k) - \overline{g}(\theta_k,\omega_k)\right\|_2  \\
    &\leq  2 R_\omega \left\|\E_{x\sim\theta_{k-\tau_k}}[g(x,\omega_k)] - \E_{x\sim\theta_k}[g(x,\omega_k)]\right\|_2 \\
    % &\leq \sup\left| \left\langle \omega - \omega_\theta^*, \overline{g}(\theta',\omega)\right\rangle \right| d_{TV}(\mu_{\theta_{k-\tau_k}} \otimes \pi_{\theta_{k-\tau_k}} \otimes \mathcal{P}, \mu_{\theta_k} \otimes \pi_{\theta_k} \otimes \mathcal{P}) \\
    % &= 2 R_\omega \left\|\E_{\substack{s \sim \mu_{\theta_{k-\tau_k}} \\ a \sim \pi_{\theta_{k-\tau_k}} \\ s' \sim \mathcal{P}}} \left[\left(r(s,a,s') + \gamma \phi(s')^\top \omega - \phi(s)^\top \omega\right)\phi(s) \right] - \E_{\substack{s \sim \mu_{\theta_k} \\ a \sim \pi_{\theta_k} \\ s' \sim \mathcal{P}}} \left[\left(r(s,a,s') + \gamma \phi(s')^\top \omega - \phi(s)^\top \omega\right)\phi(s) \right]\right\|_2 \\
    &\leq 2R_\omega \sup_{x} \left\|g(x,\omega_k) \right\|_2 \left\|\mu_{\theta_{k-\tau_k}} \otimes \pi_{\theta_{k-\tau_k}} \otimes \mathcal{P} - \mu_{\theta_k} \otimes \pi_{\theta_k} \otimes \mathcal{P}\right\|_{TV} \\
    &\leq 4 R_\omega C_\delta d_{TV}(\mu_{\theta_{k-\tau_k}} \otimes \pi_{\theta_{k-\tau_k}} \otimes \mathcal{P}, \mu_{\theta_k} \otimes \pi_{\theta_k} \otimes \mathcal{P}), \numberthis
\end{align*}
where second last inequality follows the definition of TV norm and the last inequality uses the definition of $C_\delta$ in \eqref{eq:C_delta}. 
% \begin{align*}\label{eq:C_delta-iid}
%     \|g(x,\omega_k)\|_2 
%     &\leq \|\hat{\delta}(x,\omega_k)\|_2\| \phi(s_i)\|_2 \\
%     &= \|r(x)+\gamma \phi(s')^\top \omega_k - \phi(s)^\top \omega_k\|_2 \\
%     % &\leq |r(x)|+\gamma \|\phi(s')\|_2 \|\omega_k\|_2 +\|\phi(s)\|_2 \|\omega_k\|_2 \\
%     &\leq r_{\max}+(1+\gamma)R_\omega \leq C_\delta.  \numberthis
% \end{align*}
% \begin{align*}
%     \E\left[g(x_{(k)},\omega_k)| \omega_k\right] =  \E_{\substack{s_{(k)}\sim \mu_{\theta_{k-\tau_k}}\\ a_{(k)} \sim \pi_{\theta_{k-\tau_k}} \\ s'_{(k)} \sim \mathcal{P}}}\left[(r(s_{(k)}, a_{(k)}, s'_{(k)})+\gamma \phi(s'_{(k)})^\top \omega_k - \phi(s_{(k)})^\top \omega_k) \phi(s_{(k)}) | \omega_k \right] = \overline{g}(\theta_{k-\tau_k}, \omega_k)
% \end{align*}

Define constant $C_3 \coloneqq 2 R_\omega C_\delta |\mathcal{A}| L_\pi (1+\log_\rho \kappa^{-1} + (1-\rho)^{-1})$. Then by following the third item in \cite[Lemma A.1]{wu2020finite}, we can write \eqref{eq:async-tts-tmp1-iid} as
\begin{align*}\label{eq:async-tts-tmp2-iid}
     &\E \left[\left\langle \omega_k-\omega^*_k, g(x_{(k)},\omega_k) - \overline{g}(\theta_k,\omega_k)\right\rangle| \theta_k, \omega_k,\theta_{k-\tau_k}\right] \\
    &\leq 4 R_\omega C_\delta d_{TV}(\mu_{\theta_{k-\tau_k}} \otimes \pi_{\theta_{k-\tau_k}} \otimes \mathcal{P}, \mu_{\theta_k} \otimes \pi_{\theta_k} \otimes \mathcal{P})\\
    &\leq  C_3 \left\|\theta_{k-\tau_k}-\theta_k \right\|_2 \\
    % &= C_3 \left\|\sum_{i=k-\tau_k}^{k-1}(\theta_{i+1} - \theta_i) \right\|_2 \\
    &\leq  C_3 \sum_{i=k-\tau_k}^{k-1}\alpha \|g(x_i,\omega_{i-\tau_i})\|_2 \\
    % &\leq C_3 \sum_{i=k-\tau_k}^{k-1}\alpha C_\delta \\
    % &= C_3 \tau_k\alpha C_\delta \\
    &\leq  C_3 C_\delta K_0\alpha , \numberthis
\end{align*}

Taking total expectation on both sides of \eqref{eq:async-tts-tmp2-iid} and substituting it into \eqref{eq:async-tts-tmp0} yield
\begin{align*}\label{eq:async-tts-tmp3-iid}
  \E\|\omega_{k+1}-\omega^*_{k+1}\|_2^2
    &\leq (1-2\lambda \beta)\E\left\|\omega_k - \omega^*_k \right\|_2^2 + 2\beta\Big(C_1 \frac{\alpha}{\beta} + C_2 K_0\beta\Big)\E\left\| \omega_k-\omega^*_k\right\|_2\\
    &~~~~+ 2 C_3 C_\delta K_0 \beta \alpha  + C_q\beta^2. \numberthis
\end{align*}

which along with the fact $\alpha = \frac{1}{(K+1)^{\frac{3}{5}}}$ and $\beta = \frac{1}{(K+1)^{\frac{2}{5}}}$ implies
\begin{align*}\label{eq:result/iidcritic}
    \frac{1}{K}\sum_{k=1}^K \E\left\|\omega_k-\omega_k^*\right\|_2^2
    &=
 \mathcal{O}\left(\frac{K_0^2}{K^{\frac{4}{5}}}\right)
    + \mathcal{O}\left(\frac{K_0}{K^{\frac{3}{5}}}\right)+\mathcal{O}\left(\frac{1}{K^{\frac{2}{5}}}\right). \numberthis
\end{align*}
% We further have
% \begin{align*}\label{eq:asymp-equiv-critic-iid}
%     \frac{1}{K}\sum_{k=1}^{K}\E\|\omega_k - \omega_k^*\|_2^2
%     &\leq \frac{1}{K}\left(\sum_{k=1}^{K_0-1}4R_\omega^2 + \sum_{k=K_0}^{K}\E\|\omega_k - \omega_k^*\|_2^2 \right)\\
%     &=\frac{K_0-1}{K}4R_\omega^2 + \frac{K-K_0+1}{K}\frac{1}{K-K_0+1}\sum_{k=K_0}^{K}\E\|\omega_k - \omega_k^*\|_2^2 \\
%     &= \mathcal{O}\left(\frac{K_0}{K}\right) + \mathcal{O}\left(\frac{1}{K-K_0+1}\sum_{k=K_0}^K \E\left\|\omega_k-\omega_{k}^*\right\|_2^2\right) \\
%     &= \mathcal{O}\left(\frac{1}{K-K_0+1}\sum_{k=K_0}^K \E\left\|\omega_k-\omega_{k}^*\right\|_2^2\right) \numberthis
% \end{align*}
This completes the proof.
\end{proof}

\subsection{Proof of Theorem \ref{theorem:async-tts-actor-iid-double}}\label{section:async-tts-actor-proof-double}
We first define the `optimal' TD target as:
\begin{align*}
    \delta(x,\theta) &\coloneqq r(s,a,s') + \gamma V_{\pi_\theta}(s') - V_{\pi_\theta}(s).
\end{align*}
The update in Algorithm \ref{algorithm:async-tts-worker} can be written as:
\begin{align}\label{update:async-tts-actor-update}
\theta_{k+1} = \theta_k + \alpha \big(\hat{\delta}(\hat{x}_{(k)},\omega_{k-\tau_k})\psi_{\theta_{k-\tau_k}}(\hat{s}_{(k)},\hat{a}_{(k)})+\lambda \psi_{\theta_{k-\tau_k}}(x_{(k)}^p)\big).
\end{align}
% We also define constant $C_p \coloneqq C_\delta C_\psi$. For any $\theta$, $\omega$ and $x$, the following hold:
% \begin{align*}\label{eq:C_p}
%     &\|\hat{\delta}(x,\omega)\psi_\theta (s,a)\|_2 \leq \|\hat{\delta}(x,\omega)\|_2\|\psi_\theta (s,a)\|_2 \leq C_\delta C_\psi = C_p, \\
%     &\|\nabla J(\theta)\|_2 = \|\E\left[ A_{\pi_\theta}(s,a) \psi_\theta (s,a)\right]\|_2 = \|\E\left[ \delta(x,\theta) \psi_\theta (s,a) \right]\|_2 \leq \E\|\delta(x,\theta) \psi_\theta (s,a)\|_2 \leq C_\delta. \numberthis
% \end{align*}
% We can see that both stochastic policy gradient and exact policy gradient are upper bounded by constant $C_p$. 
For brevity, we use $\omega_k^*$ as shorthand notation of $\omega_{\theta_k}^*$ in this proof. Then we are ready to give the convergence proof.

\begin{proof}
From $L_{\lambda}$-Lipschitz continuity of regularized policy gradient shown in Proposition \ref{prop:Lj-lip}, we have:
\begin{align*}
    J_\lambda(\theta_{k+1})
    &\geq J_\lambda(\theta_k) + \left\langle \nabla J_\lambda(\theta_k), \theta_{k+1}-\theta_k \right\rangle - \frac{L_{\lambda}}{2}\|\theta_{k+1}-\theta_k\|_2^2 \\
    &= J_\lambda(\theta_k) + \alpha \left\langle \nabla J_\lambda(\theta_k), \left( \hat{\delta}(\hat{x}_{(k)},\omega_{k-\tau_k})-\hat{\delta}(\hat{x}_{(k)},\omega_k^*) \right)\psi_{\theta_{k-\tau_k}}(\hat{s}_{(k)},\hat{a}_{(k)}) \right\rangle \\&~~~~+ \alpha \left\langle \nabla J_\lambda(\theta_k),\hat{\delta}(\hat{x}_{(k)},\omega_k^*)\psi_{\theta_{k-\tau_k}}(\hat{s}_{(k)},\hat{a}_{(k)}) +\lambda \psi_{\theta_{k-\tau_k}}(x_{(k)}^p)\right\rangle - \frac{L_{\lambda}}{2} \|\theta_{k+1}-\theta_k \|_2^2 \\
    &\geq J_\lambda(\theta_k) + \alpha \left\langle \nabla J_\lambda(\theta_k), \left( \hat{\delta}(\hat{x}_{(k)},\omega_{k-\tau_k})-\hat{\delta}(\hat{x}_{(k)},\omega_k^*) \right)\psi_{\theta_{k-\tau_k}}(\hat{s}_{(k)},\hat{a}_{(k)}) \right\rangle \\&~~~~+ \alpha \left\langle \nabla J_\lambda(\theta_k),\hat{\delta}(\hat{x}_{(k)},\omega_k^*)\psi_{\theta_{k-\tau_k}}(\hat{s}_{(k)},\hat{a}_{(k)}) +\lambda \psi_{\theta_{k-\tau_k}}(x_{(k)}^p)\right\rangle - \frac{L_{\lambda}}{2}C_p^2\alpha^2,
\end{align*}
where the last inequality follows the definition of $C_p$ in \eqref{eq:C_p}.

Taking expectation on both sides of the last inequality yields
\begin{align*}\label{eq:async-tts-actor-tmp0}
    \E [J_\lambda(\theta_{k+1})]
    &\geq  \E[J_\lambda(\theta_k)] + \alpha \underbracket{\E\left\langle \nabla J_\lambda(\theta_k), \left( \hat{\delta}(\hat{x}_{(k)},\omega_{k-\tau_k})-\hat{\delta}(\hat{x}_{(k)},\omega_k^*) \right)\psi_{\theta_{k-\tau_k}}(\hat{s}_{(k)},\hat{a}_{(k)}) \right\rangle}_{I_1} \\
    &~~~~+ \alpha \underbracket{\E\left\langle \nabla J_\lambda(\theta_k),\hat{\delta}(\hat{x}_{(k)},\omega_k^*)\psi_{\theta_{k-\tau_k}}(\hat{s}_{(k)},\hat{a}_{(k)}) +\lambda \nabla R(\theta_{k-\tau_k})\right\rangle}_{I_2} - \frac{L_{\lambda}}{2}C_p^2\alpha^2. \numberthis
\end{align*}
where we used the fact that $\E[\psi_{\theta_{k-\tau_k}}(x_{(k)}^p)|\theta_{k-\tau_K}]=\nabla R(\theta_{k-\tau_k})$.

We first decompose $I_1$ as
\begin{align*}
    I_1 &= \E\left\langle \nabla J_\lambda(\theta_k), \left( \hat{\delta}(\hat{x}_{(k)},\omega_{k-\tau_k})-\hat{\delta}(\hat{x}_{(k)},\omega_k^*) \right)\psi_{\theta_{k-\tau_k}}(\hat{s}_{(k)},\hat{a}_{(k)}) \right\rangle\\
    &= \underbracket{\E\left\langle \nabla J_\lambda(\theta_k), \left( \hat{\delta}(\hat{x}_{(k)},\omega_{k-\tau_k})-\hat{\delta}(\hat{x}_{(k)},\omega_k) \right)\psi_{\theta_{k-\tau_k}}(\hat{s}_{(k)},\hat{a}_{(k)}) \right\rangle}_{I_1^{(1)}} \\
    &~~~~+ \underbracket{\E\left\langle \nabla J_\lambda(\theta_k), \left( \hat{\delta}(\hat{x}_{(k)},\omega_k)-\hat{\delta}(\hat{x}_{(k)},\omega_k^*) \right)\psi_{\theta_{k-\tau_k}}(\hat{s}_{(k)},\hat{a}_{(k)}) \right\rangle}_{I_1^{(2)}}.\numberthis
\end{align*}
We bound $I_1^{(1)}$ as
\begin{align*}
    I_1^{(1)}
    &= \E\left\langle \nabla J_\lambda(\theta_k), \left( \gamma \phi(\hat{s}'_{(k)})-\phi(\hat{s}_{(k)}) \right)^\top (\omega_{k-\tau_k} - \omega_k) \psi_{\theta_{k-\tau_k}}(\hat{s}_{(k)},\hat{a}_{(k)}) \right\rangle \\
    &\geq -\E\left[\|\nabla J_\lambda(\theta_k)\|_2 \|\gamma \phi(\hat{s}'_{(k)}) - \phi(\hat{s}_{(k)})\|_2 \|\omega_k - \omega_{k-\tau_k}\|_2 \|\psi_{\theta_{k-\tau_k}}(\hat{s}_{(k)},\hat{a}_{(k)})\|_2  \right] \\
    &\geq -(1+\gamma)C_\psi \E\left[\|\nabla J_\lambda(\theta_k)\|_2 \|\omega_k - \omega_{k-\tau_k}\|_2 \right] \\
    &\geq -(1+\gamma)C_\psi C_\delta K_0  \beta \E\|\nabla J_\lambda(\theta_k)\|_2,\numberthis
\end{align*}
where the last inequality follows
\begin{align*}
    \|\omega_k - \omega_{k-\tau_k}\|_2
    &= \left\| \sum_{i=k-\tau_k}^{k-1} (\omega_{i+1}-\omega_i) \right\|_2 \\
    &\leq  \sum_{i=k-\tau_k}^{k-1} \left\| \beta g(x_i,\omega_{i-\tau_i}) \right\|_2\\
    % &\leq \beta \sum_{i=k-\tau_k}^{k-1} \left\| g(x_i,\omega_{i-\tau_i}) \right\|_2\\
    &\leq  \beta K_0 C_\delta, \numberthis
\end{align*}
where the second inequality is due to \eqref{eq:C_delta}.

Then we bound $I_1^{(2)}$ as
\begin{align*}
    I_1^{(2)}
    &=\E\left\langle \nabla J_\lambda(\theta_k), \left( \hat{\delta}(\hat{x}_{(k)},\omega_k)-\hat{\delta}(\hat{x}_{(k)},\omega_k^*)\right)\psi_{\theta_{k-\tau_k}}(\hat{s}_{(k)},\hat{a}_{(k)}) \right\rangle\\
    &= -\E\left\langle \nabla J_\lambda(\theta_k), \left(\gamma \phi(\hat{s}'_{(k)})-\phi(\hat{s}_{(k)})\right)^\top (\omega_k^* - \omega_k)\psi_{\theta_{k-\tau_k}}(\hat{s}_{(k)},\hat{a}_{(k)}) \right\rangle \\
    &\geq - \E\left[\|\nabla J_\lambda(\theta_k)\|_2 \|\gamma \phi(\hat{s}'_{(k)})-\phi(\hat{s}_{(k)})\|_2 \|\omega_k-\omega_k^*\|_2 \|\psi_{\theta_{k-\tau_k}}(\hat{s}_{(k)},\hat{a}_{(k)})\|_2\right] \\
    &\geq - (1+\gamma)C_\psi \E\left[\|\nabla J_\lambda(\theta_k)\|_2 \|\omega_k-\omega_k^*\|_2\right].\numberthis
\end{align*}
Collecting lower bounds of $I_1^{(1)}$ and $I_1^{(2)}$ gives
% \begin{align*}\label{eq:async-tts-actor-I_1}
%     I_1 
%     &\geq - (1+\gamma)C_\psi \E\left[\|\nabla J_\lambda(\theta_k)\|_2\left( C_\delta K_0  \beta + \|\omega_k-\omega_k^*\|_2\right)\right] \numberthis
% \end{align*}
\begin{align*}\label{eq:async-tts-actor-I_1}
    I_1 
    &\geq - 2C_\psi \E\Big[\|\nabla J_\lambda(\theta_k)\|_2\left( C_\delta K_0  \beta + \|\omega_k-\omega_k^*\|_2\right)\Big] \\
    &\geq -\frac{1}{2}\E\|\nabla J_\lambda(\theta_k)\|_2^2 - 2C_\psi^2\E\left[\left( C_\delta K_0  \beta + \|\omega_k-\omega_k^*\|_2\right)^2\right] \\
    &\geq -\frac{1}{2}\E\|\nabla J_\lambda(\theta_k)\|_2^2 - 4C_\psi^2 C_\delta^2 K_0^2  \beta^2 - 4 C_\psi^2\E\|\omega_k-\omega_k^*\|_2^2, \numberthis
\end{align*}
where the the second and third inequality follow Young's inequality.

Now we consider $I_2$. We first decompose $I_2$ as
\begin{align*}\label{eq:I2/theoremdoubleiid}
    I_2
    &= \E\left\langle \nabla J_\lambda(\theta_k), \hat{\delta}(\hat{x}_{(k)},\omega_k^*)\psi_{\theta_{k-\tau_k}}(\hat{s}_{(k)},\hat{a}_{(k)}) +\lambda \nabla R(\theta_{k-\tau_k})\right\rangle \\
    &=\underbracket{\E\left\langle \nabla J_\lambda(\theta_k), \left( \hat{\delta}(\hat{x}_{(k)},\omega_k^*)-\hat{\delta}(\hat{x}_{(k)},\omega_{k-\tau_k}^*)\right)\psi_{\theta_{k-\tau_k}}(\hat{s}_{(k)},\hat{a}_{(k)}) \right\rangle}_{I_2^{(1)}} \\
    &~~~~ + \underbracket{\E\left\langle \nabla J_\lambda(\theta_k), \left( \hat{\delta}(\hat{x}_{(k)},\omega_{k-\tau_k}^*)-\delta(\hat{x}_{(k)},\theta_{k-\tau_k})\right)\psi_{\theta_{k-\tau_k}}(\hat{s}_{(k)},\hat{a}_{(k)}) \right\rangle}_{I_2^{(2)}} \\
    &~~~~ + \underbracket{\E\left\langle \nabla J_\lambda(\theta_k),  \delta(\hat{x}_{(k)},\theta_{k-\tau_k}) \psi_{\theta_{k-\tau_k}}(\hat{s}_{(k)},\hat{a}_{(k)})+\lambda \nabla R(\theta_{k-\tau_k}) -\nabla J_\lambda(\theta_k)\right\rangle}_{I_2^{(3)}} + \|\nabla J_\lambda(\theta_k)\|_2^2.\numberthis
\end{align*}
We bound $I_2^{(1)}$ as
\begin{align*}
    I_2^{(1)}
    &= \E\left\langle \nabla J_\lambda(\theta_k), \left( \hat{\delta}(\hat{x}_{(k)},\omega_k^*)-\hat{\delta}(\hat{x}_{(k)},\omega_{k-\tau_k}^*)\right)\psi_{\theta_{k-\tau_k}}(\hat{s}_{(k)},\hat{a}_{(k)}) \right\rangle \\
    &= \E\left\langle \nabla J_\lambda(\theta_k), \left(\gamma \phi(\hat{s}'_{(k)}) - \phi(\hat{s}_{(k)})\right)^\top \left(\omega_k^*-\omega_{k-\tau_k}^* \right)\psi_{\theta_{k-\tau_k}}(\hat{s}_{(k)},\hat{a}_{(k)}) \right\rangle \\
    &\geq -\E\left[\|\nabla J_\lambda(\theta_k)\|_2 \|\left(\gamma \phi(\hat{s}'_{(k)}) - \phi(\hat{s}_{(k)})\right)^\top\|_2 \left\|\omega_k^*-\omega_{k-\tau_k}^* \right\|_2 \|\psi_{\theta_{k-\tau_k}}(\hat{s}_{(k)},\hat{a}_{(k)})\|_2\right] \\
    &\geq -L_V C_\psi (1+\gamma)\E\left\|\omega_k^*-\omega_{k-\tau_k}^* \right\|_2 \\
    &\geq  -L_V L_\omega C_\psi (1+\gamma)  \E\|\theta_k - \theta_{k-\tau_k}\|_2 \\
    &\geq -L_V L_\omega C_\psi C_p (1+\gamma) K_0 \alpha , \numberthis
\end{align*}
where $L_V\coloneqq \frac{r_{\max}}{1-\gamma} C_\psi+\lambda C_\psi$ is the trivial upper bound of $\|\nabla J_\lambda(\theta)\|_2$ and $\|\nabla J(\theta)\|_2$. The second last inequality follows from Proposition \ref{proposition:omega-lipschitz} and the last inequality uses \eqref{eq:C_p} as
\begin{align*}\label{eq:async-tts-actor-tmp1-iid}
     \|\theta_k - \theta_{k-\tau_k}\|_2
     &\leq \sum_{i=k-\tau_k}^{k-1} \|\theta_{i+1} - \theta_{i}\|_2 \\
     &= \sum_{i=k-\tau_k}^{k-1}\alpha \| \hat{\delta}(\hat{x}_i,\omega_{i-\tau_i})\psi_{\theta_{i-\tau_i}}(\hat{s}_i,\hat{a}_i)\|_2 \\
     &\leq \sum_{i=k-\tau_k}^{k-1} \alpha C_p \leq C_p K_0 \alpha. \numberthis
\end{align*}
We bound $I_2^{(2)}$ as
\begin{align*}
    I_2^{(2)}
    &= \E\left\langle \nabla J_\lambda(\theta_k), \left( \hat{\delta}(\hat{x}_{(k)},\omega_{k-\tau_k}^*)-\delta(\hat{x}_{(k)},\theta_{k-\tau_k})\right)\psi_{\theta_{k-\tau_k}}(\hat{s}_{(k)},\hat{a}_{(k)}) \right\rangle \\
    &\geq -\E\left[\left\|\nabla J_\lambda(\theta_k)\right\|_2 \left| \hat{\delta}(\hat{x}_{(k)},\omega_{k-\tau_k}^*)-\delta(\hat{x}_{(k)},\theta_{k-\tau_k})\right| \|\psi_{\theta_{k-\tau_k}}(\hat{s}_{(k)},\hat{a}_{(k)})\|_2 \right] \\
    &\geq -L_V C_\psi \E \left|\hat{\delta}(\hat{x}_{(k)},\omega_{k-\tau_k}^*)-\delta(\hat{x}_{(k)},\theta_{k-\tau_k}) \right| \\
    &= -L_V C_\psi \E \left|\gamma\left(\phi(\hat{s}'_{(k)})^\top \omega_{k-\tau_k}^*-V_{\pi_{\theta_{k-\tau_k}}}(\hat{s}'_{(k)}) \right) + V_{\pi_{\theta_{k-\tau_k}}}(\hat{s}_{(k)}) - \phi(\hat{s}_{(k)})^\top \omega_{k-\tau_k}^* \right| \\
    &\geq -L_V C_\psi \left(\gamma \E\left|\phi(\hat{s}'_{(k)})^\top \omega_{k-\tau_k}^*-V_{\pi_{\theta_{k-\tau_k}}}(\hat{s}'_{(k)})\right| + \E\left|V_{\pi_{\theta_{k-\tau_k}}}(\hat{s}_{(k)}) - \phi(\hat{s}_{(k)})^\top \omega_{k-\tau_k}^* \right| \right) \\
    &\geq -L_V C_\psi \bigg(\gamma \sqrt{\E\left|\phi(\hat{s}'_{(k)})^\top \omega_{k-\tau_k}^*-V_{\pi_{\theta_{k-\tau_k}}}(\hat{s}'_{(k)})\right|^2} + \sqrt{\E\left|V_{\pi_{\theta_{k-\tau_k}}}(\hat{s}_{(k)}) - \phi(\hat{s}_{(k)})^\top \omega_{k-\tau_k}^* \right|^2} \bigg) \\
    &\geq -L_V C_\psi(1+\gamma)\epsilon_{\rm app}. \numberthis
\end{align*}
We bound $I_2^{(3)}$ as
\begin{align*}\label{eq:grad-bias}
    I_2^{(3)}
    &= \E\left\langle \nabla J_\lambda(\theta_k),  \delta(\hat{x}_{(k)},\theta_{k-\tau_k}) \psi_{\theta_{k-\tau_k}}(\hat{s}_{(k)},\hat{a}_{(k)})+\lambda \nabla R(\theta_{k-\tau_k}) -\nabla J_\lambda(\theta_k)\right\rangle \\
    &= \E \left\langle \nabla J_\lambda(\theta_k), \E\left.\left[ \delta(\hat{x}_{(k)},\theta_{k-\tau_k}) \psi_{\theta_{k-\tau_k}}(\hat{s}_{(k)},\hat{a}_{(k)})\right| \theta_{k-\tau_k}, \theta_k\right]+\lambda \nabla R(\theta_{k-\tau_k}) -\nabla J_\lambda(\theta_k)\right\rangle \\
    &= \E \bigg\langle \nabla J_\lambda(\theta_k), \E_{\substack{\hat{s}_{(k)} \sim d_{\theta_{k-\tau_k}}\\ \hat{a}_{(k)} \sim \pi_{\theta_{k-\tau_k}}}}\left[ A_{\pi_{\theta_{k-\tau_k}}}(\hat{s}_{(k)},\hat{a}_{(k)})\psi_{\theta_{k-\tau_k}}(\hat{s}_{(k)},\hat{a}_{(k)}) \right]+\lambda \nabla R(\theta_{k-\tau_k}) -\nabla J_\lambda(\theta_k)\bigg\rangle, \numberthis
\end{align*}
where we used the fact that
\begin{align*}
&\E\left.\left[ \delta(\hat{x}_{(k)},\theta_{k-\tau_k}) \psi_{\theta_{k-\tau_k}}(\hat{s}_{(k)},\hat{a}_{(k)})\right| \theta_{k-\tau_k}, \theta_k\right]\\
=&\E_{\substack{\hat{s}_{(k)},\hat{a}_{(k)} \sim d_{\theta_{k-\tau_k}}\\ \hat{s}'_{(k)} \sim \mathcal{P}}}\left[ \left(r(\hat{s}_{(k)},\hat{a}_{(k)},\hat{s}'_{(k)}) + \gamma V_{\pi_{\theta_{k-\tau_k}}}(\hat{s}'_{(k)})-  V_{\pi_{\theta_{k-\tau_k}}}(\hat{s}_{(k)}) \right)\psi_{\theta_{k-\tau_k}}(\hat{s}_{(k)},\hat{a}_{(k)}) \bigg| \theta_{k-\tau_k}, \theta_k \right]\\
=&\E_{\substack{ d_{\theta_{k-\tau_k}}}}\left[ \Big(\E_{\hat{s}'_{(k)} \sim \mathcal{P}}\left[r(\hat{s}_{(k)},\hat{a}_{(k)},\hat{s}'_{(k)}) + \gamma V_{\pi_{\theta_{k-\tau_k}}}\left(\hat{s}'_{(k)}\right)\right]-  V_{\pi_{\theta_{k-\tau_k}}}(\hat{s}_{(k)}) \Big)\psi_{\theta_{k-\tau_k}}(\hat{s}_{(k)},\hat{a}_{(k)}) \bigg| \theta_{k-\tau_k}, \theta_k \right]\\
=&\E_{\substack{ d_{\theta_{k-\tau_k}}}}\left[ \left(Q_{\pi_{\theta_{k-\tau_k}}}(\hat{s}_{(k)},\hat{a}_{(k)}) -  V_{\pi_{\theta_{k-\tau_k}}}(\hat{s}_{(k)}) \right)\psi_{\theta_{k-\tau_k}}(\hat{s}_{(k)},\hat{a}_{(k)}) \bigg| \theta_{k-\tau_k}, \theta_k\right]\\
=&\E_{\substack{\ d_{\theta_{k-\tau_k}}}}\left[ A_{\pi_{\theta_{k-\tau_k}}}(\hat{s}_{(k)},\hat{a}_{(k)})\psi_{\theta_{k-\tau_k}}(\hat{s}_{(k)},\hat{a}_{(k)})\bigg| \theta_{k-\tau_k}, \theta_k \right].
\end{align*}
Continuing from \eqref{eq:grad-bias},
\begin{align*}
    I_2^{(3)}
    &=\bigg\langle \nabla J_\lambda(\theta_k),  \nabla J_\lambda(\theta_{k-\tau_k})-\nabla J_\lambda(\theta_k)\bigg\rangle \\
    &\geq -\|\nabla J_\lambda(\theta_k) \|_2 \|\nabla J_\lambda(\theta_{k-\tau_k}) -\nabla J_\lambda(\theta_k)\|_2 \\
    &\geq -L_V L_{\lambda} \|\theta_{k-\tau_k} -\theta_k\|_2 \geq -L_V L_{\lambda} C_p K_0 \alpha, \numberthis
\end{align*}
where the second last inequality is due to $L_{\lambda}$-Lipschitz continuity of policy gradient shown in Proposition \ref{prop:Lj-lip}, and the last inequality follows \eqref{eq:async-tts-actor-tmp1-iid}.

Collecting lower bounds of $I_2^{(1)}$, $I_2^{(2)}$ and $I_2^{(3)}$ gives
\begin{align}\label{eq:async-tts-actor-I_2-iid}
    I_2
    &\geq -D_1 K_0 \alpha - L_V C_\psi (1+\gamma) \epsilon_{\rm app},
\end{align}
where constant $D_1 \coloneqq L_V L_\omega C_\psi C_p (1+\gamma) + L_V L_{\lambda} C_p$.

Substituting (\ref{eq:async-tts-actor-I_1}) and (\ref{eq:async-tts-actor-I_2-iid}) into (\ref{eq:async-tts-actor-tmp0}) yields
\begin{align*}\label{eq:async-tts-actor-tmp2-iid}
     \E [J_\lambda(\theta_{k+1})]
    &\!\geq\!  \E[J_\lambda(\theta_k)] + \frac{\alpha}{2} \E\|\nabla J_\lambda(\theta_k)\|_2^2 - 4 C_\psi^2 \alpha\E\|\omega_k-\omega_k^*\|_2^2- 4C_\psi^2 C_\delta^2 K_0^2  \alpha\beta^2\\
    &~~~ \!-\! 2 L_V C_\psi  \epsilon_{\rm app}\alpha - (D_1 K_0+\frac{L_{\lambda}}{2}C_p^2)\alpha^2 . \numberthis
\end{align*}
After telescoping, we have
\begin{align}\label{eq:sumalphakgrad/iid}
    \sum_{k=1}^K \frac{1}{2} \E\|\nabla J_\lambda(\theta_k)\|_2^2 
    &\leq \frac{1}{\alpha} \big(J^* - J_\lambda (\theta_{K_0})\big) + 4 C_\psi^2 \sum_{k=1}^K\E\|\omega_k-\omega_k^*\|_2^2 
    + 4 K C_\psi^2 C_\delta^2 K_0^2  \beta^2 \nonumber\\
    &~~~ +  2 K L_V C_\psi \epsilon_{\rm app}+K(D_1 K_0+\frac{L_{\lambda}}{2}C_p^2)\alpha .
\end{align}
Select $\alpha = K^{-\frac{3}{5}}$ and $\beta = K^{-\frac{2}{5}}$, we have
\begin{align}
    \frac{1}{K}\sum_{k=1}^K \E\|\nabla J_\lambda(\theta_k)\|_2^2 =\mathcal{O}\Big(\frac{1}{K^{\frac{2}{5}}}\Big)+ \mathcal{O}\Big(\frac{1}{K}\sum_{k=1}^K\E\|\omega_k-\omega_k^*\|_2^2\Big)+\mathcal{O}\Big(\frac{K_0^2}{K^{\frac{4}{5}}}\Big)+\mathcal{O}\Big(\frac{K_0}{K^{\frac{3}{5}}}\Big)+\mathcal{O}(\epsilon_{\rm app}).
\end{align}
This completes the proof.
\end{proof}

\subsection{Proof of Theorem \ref{theorem:async-tts-critic}}
Given the definition in Section \ref{section:async-tts-critic-proof-double}, we now give the convergence proof of critic update in Algorithm \ref{algorithm:async-tts-worker} with linear function approximation and Markovian sampling.

\begin{proof}
By following the derivation of (\ref{eq:async-tts-tmp0}), we have
\begin{align*}\label{eq:async-tts-tmp0-markov}
  \E\|\omega_{k+1}-\omega^*_{k+1}\|_2^2
    &\leq (1-2\lambda \beta)\E\left\|\omega_k - \omega^*_k \right\|_2^2 + 2\beta\Big(C_1 \frac{\alpha}{\beta} + C_2 K_0\beta\Big)\E\left\| \omega_k-\omega^*_k\right\|_2\\&~~~~+ 2 \beta \E \left\langle \omega_k-\omega^*_k, g(x_{(k)},\omega_k) - \overline{g}(\theta_k,\omega_k)\right\rangle + C_q\beta^2, \numberthis
\end{align*}
where $C_1 \coloneqq C_p L_{\omega}$, $C_2 \coloneqq C_\delta (1+\gamma)$ and $C_q \coloneqq 2C_\delta^2 + 2L_{\omega}^2 C_p^2 \max_{(k)}\frac{\alpha^2}{\beta^2}$.

Now we consider the third item in the last inequality. For some $m \in \mathbb{N}^+$, we define $M \coloneqq (K_0+1)m + K_0$. Following Lemma \ref{lemma:tts-critic-noise} (to be presented in Section \ref{sec:support:3}),  for some $d_m \leq M$ and positive constants $C_4, C_5, C_6, C_7$, we have
\begin{align*}\label{eq:stepsize*C_p}
  &\E \left\langle \omega_k-\omega_k^*,g(x_{(k)},\omega_k)-\overline{g}(\theta_k,\omega_k) \right\rangle \\
  &\leq  C_4 \E\|\theta_k - \theta_{k-d_m}\|_2 + C_5 \sum_{i=\tau_k}^{d_m}\E\|\theta_{k-i}-\theta_{k-d_m}\|_2 + C_6 \E\| \omega_k-\omega_{k-d_m} \|_2 + C_7 \kappa \rho^{m-1} \\
  &\leq C_4 \sum_{i={k-d_m}}^{k-1}\!\! \E\|\theta_{i+1}-\theta_i\|_2 + C_5 \sum_{i=\tau_k}^{d_m-1}\sum_{j=k-d_m}^{k-i-1}\!\E\|\theta_{j+1}-\theta_j\|_2 + C_6\! \sum_{i=k-d_m}^{k-1}\!\E\|\omega_{i+1}-\omega_i\|_2+ C_7\kappa \rho^{m-1} \\
%   &\leq C_4 \sum_{i={k-d_m}}^{k-1} \alpha C_p + C_5 \sum_{i=\tau_k}^{d_m-1}\sum_{j=k-d_m}^{k-i-1}\alpha C_p + C_6 \sum_{i=k-d_m}^{k-1}\beta C_\delta + C_7\kappa \rho^{m-1} \\
%       &\leq C_4 \alpha\sum_{i={k-d_m}}^{k-1}  C_p + C_5 \alpha\sum_{i=\tau_k}^{d_m-1}\sum_{j=k-d_m}^{k-i-1} C_p + C_6 \beta\sum_{i=k-d_m}^{k-1} C_\delta + C_7\kappa \rho^{m-1} \\
%   &\leq C_4 \sum_{i={k-d_m}}^{k-1} \alpha C_p + C_5 \sum_{i=\tau_k}^{d_m-1}\sum_{j=k-d_m}^{k-\tau_k-1}\alpha C_p + C_6 \sum_{i=k-d_m}^{k-1}\beta C_\delta + C_7\kappa \rho^{m-1} \\
%   &\leq C_4 \alpha C_p \sum_{i={k-d_m}}^{k-1} 1 + C_5 \alpha C_p \sum_{i=\tau_k}^{d_m-1}\sum_{j=k-d_m}^{k-1} 1 + C_6 \beta C_\delta \sum_{i=k-d_m}^{k-1} 1 + C_7\kappa \rho^{m-1} \\
  &\leq C_4 d_m C_p \alpha + C_5 (d_m-\tau_k)^2 C_p \alpha + C_6 d_m C    _\delta\beta + C_7\kappa \rho^{m-1} \\
%   &\leq \left[ C_4 \left( (K_0+1)m + K_0 \right) + C_5 \left( (K_0+1)m + K_0 \right)^2 \right] C_p \alpha_{k-(K_0+1)m-K_0} + C_6 \left( (K_0+1)m + K_0 \right) C_\delta \beta_{k-(K_0+1)m-K_0}  + C_7\kappa \rho^{m-1}
  &\leq \left( C_4  M + C_5 M^2 \right) C_p \alpha + C_6  M  C_\delta \beta  + C_7\kappa \rho^{m-1}, \numberthis
\end{align*}
where the last inequality is due to $\tau_k \geq 0$ and $d_m \leq M$. 

Further letting $m = m_K$ which is defined in \eqref{eq:m_K-definition} yields
\begin{align*}\label{eq:async-tts-markov-noise-bound}
    &\E \left\langle \omega_k-\omega_k^*,g(x_{(k)},\omega_k)-\overline{g}(\theta_k,\omega_k) \right\rangle \\
    &= \left( C_4  M_K + C_5 M_K^2 \right) C_p \alpha + C_6 C_\delta M_K \beta  +  C_7\kappa \rho^{m_K-1}\\
    &\leq \left( C_4  M_K + C_5 M_K^2 \right) C_p \alpha + C_6 C_\delta M_K \beta  + C_7 \alpha, \numberthis
\end{align*}
where $M_K = (K_0 + 1)m_K+K_0$, and the last inequality follows the from $m_K=\mathcal{O}(\log K)$.

Substituting (\ref{eq:async-tts-markov-noise-bound}) into (\ref{eq:async-tts-tmp0-markov}) gives
\begin{align}
    \E\|\omega_{k+1}-\omega^*_{k+1}\|_2^2
    &\leq (1-2\lambda \beta)\E\left\|\omega_k - \omega^*_k \right\|_2^2 + 2\beta\Big(C_1 \frac{\alpha}{\beta} + C_2 K_0\beta\Big)\E\left\| \omega_k-\omega^*_k\right\|_2 \nonumber\\
    &~~~~+  2\beta\Big(\left( C_4  M_K + C_5 M_K^2 \right) C_p \alpha + C_6 C_\delta M_K \beta  + C_7 \alpha \Big) + C_q\beta^2. \numberthis
\end{align}

Select $\alpha = K^{-\frac{3}{5}}$ and $\beta = K^{-\frac{2}{5}}$. After  telescoping, we have
\begin{align*}
    &\frac{1}{K}\sum_{k=1}^K \E\left\|\omega_k-\omega_{k}^*\right\|_2^2 =
    \mathcal{O}\left(\frac{1}{K^{\frac{2}{5}}}\right)
    + \mathcal{O}\left(\frac{K_0^2\log^2 K}{K^{\frac{3}{5}}}\right)
    + \mathcal{O}\left(\frac{K_0 \log K}{K^{\frac{2}{5}}}\right).
\end{align*}
This completes the proof.
\end{proof}

\subsection{Proof of Theorem \ref{theorem:async-tts-actor}}
Given the definition in section \ref{section:async-tts-actor-proof-double}, we now give the convergence proof of actor update in Algorithm \ref{algorithm:async-tts-worker} with linear value function approximation and Markovian sampling method.

\begin{proof}
By following the derivation of (\ref{eq:async-tts-actor-tmp0}), we have
\begin{align*}\label{eq:async-tts-actor-tmp0-markov}
       \E [J_\lambda(\theta_{k+1})]
    &\geq  \E[J_\lambda(\theta_k)] + \alpha \underbracket{\E\left\langle \nabla J_\lambda(\theta_k), \left( \hat{\delta}(\hat{x}_{(k)},\omega_{k-\tau_k})-\hat{\delta}(\hat{x}_{(k)},\omega_k^*) \right)\psi_{\theta_{k-\tau_k}}(\hat{s}_{(k)},\hat{a}_{(k)}) \right\rangle}_{I_1} \\
    &~~~~+ \alpha \underbracket{\E\left\langle \nabla J_\lambda(\theta_k),\hat{\delta}(\hat{x}_{(k)},\omega_k^*)\psi_{\theta_{k-\tau_k}}(\hat{s}_{(k)},\hat{a}_{(k)}) +\lambda \nabla R(\theta_{k-\tau_k})\right\rangle}_{I_2} - \frac{L_{\lambda}}{2}C_p^2\alpha^2. \numberthis
\end{align*}

The item $I_1$ can be bounded by following (\ref{eq:async-tts-actor-I_1}) as
\begin{align*}\label{eq:async-tts-actor-I_1-markov}
       I_1 
    &\geq -\frac{1}{2}\E\|\nabla J_\lambda(\theta_k)\|_2^2 - 4C_\psi^2 C_\delta^2 K_0^2  \beta^2 - 4 C_\psi^2\E\|\omega_k-\omega_k^*\|_2^2. \numberthis
\end{align*}

Next we consider $I_2$. We first decompose it as
\begin{align}\label{eq:I_2-markov}
    I_2
    &= \E\left\langle \nabla J_\lambda (\theta_k), \hat{\delta}(x_{(k)},\omega_k^*)\psi_{\theta_{k-\tau_k}}(s_{(k)},a_{(k)})+\lambda \nabla R(\theta_{k-\tau_k}) \right\rangle \nonumber\\
    &=\underbracket{\E\left\langle \nabla J_\lambda(\theta_k), \left( \hat{\delta}(x_{(k)},\omega_k^*)-\delta(x_{(k)},\theta_k)\right)\psi_{\theta_{k-\tau_k}}(s_{(k)},a_{(k)}) \right\rangle}_{I_2^{(1)}} \nonumber\\
    &~~~~+\underbracket{\E\left\langle \nabla J_\lambda(\theta_k), \delta(x_{(k)},\theta_k)\psi_{\theta_{k-\tau_k}}(s_{(k)},a_{(k)}) -\nabla J(\theta_k) \right\rangle}_{I_2^{(2)}} \nonumber\\
    &~~~~+\underbracket{\E\left\langle \nabla J_\lambda(\theta_k), \lambda \nabla R(\theta_{k-\tau_k}) -\lambda\nabla R(\theta_k) \right\rangle}_{I_2^{(3)}}+\E\|\nabla J_\lambda(\theta_k)\|_2^2.
\end{align}
For some $m \in \mathbb{N}^+$, define $M \coloneqq (K_0+1)m+K_0$. Following Lemma \ref{lemma:tts-epsilon}, for some $d_m \leq M$ and positive constants $D_2, D_3, D_4, D_5$, $I_2^{(1)}$ can be bounded as
\begin{align*}\label{eq:async-tts-actor-I_2^1}
    I_2^{(1)}&=
    \E\left\langle \nabla J_\lambda(\theta_k), \left( \hat{\delta}(x_{(k)},\omega_k^*)-\delta(x_{(k)},\theta_k)\right)\psi_{\theta_{k-\tau_k}}(s_{(k)},a_{(k)}) \right\rangle \\
    &\geq -D_2 \E\|\theta_{k-\tau_k}-\theta_{k-d_m}\|_2 - D_3\E\|\theta_k - \theta_{k-d_m}\|_2 - D_4 \sum_{i=k-d_m}^{k-\tau_k}\E\|\theta_{i}-\theta_{k-d_m}\|_2\\
    &~~~~ - D_5 \kappa \rho^{m-1} - L_V C_\psi(1+\gamma)\epsilon_{\rm app} \\
    &\geq -D_2 (d_m-\tau_k) C_p \alpha  - D_3 d_m C_p \alpha   - D_4 (d_m-\tau_k)^2 C_p \alpha  \\
    &~~~~ - D_5 \kappa \rho^{m-1} - (1+\gamma)L_V C_\psi\epsilon_{\rm app}, \numberthis
\end{align*}
where the derivation of the last inequality is similar to that of (\ref{eq:stepsize*C_p}). By setting $m=m_K$ in (\ref{eq:async-tts-actor-I_2^1}), and following the fact that $d_{m_K} \leq M_K$ and $\tau_k \geq 0$, we have
\begin{align*}\label{eq:tts-actor-bound-to-step}
    I_2^{(1)}
    &\geq -D_2 M_K C_p \alpha -D_3 M_K C_p \alpha  - D_4 M_K^2  C_p \alpha - D_5 \kappa \rho^{m_K-1} - (1+\gamma)L_V C_\psi\epsilon_{\rm app} \\
    &= -\left( (D_2 + D_3) C_p M_K  + D_4 C_p M_K^2 \right) \alpha - D_5 \kappa \rho^{m_K-1} - (1+\gamma)L_V C_\psi\epsilon_{\rm app} \\
    &\geq -\left( (D_2 + D_3) C_p M_K  + D_4 C_p M_K^2 \right) \alpha - D_5 \alpha - (1+\gamma)L_V C_\psi \epsilon_{\rm app}, \numberthis
\end{align*}
where the last inequality is due to the definition of $m_K$.

Following Lemma \ref{lemma:tts-actor-noise}, for some positive constants $D_6, D_7$ and $D_8$, we bound $I_2^{(2)}$ as
\begin{align*}
    I_2^{(2)}&=
    \E\left\langle \nabla J_\lambda(\theta_k), \delta(x_{(k)},\theta_k)\psi_{\theta_{k-\tau_k}}(s_{(k)},a_{(k)}) - \nabla J(\theta_k) \right\rangle \\
    &\geq \!-\!D_6 \E\|\theta_{k-\tau_k} \!-\! \theta_{k-d_m}\|_2 \!-\! D_7 \E\|\theta_k\!-\!\theta_{k-d_m}\|_2
    \!-\! D_8 \sum_{i=\tau_k}^{d_m} \E\|\theta_{k-i}\!-\!\theta_{k-d_m}\|_2 \!-\! D_9 \kappa \rho^{m-1}.
\end{align*}
Similar to the derivation of (\ref{eq:tts-actor-bound-to-step}), we have
\begin{align*}\label{eq:async-tts-actor-I_2^2}
     I_2^{(2)}
     &\geq -\left( D_6  C_p M_K + D_7 C_p M_K  + D_8 C_p M_K^2 \right) \alpha - D_9 \alpha. \numberthis
\end{align*}
Term $I_2^{(3)}$ can be bounded as
\begin{align}
    I_2^{(3)} \geq -\lambda L_V \|\nabla R(\theta_k) - \nabla R(\theta_{k-\tau_k})\|_2 &\geq -\lambda L_V L_\psi\|\theta_k- \theta_{k-\tau_k}\|_2 \nonumber\\
    &\geq -\lambda L_V L_\psi K_0  C_p \alpha.
\end{align}

Collecting the lower bounds of $I_2^{(1)}$, $I_2^{(2)}$ and $I_2^{(3)}$ yields
\begin{align*}\label{eq:async-tts-actor-I_2-markov}
    I_2 
    &\geq- D_K \alpha - (1+\gamma)L_V C_\psi \epsilon_{\rm app} + \E\|\nabla J_\lambda(\theta_k)\|_2^2, \numberthis
\end{align*}
where we define $D_K \coloneqq C_p(D_4 + D_8) M_K^2  + C_p(D_2 + D_3 + D_6 + D_7 )M_K + \lambda L_V L_\psi K_0  C_p + D_5 + D_9$ for brevity.

Substituting \eqref{eq:async-tts-actor-I_1-markov} and \eqref{eq:async-tts-actor-I_2-markov} into \eqref{eq:async-tts-actor-tmp0-markov} yields
\begin{align*}
       \E [J_\lambda(\theta_{k+1})]
    &\!\geq\!  \E[J_\lambda(\theta_k)] + \frac{\alpha}{2} \E\|\nabla J_\lambda(\theta_k)\|_2^2 - 4 C_\psi^2 \alpha\E\|\omega_k-\omega_k^*\|_2^2- 4C_\psi^2 C_\delta^2 K_0^2  \alpha\beta^2\\
    &~~~ \!-\! 2 L_V C_\psi  \epsilon_{\rm app}\alpha - D_K \alpha^2 . \numberthis
\end{align*}
Choose step size $\alpha=K^{-\frac{3}{5}}$, $\beta=K^{-\frac{2}{5}}$. With $D_K = \mathcal{O}(M_K^2)=\mathcal{O}(K_0^2 \log^2 K)$, the last inequality implies
\begin{align}
    \frac{1}{K}\sum_{k=1}^K\E\|\nabla J_\lambda(\theta_k)\|_2^2 = \mathcal{O}\Big(\frac{1}{K}\sum_{k=1}^K\E\|\omega_k - \omega_k^*\|_2^2\Big)+\mathcal{O}\Big(\frac{K_0^2 \log^2 K}{K^{\frac{3}{5}}} \Big)+\mathcal{O}\Big(\epsilon_{\rm app} \Big).
\end{align}
This completes the proof.
\end{proof}

\subsection{Proof of Theorem \ref{theorem:global}}

\begin{proof}
We define an event $E_k$ as $\|\nabla J_\lambda (\theta_k)\| \leq \frac{  \lambda}{2|\mathcal{S}||\mathcal{A}|}$ and its complement $E_k^c$ as $\|\nabla J_\lambda (\theta_k)\| > \frac{ \lambda}{2|\mathcal{S}||\mathcal{A}|}$. We use $\mathbf{1}_{E_k}$ to indicate whether the event happens or not, i.e. $\mathbf{1}_{E_k}=1$ if $E_k$ happens and $\mathbf{1}_{E_k}=0$ if $E_k^c$ happens. Then we have for the optimality gap:
\begin{align}\label{eq:sumj*-jt/global}
    \sum_{k=1}^{K} \E\big[J^* - J (\theta_k)\big] &= \sum_{k=1}^{K}\E\big[\big(J^* - J (\theta_k) \big) \mathbf{1}_{E_k}\big] + \sum_{k=1}^{K}\E\big[\big(J^* - J (\theta_k) \big) \mathbf{1}_{E_k^c}\big] \nonumber\\
    &\leq \frac{2\lambda}{1-\gamma} \Big\|\frac{d_{\pi^*}}{\eta} \Big\|_{\infty}  \sum_{k=1}^{K}\E\big[\mathbf{1}_{E_k}\big] + \sum_{k=1}^{K}\E\big[\big(J^* - J (\theta_k) \big) \mathbf{1}_{E_k^c}\big] \nonumber\\
    &\leq \frac{2\lambda}{1-\gamma} \Big\|\frac{d_{\pi^*}}{\eta} \Big\|_{\infty}  \sum_{k=1}^{K}\E\big[\mathbf{1}_{E_k}\big] + J^*\sum_{k=1}^{K} \E\big[\mathbf{1}_{E_k^c} \big]\nonumber\\
    &\leq \frac{2\lambda}{1-\gamma} \Big\|\frac{d_{\pi^*}}{\eta} \Big\|_{\infty}  K + J^*\sum_{k=1}^{K} \E\big[\mathbf{1}_{E_k^c}\big],
\end{align}
where the first inequality follows from Lemma \ref{lemma:gradientdominance}.

Now it suffices to bound $\sum_{k=1}^{K} \E\big[\mathbf{1}_{E_k^c}\big]$.
\begin{align}
    \sum_{k=1}^{K} \E\|\nabla J_\lambda (\theta_k)\|^2 &\geq \sum_{k=1}^{K} \E\big[\|\nabla J_\lambda (\theta_k)\|^2 \mathbf{1}_{E_k^c}\big] \nonumber\\
    &\geq \sum_{k=1}^{K} \frac{\lambda^2}{4|\mathcal{S}|^2|\mathcal{A}|^2}\E\big[\mathbf{1}_{E_k^c}\big]
\end{align}
Substituting the above inequality into \eqref{eq:sumj*-jt/global} and dividing both sides by $K$ give
\begin{align}\label{eq:final/global}
    \frac{1}{K}\sum_{k=1}^{K} \E\big[J^* - J (\theta_k)\big] \leq \frac{2\lambda}{1-\gamma} \Big\|\frac{d_{\pi^*}}{\eta} \Big\|_{\infty}+\frac{4|\mathcal{S}|^2|\mathcal{A}|^2}{\lambda^2}\frac{1}{K}\sum_{k=1}^{K} \E\|\nabla J_\lambda (\theta_k)\|^2
\end{align}
It is known that the softmax policy satisfies Assumption \ref{assumption:omega}, thus we immediately know that Theorem \ref{theorem:async-tts-actor-iid-double} and Theorem \ref{theorem:async-tts-actor} hold. Furthermore, by assumption \ref{assumption:linear approximable}, we have $\phi(s)^\top\omega_\theta^*=V_{\pi_\theta}(s)$, leading to $\epsilon_{\rm app}=0$. Applying Theorem \ref{theorem:async-tts-actor-iid-double} and \ref{theorem:async-tts-actor} to \eqref{eq:final/global} completes the proof.
\end{proof}

\section{Supporting Lemmas}
\subsection{Supporting Lemmas for Theorem \ref{theorem:async-tts-critic}}\label{sec:support:3}
\begin{Lemma}\label{lemma:tts-critic-noise}
For any $m \geq 1$ and $k \geq (K_0+1)m+K_0+1$, we have
\begin{align*}
    \E \left\langle \omega_k-\omega_{\theta_k}^*,g(x_{(k)},\omega_k)-\overline{g}(\theta_k,\omega_k) \right\rangle &\leq  C_4 \E\|\theta_k - \theta_{k-d_m}\|_2 + C_5 \sum_{i=\tau_k}^{d_m}\E\|\theta_{k-i}-\theta_{k-d_m}\|_2 \\&~~~~+ C_6 \E\| \omega_k-\omega_{k-d_m} \|_2 + C_7 \kappa \rho^{m-1} ,
\end{align*}
where constant $d_m \leq (K_0+1)m+K_0$, and $C_4 \coloneqq 2 C_\delta L_{\omega} + 4R_\omega C_\delta |\mathcal{A}| L_\pi (1+\log_\rho \kappa^{-1} + (1-\rho)^{-1})$, $C_5 \coloneqq 4R_\omega C_\delta |\mathcal{A}|L_\pi$ and $C_6 \coloneqq  4(1+\gamma)R_\omega + 2 C_\delta$, $C_7 \coloneqq 8R_\omega C_\delta$.
\end{Lemma}
\begin{proof}
Consider the collection of random samples $\{x_{(k-K_0-1)},x_{(k-K_0)},...,x_{(k)}\}$. Suppose $x_{(k)}$ is sampled by worker $n$, then due to Assumption \ref{assumption:delay}, $\{x_{(k-K_0-1)},x_{(k-K_0)},...,x_{(k-1)}\}$ will contain at least another sample drawn by worker $n$. Therefore, $\{x_{(k-(K_0+1)m)},x_{(k-(K_0+1)m+1)},...,x_{(k-1)}\}$ will contain at least $m$ samples from worker $n$. 

% Suppose $x_{(k)}$ is the $t_k$th tuple on the sample path of worker $n$, 
Consider the Markov chain formed by $m+1$ samples in $\{x_{(k-(K_0+1)m)},x_{(k-(K_0+1)m+1)},...,x_{(k)}\}$:
\begin{align*}
    % s_{t_k-m} \xrightarrow{\theta_{k-d_m}} a_{t_k-m} \xrightarrow{\mathcal{P}} s_{t_k-m+1} \xrightarrow{\theta_{k-d_{m-1}}} a_{t_k-m+1} \cdots s_{t_k} \xrightarrow{\theta_{k-d_0}} a_{t_k} \xrightarrow{\mathcal{P}} s_{t_k+1},
        s_{t-m} \xrightarrow{\theta_{k-d_m}} a_{t-m} \xrightarrow{\mathcal{P}} s_{t-m+1} \xrightarrow{\theta_{k-d_{m-1}}} a_{t-m+1} \cdots s_{t-1} \xrightarrow{\theta_{k-d_1}} a_{t-1}  \xrightarrow{\mathcal{P}} s_{t} \xrightarrow{\theta_{k-d_0}} a_{t} \xrightarrow{\mathcal{P}} s_{t+1},
\end{align*}
where $(s_t,a_t,s_{t+1})=(s_{(k)},a_{(k)},s'_{(k)})$, and $\{d_j\}_{j=0}^{m}$ is some increasing sequence with $d_0 \coloneqq \tau_k$.

Suppose $\theta_{k-d_m}$ was used to generate the $k_m$th update, then we have $x_{t-m}=x_{(k_m)}$. Following Assumption \ref{assumption:delay}, we have $\tau_{k_m}=k_m-(k- d_m) \leq K_0$. Since $x_{(k_m)}$ is in $\{x_{(k-(K_0+1)m)},...,x_{(k)}\}$, we have $k_m \geq  k-(K_0+1)m$.  Combining these two inequalities, we have
\begin{align}\label{eq:M_K-derivation}
    d_m\leq (K_0+1)m+K_0. 
\end{align} 
Given $(s_{t-m}, a_{t-m},s_{t-m+1})$ and $\theta_{k-d_m}$, we construct an auxiliary Markov chain as Lemma \ref{lemma:auxiliary-chain}:
\begin{align*}
    % s_{t_k-m} \xrightarrow{\theta_{k-d_m}} a_{t_k-m} \xrightarrow{\mathcal{P}} s_{t_k-m+1} \xrightarrow{\theta_{k-d_m}} \widetilde{a}_{t_k-m+1} \cdots \widetilde{s}_{t_k} \xrightarrow{\theta_{k-d_m}} \widetilde{a}_{t_k} \xrightarrow{\mathcal{P}} \widetilde{s}_{t_k+1}.
    s_{t-m} \xrightarrow{\theta_{k-d_m}} a_{t-m} \xrightarrow{\mathcal{P}} s_{t-m+1} \xrightarrow{\theta_{k-d_m}} \widetilde{a}_{t-m+1} \cdots \widetilde{s}_{t-1} \xrightarrow{\theta_{k-d_m}} \widetilde{a}_{t-1}\xrightarrow{\mathcal{P}} \widetilde{s}_t \xrightarrow{\theta_{k-d_m}} \widetilde{a}_{t} \xrightarrow{\mathcal{P}} \widetilde{s}_{t+1}.
\end{align*}
For brevity, we define
\begin{align*}
    \Delta_1(x,\theta,\omega) \coloneqq \left\langle \omega-\omega_\theta^*,g(x,\omega)-\overline{g}(\theta,\omega) \right\rangle.
\end{align*}

Throughout this proof, we use $\theta$, $\theta'$, $\omega$, $\omega'$, $x$ and $\widetilde{x}$ as shorthand notations of $\theta_k$, $\theta_{k-d_m}$, $\omega_k$, $\omega_{k-d_m}$, $x_t$ and $\widetilde{x}_t \coloneqq (\widetilde{s}_t,\widetilde{a}_t,\widetilde{s}_{t+1})$, respectively.

First we decompose $\Delta_1(x,\theta,\omega)$ as
\begin{align*}\label{eq:tmp0-lemma4}
    \Delta_1(x,\theta,\omega)
    &= \underbracket{\Delta_1(x,\theta,\omega)-\Delta_1(x,\theta',\omega)}_{I_1}+\underbracket{\Delta_1(x,\theta',\omega)-\Delta_1(x,\theta',\omega')}_{I_2} \\&~~~~+ \underbracket{\Delta_1(x,\theta',\omega')-\Delta_1(\widetilde{x},\theta',\omega')}_{I_3} + \underbracket{\Delta_1(\widetilde{x},\theta',\omega')}_{I_4}. \numberthis
\end{align*}
We bound $I_1$ in \eqref{eq:tmp0-lemma4} as
\begin{align*}\label{eq:support-noise-T1}
    \Delta_1(x,\theta,\omega)-\Delta_1(x,\theta',\omega)
    &= \left\langle \omega-\omega_\theta^*,g(x,\omega)-\overline{g}(\theta,\omega)\right\rangle - \left\langle \omega-\omega_{\theta'}^*, g(x,\omega)-\overline{g}(\theta',\omega)\right\rangle   \\
    &\leq \left| \left\langle \omega-\omega_\theta^*,g(x,\omega)-\overline{g}(\theta,\omega)\right\rangle - \left\langle \omega-\omega_{\theta'}^*,g(x,\omega)-\overline{g}(\theta,\omega)\right\rangle \right| \\&~~~~+ \left|\left\langle \omega-\omega_{\theta'}^*,g(x,\omega)-\overline{g}(\theta,\omega)\right\rangle - \left\langle \omega-\omega_{\theta'}^*,g(x,\omega)-\overline{g}(\theta',\omega)\right\rangle \right|. \numberthis
\end{align*}
For the first term in (\ref{eq:support-noise-T1}), we have
\begin{align*}
    \left| \left\langle \omega-\omega_\theta^*,g(x,\omega)-\overline{g}(\theta,\omega)\right\rangle - \left\langle \omega-\omega_{\theta'}^*,g(x,\omega)-\overline{g}(\theta,\omega)\right\rangle \right| 
    &=\left| \left\langle \omega_\theta^*-\omega_{\theta'}^*,g(x,\omega)-\overline{g}(\theta,\omega)\right\rangle\right| \\
    &\leq \| \omega_\theta^*-\omega_{\theta'}^*\|_2 \|g(x,\omega)-\overline{g}(\theta,\omega)\| \\
    &\leq 2C_\delta \|\omega_\theta^*-\omega_{\theta'}^*\|_2 \\
    &\leq 2 C_\delta L_{\omega}\|\theta-\theta'\|_2,\numberthis
\end{align*}
where the last inequality is due to Proposition \ref{proposition:omega-lipschitz}. 

We use $x \sim \theta'$ as shorthand notations to represent that $s \sim \mu_{\theta'}$, $a \sim \pi_{\theta'}$, $s' \sim \mathcal{P}$. For the second term in (\ref{eq:support-noise-T1}), we have
\begin{align*}
    &~~~~~\left|\left\langle \omega-\omega_{\theta'}^*,g(x,\omega)-\overline{g}(\theta,\omega)\right\rangle - \left\langle \omega-\omega_{\theta'}^*,g(x,\omega)-\overline{g}(\theta',\omega)\right\rangle \right| \\
    &=\left|\left\langle \omega-\omega_{\theta'}^*,\overline{g}(\theta',\omega)-\overline{g}(\theta,\omega)\right\rangle \right|\\
    &\leq\| \omega-\omega_{\theta'}^*\|_2 \|\overline{g}(\theta',\omega)-\overline{g}(\theta,\omega)\|_2\\
    &\leq 2R_\omega \|\overline{g}(\theta',\omega)-\overline{g}(\theta,\omega)\|_2\\
    % &\leq 2R_\omega \|\overline{g}(\theta',\omega)-\overline{g}(\theta,\omega)\|_2 \\\
    &= 2R_\omega \left\|\E_{x\sim\theta'}\left[g(x,\omega)\right]-\E_{x\sim\theta}\left[g(x,\omega)\right]\right\|_2 \\
    &\leq 2R_\omega \sup_{x}\|g(x,\omega)\|_2 \|\mu_{\theta'} \otimes \pi_{\theta'} \otimes \mathcal{P} - \mu_{\theta} \otimes \pi_{\theta} \otimes \mathcal{P}\|_{TV} \\
    &\leq 2R_\omega C_\delta \|\mu_{\theta'} \otimes \pi_{\theta'} \otimes \mathcal{P} - \mu_{\theta} \otimes \pi_{\theta} \otimes \mathcal{P}\|_{TV} \\
    &= 4R_\omega C_\delta d_{TV} \left(\mu_{\theta'} \otimes \pi_{\theta'} \otimes \mathcal{P}, \mu_{\theta} \otimes \pi_{\theta} \otimes \mathcal{P} \right) \\
    &\leq 4R_\omega C_\delta |\mathcal{A}|L_\pi (1+\log_\rho \kappa^{-1} + (1-\rho)^{-1}) \|\theta-\theta'\|_2,\numberthis
\end{align*}
where the third inequality follows the definition of TV norm, the second last inequality follows \eqref{eq:C_delta}, and the last inequality follows \cite[Lemma A.1]{wu2020finite}.

Collecting the upper bounds of the two terms in (\ref{eq:support-noise-T1}) yields
\begin{align*}
    I_1
    &\leq \left[2 C_\delta L_{\omega} + 4R_\omega C_\delta |\mathcal{A}| L_\pi (1+\log_\rho \kappa^{-1} + (1-\rho)^{-1})\right] \|\theta-\theta'\|_2.\numberthis
\end{align*}
Next we bound $\E[I_2]$ in \eqref{eq:tmp0-lemma4} as
\begin{align*}\label{eq:support-noise-T2}
    \E[I_2]
    &=\E[\Delta_1(x,\theta',\omega)-\Delta_1(x,\theta',\omega')] \\
    &= \E\left\langle \omega-\omega_{\theta'}^*, g(x,\omega)-\overline{g}(\theta',\omega) \right\rangle - \left\langle \omega'-\omega_{\theta'}^*, g(x,\omega')-\overline{g}(\theta',\omega') \right\rangle  \\
    &\leq \E\left| \left\langle \omega-\omega_{\theta'}^*,g(x,\omega)-\overline{g}(\theta',\omega) \right\rangle - \left\langle \omega-\omega_{\theta'}^*, g(x,\omega')-\overline{g}(\theta',\omega') \right\rangle  \right| \\&~~~~+ \E\left| \left\langle \omega-\omega_{\theta'}^*,g(x,\omega')-\overline{g}(\theta',\omega') \right\rangle - \left\langle \omega'-\omega_{\theta'}^*, g(x,\omega')-\overline{g}(\theta',\omega') \right\rangle \right|. \numberthis
\end{align*}
We bound the first term in (\ref{eq:support-noise-T2}) as
\begin{align*}
    &\E\left| \left\langle \omega-\omega_{\theta'}^*,g(x,\omega)-\overline{g}(\theta',\omega) \right\rangle - \left\langle \omega-\omega_{\theta'}^*, g(x,\omega')-\overline{g}(\theta',\omega') \right\rangle  \right| \\
    &= \E\left| \left\langle \omega-\omega_{\theta'}^*,g(x,\omega)-g(x,\omega')+ \overline{g}(\theta',\omega')-\overline{g}(\theta',\omega) \right\rangle \right| \\
    &\leq 2R_\omega \left(\E\|g(x,\omega)-g(x,\omega')\|_2+ \E\|\overline{g}(\theta',\omega')-\overline{g}(\theta',\omega)\|_2 \right) \\
    &\leq 2R_\omega \left(\E\|g(x,\omega)-g(x,\omega')\|_2+ \E\left\|\E_{x \sim \theta'}[g(x,\omega')]-\E_{x \sim \theta'}[g(x,\omega)]\right\|_2 \right) \\
    &= 2R_\omega \left(\E\|(\gamma \phi(s')-\phi(s))^\top (\omega-\omega')\|_2+ \E\left\|\E_{x \sim \theta'}\left[(\gamma \phi(s')-\phi(s))^\top\right]  (\omega'-\omega)\right\|_2\right) \\
    &\leq 2R_\omega \left( (1+\gamma)\E\|\omega-\omega'\|_2 + (1+\gamma)\E\|\omega-\omega'\|_2\right) \\
    &= 4 R_\omega (1+\gamma) \E\|\omega-\omega'\|_2.\numberthis
\end{align*}
We bound the second term in (\ref{eq:support-noise-T2}) as
\begin{align*}
    &\E\left| \left\langle \omega-\omega_{\theta'}^*,g(x,\omega')-\overline{g}(\theta',\omega') \right\rangle - \left\langle \omega'-\omega_{\theta'}^*, g(x,\omega')-\overline{g}(\theta',\omega') \right\rangle \right| \\
    &= \E\left|\left\langle \omega-\omega', g(x,\omega')-\overline{g}(\theta',\omega')\right\rangle \right|\\
    &\leq 2 C_\delta \E\|\omega-\omega'\|_2.\numberthis
\end{align*}
Collecting the upper bounds of the two terms in  (\ref{eq:support-noise-T2}) yields
\begin{equation}
    \E[I_2] \leq \left( 4(1+\gamma)R_\omega + 2 C_\delta \right) \E\|  \omega_k-\omega_{k-d_m} \|_2.
\end{equation}
We first bound $I_3$ as
\begin{align*}\label{eq:EI_3}
    \E[I_3| \theta', \omega', s_{t-m+1}]
    &=\E\left[\Delta_1(x,\theta',\omega')-\Delta_1(\widetilde{x},\theta',\omega')| \theta',\omega',  s_{t-m+1}\right] \\
    &\leq \left| \E\left[\Delta_1(x,\theta',\omega')| \theta',\omega',  s_{t-m+1}\right]-\E\left[\Delta_1(\widetilde{x},\theta',\omega')| \theta', \omega', s_{t-m+1}\right]\right|\\
    &\leq \sup_{x}| \Delta_1(x,\theta',\omega') | \left\|\mathbb{P}(x \in \cdot | \theta', \omega', s_{t-m+1}) - \mathbb{P}(\widetilde{x} \in \cdot | \theta', \omega', s_{t-m+1}) \right\|_{TV} \\
    &\leq 8R_\omega C_\delta d_{TV}\left( \mathbb{P}(x \in \cdot | \theta', s_{t-m+1}), \mathbb{P}(\widetilde{x} \in \cdot | \theta', s_{t-m+1}) \right), \numberthis
\end{align*}
where the second last inequality follows the definition of TV norm, and the last inequality follows 
\begin{align*}
    | \Delta_1(x,\theta',\omega') | \leq \|\omega'-\omega_{\theta'}^*\|_2 \|g(x,\omega')-\overline{g}(\theta',\omega')\|_2 \leq 4R_\omega C_\delta.\numberthis
\end{align*}

By following (\ref{eq:auxiliary-markov-chain-4}) in Lemma \ref{lemma:auxiliary-chain}, we have
\begin{align*}
    d_{TV}\left( \mathbb{P}(x \in \cdot | \theta', s_{t-m+1}), \mathbb{P}(\widetilde{x} \in \cdot | \theta', s_{t-m+1}) \right) 
    &\leq \frac{1}{2} |\mathcal{A}| L_\pi \sum_{i=\tau_k}^{d_m}\E\left.\left[\|\theta_{k-i}-\theta_{k-d_m}\|_2 \right| \theta', s_{t-m+1}\right].
\end{align*}
Substituting the last inequality into \eqref{eq:EI_3}, then taking total expectation on both sides yield
\begin{equation}
    \E[I_3] \leq 4R_\omega C_\delta |\mathcal{A}|L_\pi \sum_{i=\tau_k}^{d_m}\E\|\theta_{k-i}-\theta_{k-d_m}\|_2.
\end{equation}
Next we bound $I_4$. Define $\overline{x} \coloneqq (\overline{s},\overline{a},\overline{s}')$ where $\overline{s} \sim \mu_{\theta'}$, $\overline{a} \sim \pi_{\theta'}$ and $\overline{s}' \sim \mathcal{P}$. It is immediate that
\begin{align*}\label{eq:delta_1-upper-bound}
    \E[\Delta_1(\overline{x}, \theta', \omega') | \theta', \omega', s_{t-m+1}]
    &= \left\langle \omega'-\omega_{\theta'}^*, \E[g(\overline{x}, \omega')| \theta', \omega', s_{t-m+1}]-\overline{g}(\theta',\omega') \right\rangle \\
    &= \left\langle \omega'-\omega_{\theta'}^*, \overline{g}(\theta',\omega')-\overline{g}(\theta',\omega') \right\rangle = 0 . \numberthis
\end{align*}
Then we have
\begin{align*}\label{EI_4}
    \E[I_4 | \theta', \omega', s_{t-m+1}] 
    &= \E\left[ \Delta_1(\widetilde{x},\theta',\omega')-\Delta_1(\overline{x}, \theta', \omega') | \theta', \omega', s_{t-m+1}\right] \\
    &\leq  \left| \E\left[ \Delta_1(\widetilde{x},\theta',\omega') | \theta', \omega', s_{t-m+1}\right] -\E\left[\Delta_1(\overline{x}, \theta', \omega')| \theta', \omega', s_{t-m+1}\right] \right| \\
    &\leq \sup_{x}| \Delta_1(x,\theta',\omega') | \left\| \mathbb{P}(\widetilde{x} \in \cdot | \theta',s_{t-m+1}) - \mathbb{P}(\overline{x} \in \cdot | \theta',s_{t-m+1}) \right\|_{TV} \\
    &\leq 8R_\omega C_\delta d_{TV} \left( \mathbb{P}(\widetilde{x} \in \cdot | \theta',s_{t-m+1}),\mathbb{P}(\overline{x} \in \cdot | \theta',s_{t-m+1}) \right) \\
    &= 8R_\omega C_\delta d_{TV} \left( \mathbb{P}(\widetilde{x} \in \cdot | \theta',s_{t-m+1}), \mu_{\theta'} \otimes \pi_{\theta'} \otimes \mathcal{P} \right), \numberthis
\end{align*}
where the second inequality follows the definition of TV norm, and the third inequality follows \eqref{eq:delta_1-upper-bound}. 

The auxiliary Markov chain with policy $\pi_{\theta'}$ starts from initial state $s_{t-m+1}$, and $\widetilde{s}_t$ is the $(m-1)$th state on the chain. Following Lemma \ref{lemma:geom-s-a-s'}, we have:
\begin{align*}
    &d_{TV} \left( \mathbb{P}(\widetilde{x} \in \cdot | \theta',s_{t-m+1}), \mu_{\theta'} \otimes\pi_{\theta'} \otimes \mathcal{P} \right) \\
    &= d_{TV} \left( \mathbb{P}\left((\widetilde{s}_t,\widetilde{a}_t,\widetilde{s}_{t+1}) \in \cdot | \theta',s_{t-m+1}\right), \mu_{\theta'} \otimes\pi_{\theta'} \otimes \mathcal{P} \right)
    \leq  \kappa \rho^{m-1}.\numberthis
\end{align*}
Substituting the last inequality into \eqref{EI_4} and taking total expectation on both sides yield
\begin{align*}
    \E[I_4] \leq 8 R_\omega C_\delta \kappa \rho^{m-1}. \numberthis
\end{align*}
Taking total expectation on (\ref{eq:tmp0-lemma4}) and collecting bounds of $I_1$, $I_2$, $I_3$, $I_4$ yield
\begin{align*}
\E\left[\Delta_1(x,\theta,\omega)\right] 
&\leq C_4 \E\|\theta_k - \theta_{k-d_m}\|_2 + C_5 \sum_{i=\tau_k}^{d_m}\E\|\theta_{k-i}-\theta_{k-d_m}\|_2 \\&~~~~+ C_6 \E\| \omega_k-\omega_{k-d_m} \|_2 + C_7 \kappa \rho^{m-1} , \numberthis
\end{align*}
where $C_4 \coloneqq 2 C_\delta L_{\omega} + 4R_\omega C_\delta |\mathcal{A}| L_\pi (1+\log_\rho \kappa^{-1} + (1-\rho)^{-1})$, $C_5 \coloneqq 4R_\omega C_\delta |\mathcal{A}|L_\pi$, $C_6 \coloneqq  4(1+\gamma)R_\omega + 2 C_\delta$ and $C_7 \coloneqq 8R_\omega C_\delta$.
\end{proof}

\subsection{Supporting Lemmas for Theorem \ref{theorem:async-tts-actor}}
\begin{Lemma}\label{lemma:tts-epsilon}
For any $m \geq 1$ and $k \geq (K_0+1)m+K_0+1$, we have
\begin{align*}
&    \E\left\langle \nabla J_\lambda(\theta_k), \left( \hat{\delta}(\hat{x}_{(k)},\omega_k^*)-\delta(\hat{x}_{(k)},\theta_k)\right)\psi_{\theta_{k-\tau_k}}(\hat{s}_{(k)},\hat{a}_{(k)}) \right\rangle  \geq -D_2 \E\|\theta_{k-\tau_k}-\theta_{k-d_m}\|_2\\
    &~~~~~~~~~~ - D_3 \E\|\theta_k - \theta_{k-d_m}\|_2- D_4 \sum_{i=\tau_k}^{d_m}\E\|\theta_{k-i}-\theta_{k-d_m}\|_2  - D_5 \kappa \rho^{m-1} - L_V C_\psi(1+\gamma)\epsilon_{\rm app},
\end{align*}
where $D_2 \coloneqq 2 L_V L_\psi C_\delta$, $D_3 \coloneqq \left( 2C_\delta C_\psi L_{\lambda} + L_V C_\psi(L_{\omega}+L_V)(1+\gamma) \right)$, $D_4 \coloneqq 2 L_V C_\psi C_\delta |\mathcal{A}|L_\pi$ and $D_5 \coloneqq 4L_V C_\psi C_\delta$.
\end{Lemma}
\begin{proof}
For the worker that contributes to the $k$th update, we construct its Markov chain:
\begin{align*}
    \hat{s}_{t-m} \xrightarrow{\theta_{k-d_m}} \hat{a}_{t-m} \xrightarrow{\hat{\mathcal{P}}} \hat{s}_{t-m+1} \xrightarrow{\theta_{k-d_{m-1}}} \hat{a}_{t-m+1} \cdots \hat{s}_{t-1} \xrightarrow{\theta_{k-d_1}} \hat{a}_{t-1}  \xrightarrow{\hat{\mathcal{P}}} \hat{s}_{t} \xrightarrow{\theta_{k-d_0}} a_{t} \xrightarrow{\hat{\mathcal{P}}} \hat{s}_{t+1},
\end{align*}
where $(s_t,a_t,\hat{s}_{t+1})=(\hat{s}_{(k)},\hat{a}_{(k)},s'_{(k)})$, and $\{d_j\}_{j=0}^{m}$ is some increasing sequence with $d_0 \coloneqq \tau_k$.
By (\ref{eq:M_K-derivation}) in Lemma \ref{lemma:tts-critic-noise}, we have $d_m \leq (K_0+1)m+K_0$. 

Given $(\hat{s}_{t-m}, \hat{a}_{t-m},\hat{s}_{t-m+1})$ and $\theta_{k-d_m}$, we construct an auxiliary Markov chain:
\begin{align*}
    \hat{s}_{t-m} \xrightarrow{\theta_{k-d_m}} \hat{a}_{t-m} \xrightarrow{\hat{\mathcal{P}}} \hat{s}_{t-m+1} \xrightarrow{\theta_{k-d_m}} \widetilde{a}_{t-m+1} \cdots \widetilde{s}_{t-1} \xrightarrow{\theta_{k-d_m}} \widetilde{a}_{t-1}\xrightarrow{\hat{\mathcal{P}}} \widetilde{s}_{t} \xrightarrow{\theta_{k-d_m}} \widetilde{a}_{t} \xrightarrow{\hat{\mathcal{P}}} \widetilde{s}_{t+1}.
\end{align*}
% For brevity, we define $\Delta_2(x,\theta) \coloneqq \gamma \left(\phi(s')^\top \omega_\theta^* - V_{\pi_\theta}(s')\right)+V_{\pi_\theta}(s)-\phi(s)^\top \omega_\theta^*$. Throughout this proof, we use $\theta$, $\theta'$, $\omega_\theta^*$, $\omega_{\theta'}^*$ and $x$ as shorthand notations of $\theta_k$, $\theta_{k-d_m}$, $\omega_k^*$, $\omega_{k-d_m}^*$ and $\hat{x}_{(k)}$ respectively. We also define $\overline{x} \coloneqq (\overline{s},\overline{a},\overline{s}')$, where $\overline{s} \sim \mu_{\theta'}$, $\overline{a} \sim \pi_{\theta'}$ and $\overline{s}' \sim \hat{\mathcal{P}}$.
% where we have $\|\delta(x,\theta)\|_2 \leq r_{\max} + (1+\gamma)r_{\max} \leq C_{\delta}$ for any $x$ and $\theta$.
First we have
\begin{align*}\label{eq:tmp0-lemma5}
    &\left\langle \nabla J_\lambda(\theta_k), \left( \hat{\delta}(\hat{x}_{(k)},\omega_k^*)-\delta(\hat{x}_{(k)},\theta_k)\right)\psi_{\theta_{k-\tau_k}}(\hat{s}_{(k)},\hat{a}_{(k)}) \right\rangle \\
    &= \left\langle \nabla J_\lambda(\theta_k), \left( \hat{\delta}(\hat{x}_{(k)},\omega_k^*)-\delta(\hat{x}_{(k)},\theta_k)\right)\left( \psi_{\theta_{k-\tau_k}}(\hat{s}_{(k)},\hat{a}_{(k)})-\psi_{\theta_{k-d_m}}(\hat{s}_{(k)},\hat{a}_{(k)})\right) \right\rangle \\
    &~~~~+ \left\langle \nabla J_\lambda(\theta_k), \left( \hat{\delta}(\hat{x}_{(k)},\omega_k^*)-\delta(\hat{x}_{(k)},\theta_k)\right)\psi_{\theta_{k-d_m}}(\hat{s}_{(k)},\hat{a}_{(k)}) \right\rangle. \numberthis
\end{align*}
We first bound the fist term in \eqref{eq:tmp0-lemma5} as
\begin{align*}\label{eq:tmp0_1-lemma5}
    &\left\langle \nabla J_\lambda(\theta_k), \left( \hat{\delta}(\hat{x}_{(k)},\omega_k^*)-\delta(\hat{x}_{(k)},\theta_k)\right)\left( \psi_{\theta_{k-\tau_k}}(\hat{s}_{(k)},\hat{a}_{(k)})-\psi_{\theta_{k-d_m}}(\hat{s}_{(k)},\hat{a}_{(k)})\right) \right\rangle \\
    &\geq -\|J_\lambda(\theta_k)\|_2 |\hat{\delta}(\hat{x}_{(k)},\omega_k^*)-\delta(\hat{x}_{(k)},\theta_k)| \| \psi_{\theta_{k-\tau_k}}(\hat{s}_{(k)},\hat{a}_{(k)})-\psi_{\theta_{k-d_m}}(\hat{s}_{(k)},\hat{a}_{(k)})\|_2 \\
    &\geq -\|J_\lambda(\theta_k)\|_2 \left(|\hat{\delta}(\hat{x}_{(k)},\omega_k^*)|+|\delta(\hat{x}_{(k)},\theta_k)|\right) \| \psi_{\theta_{k-\tau_k}}(\hat{s}_{(k)},\hat{a}_{(k)})-\psi_{\theta_{k-d_m}}(\hat{s}_{(k)},\hat{a}_{(k)})\|_2 \\
    &\geq -L_V \left(|\hat{\delta}(\hat{x}_{(k)},\omega_k^*)|+|\delta(\hat{x}_{(k)},\theta_k)|\right) \| \psi_{\theta_{k-\tau_k}}(\hat{s}_{(k)},\hat{a}_{(k)})-\psi_{\theta_{k-d_m}}(\hat{s}_{(k)},\hat{a}_{(k)})\|_2 \\
    &\geq -2 L_V C_\delta \|\psi_{\theta_{k-\tau_k}}(\hat{s}_{(k)},\hat{a}_{(k)})-\psi_{\theta_{k-d_m}}(\hat{s}_{(k)},\hat{a}_{(k)})\|_2\\
    &\geq -2 L_V L_\psi C_\delta \|\theta_{k-\tau_k}-\theta_{k-d_m}\|_2, \numberthis
\end{align*}
where the last inequality follows Assumption \ref{assumption:omega} and second last inequality follows
% \begin{subequations}
\begin{align*}
    % \label{eq:delta(x,\omega_\theta^*)}
    |\hat{\delta}(x,\omega_\theta^*)|
    \leq |r(x)|+\gamma \|\phi(s')\|_2 \|\omega_\theta^*\|_2 +\|\phi(s)\|_2 \|\omega_\theta^*\|_2 
    \leq r_{\max}+(1+\gamma)R_\omega \leq C_\delta, \\
    % \label{eq:delta'(x,\theta)}
    |\delta(x,\theta)|
    \leq |r(x)| + \gamma |V_{\pi_{\theta}}(s')| +  |V_{\pi_{\theta}}(s)|
    \leq r_{\max} + (1+\gamma)\frac{r_{\max}}{1-\gamma} \leq C_{\delta}.
\end{align*}
% \end{subequations}
Substituting \eqref{eq:tmp0_1-lemma5} into \eqref{eq:tmp0-lemma5} gives
\begin{align*}\label{eq:tmp1-lemma5}
    &\left\langle \nabla J_\lambda(\theta_k), \left( \hat{\delta}(\hat{x}_{(k)},\omega_k^*)-\delta(\hat{x}_{(k)},\theta_k)\right)\psi_{\theta_{k-\tau_k}}(\hat{s}_{(k)},\hat{a}_{(k)}) \right\rangle \\
    &\geq -2 L_V L_\psi C_\delta \|\theta_{k-\tau_k}-\theta_{k-d_m}\|_2 + \left\langle \nabla J_\lambda(\theta_k), \left( \hat{\delta}(\hat{x}_{(k)},\omega_k^*)-\delta(\hat{x}_{(k)},\theta_k)\right)\psi_{\theta_{k-d_m}}(\hat{s}_{(k)},\hat{a}_{(k)}) \right\rangle. \numberthis
\end{align*}
Then we start to bound the second term in \eqref{eq:tmp1-lemma5}. For brevity, we define
\begin{align*}
    \Delta_2(x,\theta) \coloneqq \left\langle \nabla J_\lambda(\theta), \left(\hat{\delta}(x,\omega_\theta^*)-\delta(x,\theta)\right) \psi_{\theta_{k-d_m}}(s,a) \right\rangle.
\end{align*}

In the following proof, 
we use $\theta$, $\theta'$, $\omega_\theta^*$, $\omega_{\theta'}^*$, $x$ as shorthand notations for $\theta_k$, $\theta_{k-d_m}$, $\omega_k^*$, $\omega_{k-d_m}^*$, $x_t$ respectively. We also define $\overline{x} \coloneqq (\overline{s},\overline{a},\overline{s}')$, where $\overline{s} \sim d_{\theta'}$, $\overline{a} \sim \pi_{\theta'}$ and $\overline{s}' \sim \mathcal{P}$. Define $\widetilde{x}\coloneqq (\widetilde{s}_{t-d_m},\widetilde{a}_{t-d_m},\widetilde{s}'_{t-d_m})$ where $\widetilde{s}_{t-d_m},\widetilde{a}_{t-d_m}$ are state and action on the auxiliary Markov chain, and $\widetilde{s}'_{t-d_m}\sim \mathcal{P}(\cdot|\widetilde{s}_{t-d_m},\widetilde{a}_{t-d_m})$ is a virtual sample. We decompose the second term in (\ref{eq:tmp1-lemma5}) as
\begin{align*}
    \Delta_2(x,\theta) = \underbracket{\Delta_2(x,\theta) - \Delta_2(x,\theta')}_{I_1} + \underbracket{\Delta_2(x,\theta') - \Delta_2(\widetilde{x},\theta')}_{I_2} + \underbracket{\Delta_2(\widetilde{x},\theta') - \Delta_2(\overline{x},\theta')}_{I_3} + \underbracket{\Delta_2(\overline{x},\theta')}_{I_4}.
\end{align*}
We bound the term $I_1$ as
\begin{align*}
    I_1 
    &= \left\langle \nabla J_\lambda(\theta), \left(\hat{\delta}(x,\omega_\theta^*)-\delta(x,\theta)\right)\psi_{\theta'}(s,a) \right\rangle 
    - \left\langle \nabla J_\lambda(\theta'), \left(\hat{\delta}(x,\omega_{\theta'}^*)-\delta(x,\theta')\right)\psi_{\theta'}(s,a) \right\rangle \\
    &= \left\langle \nabla J_\lambda(\theta), \left(\hat{\delta}(x,\omega_\theta^*)-\delta(x,\theta)\right)\psi_{\theta'}(s,a) \right\rangle 
    - \left\langle \nabla J_\lambda(\theta'), \left(\hat{\delta}(x,\omega_\theta^*)-\delta(x,\theta)\right)\psi_{\theta'}(s,a) \right\rangle \\
    &~~~~+ \left\langle \nabla J_\lambda(\theta'), \left(\hat{\delta}(x,\omega_\theta^*)-\delta(x,\theta)\right)\psi_{\theta'}(s,a) \right\rangle 
    - \left\langle \nabla J_\lambda(\theta'), \left(\hat{\delta}(x,\omega_{\theta'}^*)-\delta(x,\theta')\right)\psi_{\theta'}(s,a) \right\rangle. \numberthis
\end{align*}
For the first term in $I_1$, we have
\begin{align*}
    &\left\langle \nabla J_\lambda(\theta), \left(\hat{\delta}(x,\omega_\theta^*)-\delta(x,\theta)\right)\psi_{\theta'}(s,a) \right\rangle - \left\langle \nabla J_\lambda(\theta'), \left(\hat{\delta}(x,\omega_\theta^*)-\delta(x,\theta)\right)\psi_{\theta'}(s,a) \right\rangle \\
    &= \left\langle \nabla J_\lambda(\theta)- \nabla J_\lambda(\theta'), \left(\hat{\delta}(x,\omega_\theta^*)-\delta(x,\theta)\right)\psi_{\theta'}(s,a) \right\rangle \\
    &\geq -\|\nabla J_\lambda(\theta)- \nabla J_\lambda(\theta')\|_2 \|\hat{\delta}(x,\omega_\theta^*)-\delta(x,\theta)\|_2 \|\psi_{\theta'}(s,a)\|_2 \\
    &\geq -2C_\delta C_\psi \|\nabla J_\lambda(\theta)-\nabla J_\lambda(\theta')\|_2 \\
    &\geq -2 C_\delta C_\psi L_{\lambda} \|\theta-\theta'\|_2,\numberthis
\end{align*}
where the last inequality is due to the $L_{\lambda}$-Lipschitz continuity of policy gradient shown in Proposition \ref{prop:Lj-lip}.

For the second term in $I_1$, we have
\begin{align*}
    &\left\langle \nabla J_\lambda(\theta'), \left(\hat{\delta}(x,\omega_\theta^*)-\delta(x,\theta)\right)\psi_{\theta'}(s,a) \right\rangle - \left\langle \nabla J_\lambda(\theta'), \left(\hat{\delta}(x,\omega_{\theta'}^*)-\delta(x,\theta')\right)\psi_{\theta'}(s,a) \right\rangle \\
    &= \left\langle \nabla J_\lambda(\theta'), \left(\hat{\delta}(x,\omega_\theta^*)-\hat{\delta}(x,\omega_{\theta'}^*)+\delta(x,\theta')-\delta(x,\theta)\right)\psi_{\theta'}(s,a) \right\rangle \\
    &\geq -L_V C_\psi \left| \hat{\delta}(x,\omega_\theta^*)-\hat{\delta}(x,\omega_{\theta'}^*)+\delta(x,\theta')-\delta(x,\theta) \right| \\
    &\geq -L_V C_\psi\left| \gamma \phi(s')^\top(\omega_\theta^*-\omega_{\theta'}^*) + \phi(s)^\top(\omega_{\theta'}^*-\omega_\theta^*) + \gamma V_{\pi_{\theta'}}(s')-\gamma V_{\pi_\theta}(s') + V_{\pi_\theta}(s) -  V_{\pi_{\theta'}}(s) \right| \\
    &\geq -L_V C_\psi \left(\gamma\|\omega_\theta^*-\omega_{\theta'}^*\|_2 + \|\omega_{\theta'}^*-\omega_\theta^*\|_2 + \gamma |V_{\pi_{\theta'}}(s')-V_{\pi_\theta}(s')| + |V_{\pi_\theta}(s) -  V_{\pi_{\theta'}}(s)| \right) \\
    &\geq -L_V C_\psi \left( \gamma L_{\omega}\|\theta-\theta'\|_2 + L_{\omega}\|\theta-\theta'\|_2 +  \gamma L_V\|\theta-\theta'\|_2 + L_V\|\theta-\theta'\|_2  \right)\\
    &= -L_V C_\psi(L_{\omega}+L_V)(1+\gamma)\|\theta-\theta'\|_2,\numberthis
\end{align*}
where the last inequality is due to the $L_{\omega}$-Lipschitz continuity of $\omega_\theta^*$ shown in Proposition \ref{proposition:omega-lipschitz} and $L_V$-Lipschitz continuity of $V_{\pi_\theta}(s)$. Collecting the upper bounds of $I_1$ yields
\begin{align*}
    I_1 \geq -\left( 2C_\delta C_\psi L_{\lambda} + L_V C_\psi(L_{\omega}+L_V)(1+\gamma) \right)\|\theta-\theta'\|_2.\numberthis
\end{align*}
First we bound $I_2$ as
\begin{align*}\label{eq:I_2}
    \E[I_2 | \theta',\hat{s}_{t-m+1}]
    &= \E\left[\Delta_2(x,\theta') - \Delta_2(\widetilde{x},\theta') | \theta',\hat{s}_{t-m+1}\right] \\
    &\geq -\left|\E\left.\left[\Delta_2(x,\theta') \right| \theta',\hat{s}_{t-m+1}\right] - \E\left.\left[\Delta_2(\widetilde{x},\theta') \right| \theta',\hat{s}_{t-m+1}\right] \right|\\
    &\geq -\sup_{x} |\Delta_2(x,\theta')| \left\| \mathbb{P}(x\in\cdot|\theta',\hat{s}_{t-m+1})-\mathbb{P}(\widetilde{x}\in\cdot|\theta',\hat{s}_{t-m+1}) \right\|_{TV}\\
    &\geq -4 L_V C_\psi C_\delta d_{TV}\left( \mathbb{P}(x\in\cdot|\theta',\hat{s}_{t-m+1}), \mathbb{P}(\widetilde{x}\in\cdot|\theta',\hat{s}_{t-m+1})\right) \\
    &=-4 L_V C_\psi C_\delta d_{TV}\left( \mathbb{P}((\hat{s}_t,\hat{a}_t)\in\cdot|\theta',\hat{s}_{t-m+1}), \mathbb{P}((\widetilde{s}_t,\widetilde{a}_t)\in\cdot|\theta',\hat{s}_{t-m+1})\right) \\
    &\geq -2 L_V C_\psi C_\delta |\mathcal{A}|L_\pi \sum_{i=\tau_k}^{d_m}\E\left.\left[\|\theta_{k-i}-\theta_{k-d_m}\|_2  \right| \theta',\hat{s}_{t-m+1} \right], \numberthis
\end{align*}
where the second inequality is due to the definition of TV norm, the last inequality follows (\ref{eq:auxiliary-markov-chain-4}) in Lemma \ref{lemma:auxiliary-chain}, and the second last inequality follows the fact that
\begin{align*}\label{eq:Delta_2bound}
    |\Delta_2(x,\theta')| 
    % = \left\langle \nabla J_\lambda(\theta'), \left(\hat{\delta}(x,\omega_\theta^*)-\delta(x,\theta')\right) \psi_{\theta'}(s,a) \right\rangle 
    \leq \|\nabla J_\lambda(\theta')\|_2 |\hat{\delta}(x,\omega_{\theta'}^*)-\delta(x,\theta')| \|\psi_{\theta'}(s,a)\|_2
    \leq 2L_V C_\delta C_\psi. \numberthis
\end{align*}
Taking total expectation on both sides of \eqref{eq:I_2} yields
\begin{align*}
    \E[I_2] &\geq -2 L_V C_\psi C_\delta |\mathcal{A}|L_\pi \sum_{i=\tau_k}^{d_m}\E\|\theta_{k-i}-\theta_{k-d_m}\|_2 .\numberthis
\end{align*}
Next we bound $I_3$ as
\begin{align*}\label{eq:EI_3-1}
    \E[I_3  | \theta',\hat{s}_{t-m+1}]
    &= \E\left.\left[\Delta_2(\widetilde{x},\theta') - \Delta_2(\overline{x},\theta') \right| \theta',\hat{s}_{t-m+1} \right] \\
    &\geq -\left|\E\left.\left[\Delta_2(\widetilde{x},\theta')\right| \theta',\hat{s}_{t-m+1} \right] - \E\left.\left[\Delta_2(\overline{x},\theta') \right| \theta',\hat{s}_{t-m+1} \right] \right| \\
    &\geq -\sup_{x} |\Delta_2(x,\theta')| \left\| \mathbb{P}(\widetilde{x}\in\cdot |\theta',\hat{s}_{t-m+1}) - \mathbb{P}(\overline{x}\in\cdot |\theta',\hat{s}_{t-m+1}) \right\|_{TV}\\
    &\geq  -4L_V C_\psi C_\delta d_{TV}\left( \mathbb{P}(\widetilde{x}\in\cdot |\theta',\hat{s}_{t-m+1}),d_{\theta'}\otimes \pi_{\theta'} \otimes \mathcal{P} \right), \numberthis
\end{align*}
where the second inequality is due to the definition of TV norm, and the last inequality follows \eqref{eq:Delta_2bound}.

The auxiliary Markov chain with policy $\pi_{\theta'}$ starts from initial state $\hat{s}_{t-m+1}$, and $\widetilde{s}_t$ is the $(m-1)$th state on the chain. Following Lemma \ref{lemma:geom-s-a-s'}, we have:
\begin{align*}
    d_{TV} \left( \mathbb{P}(\widetilde{x} \in \cdot | \theta',\hat{s}_{t-m+1}), d_{\theta'} \otimes\pi_{\theta'} \otimes \hat{\mathcal{P}} \right)
    &= d_{TV} \left( \mathbb{P}\left((\widetilde{s}_t,\widetilde{a}_t,\widetilde{s}_{t+1}) \in \cdot | \theta',\hat{s}_{t-m+1}\right), d_{\theta'} \otimes\pi_{\theta'} \otimes \mathcal{P} \right) \\
    &\leq  \kappa \rho^{m-1}.\numberthis
\end{align*}
Substituting the last inequality into \eqref{eq:EI_3-1} and taking total expectation on both sides yield
\begin{align*}
    \E[I_3] &\geq -4L_V C_\psi C_\delta \kappa \rho^{m-1}.\numberthis
\end{align*}
We bound $I_4$ as
\begin{align*}
    \E[I_4 | \theta']
    &= \E\left.\left[\left\langle \nabla J_\lambda(\theta'), \left(\hat{\delta}(\overline{x},\omega_{\theta'}^*)-\delta(\overline{x},\theta')\right) \psi_{\theta'}(s,a) \right\rangle  \right| \theta'\right]\\
    &\geq -L_V C_\psi \E \left.\left[\left|\hat{\delta}(\overline{x},\omega_{\theta'}^*)-\delta(\overline{x},\theta') \right| \right| \theta' \right] \\
    &= -L_V C_\psi \E\left.\left[\left|\gamma\left(\phi(\overline{s}')^\top \omega_{\theta'}^*-V_{\pi_{\theta'}}(\overline{s}') \right) + V_{\pi_{\theta'}}(\overline{s}) - \phi(\overline{s})^\top \omega_{\theta'}^* \right| \right| \theta'\right] \\
    &\geq -L_V C_\psi \left(\gamma \E\left.\left[|\phi(\overline{s}')^\top \omega_{\theta'}^*-V_{\pi_{\theta'}}(\overline{s}')| \right| \theta'\right]+ \E\left.\left[|V_{\pi_{\theta'}}(\overline{s}) - \phi(\overline{s})^\top \omega_{\theta'}^*| \right| \theta'\right]\right) \\
    &\geq -L_V C_\psi \left(\gamma \sqrt{\E\left.\left[|\phi(\overline{s}')^\top \omega_{\theta'}^*-V_{\pi_{\theta'}}(\overline{s}')|^2 \right| \theta' \right]} + \sqrt{\E\left.\left[|V_{\pi_{\theta'}}(\overline{s}) - \phi(\overline{s})^\top \omega_{\theta'}^*|^2\right| \theta' \right]} \right) \\
    &= -L_V C_\psi \left(\gamma \sqrt{\E_{\overline{s}' \sim \mu_{\theta'}}|\phi(\overline{s}')^\top \omega_{\theta'}^*-V_{\pi_{\theta'}}(\overline{s}')|^2 } + \sqrt{\E_{\overline{s} \sim \mu_{\theta'}}|V_{\pi_{\theta'}}(\overline{s}) - \phi(\overline{s})^\top \omega_{\theta'}^*|^2} \right) \\
    &\geq -L_V C_\psi(1+\gamma)\epsilon_{\rm app}, \numberthis
\end{align*}
where the second last inequality follows Jensen's inequality.

Taking total expectation on both sides of (\ref{eq:tmp1-lemma5}), and collecting lower bounds of $I_1$, $I_2$, $I_3$ and $I_4$ yield
\begin{align*}
    &\E\left\langle \nabla J_\lambda(\theta_k), \left( \hat{\delta}(\hat{x}_{(k)},\omega_k^*)-\delta(\hat{x}_{(k)},\theta_k)\right)\psi_{\theta_{k-\tau_k}}(\hat{s}_{(k)},\hat{a}_{(k)}) \right\rangle \\ &\geq -D_2 \E\|\theta_{k-\tau_k}-\theta_{k-d_m}\|_2 - D_3\E\|\theta_k - \theta_{k-d_m}\|_2 - D_4 \sum_{i=\tau_k}^{d_m}\E\|\theta_{k-i}-\theta_{k-d_m}\|_2\\
    &~~~~ - D_5 \kappa \rho^{m-1} - L_V C_\psi(1+\gamma)\epsilon_{\rm app}, \numberthis
\end{align*}
where $D_2 \coloneqq 2 L_V L_\psi C_\delta$, $D_3 \coloneqq \left( 2C_\delta C_\psi L_{\lambda} + L_V C_\psi(L_{\omega}+L_V)(1+\gamma) \right)$, $D_4 \coloneqq 2 L_V C_\psi C_\delta |\mathcal{A}|L_\pi$ and $D_5 \coloneqq 4L_V C_\psi C_\delta$.
\end{proof}

\begin{Lemma}\label{lemma:tts-actor-noise}
For any $m \geq 1$ and $k \geq (K_0+1)m+K_0+1$, we have
\begin{align*}
    &\E\left\langle \nabla J_\lambda(\theta_k), \delta(\hat{x}_{(k)},\theta_k)\psi_{\theta_{k-\tau_k}}(\hat{s}_{(k)},\hat{a}_{(k)}) - \nabla J(\theta_k) \right\rangle \\
    &\geq \!-\!D_6 \E\|\theta_{k-\tau_k} \!-\! \theta_{k-d_m}\|_2 \!-\! D_7 \E\|\theta_k\!-\!\theta_{k-d_m}\|_2
    \!-\! D_8 \sum_{i=\tau_k}^{d_m} \E\|\theta_{k-i}-\theta_{k-d_m}\|_2 \!-\! D_9 \kappa \rho^{m-1},
\end{align*}
where $d_m \leq (K_0+1)m+K_0$, $D_6 \coloneqq L_V C_\delta L_\psi$, $D_7 \coloneqq C_p L_{\lambda} +(1+\gamma)L_V^2 C_\psi + 2L_V L_{\lambda}$, $D_8 \coloneqq L_V C_p |\mathcal{A}| L_\pi $ and $D_9 \coloneqq 2 L_V C_p$.
\end{Lemma}
\begin{proof}
For the worker that contributes to the $k$th update, we construct its Markov chain:
\begin{align*}
    % s_{t_k-m} \xrightarrow{\theta_{k-d_m}} a_{t_k-m} \xrightarrow{\hat{\mathcal{P}}} s_{t_k-m+1} \xrightarrow{\theta_{k-d_{m-1}}} a_{t_k-m+1} \cdots s_{t_k} \xrightarrow{\theta_{k-d_0}} a_{t_k} \xrightarrow{\hat{\mathcal{P}}} s_{t_k+1},
    \hat{s}_{t-m} \xrightarrow{\theta_{k-d_m}} \hat{a}_{t-m} \xrightarrow{\hat{\mathcal{P}}} \hat{s}_{t-m+1} \xrightarrow{\theta_{k-d_{m-1}}} \hat{a}_{t-m+1} \cdots \hat{s}_{t-1} \xrightarrow{\theta_{k-d_1}} \hat{a}_{t-1}  \xrightarrow{\hat{\mathcal{P}}} s_t \xrightarrow{\theta_{k-d_0}} a_t \xrightarrow{\hat{\mathcal{P}}} \hat{s}_{t+1},
\end{align*}
where $(s_t,a_t,\hat{s}_{t+1})=(\hat{s}_{(k)},\hat{a}_{(k)},s'_{(k)})$, and $\{d_j\}_{j=0}^{m}$ is some increasing sequence with $d_0 \coloneqq \tau_k$. By (\ref{eq:M_K-derivation}) in Lemma \ref{lemma:tts-critic-noise}, we have $d_m \leq (K_0+1)m+K_0$.

Given $(\hat{s}_{t-m}, \hat{a}_{t-m},\hat{s}_{t-m+1})$ and $\theta_{k-d_m}$, we construct an auxiliary Markov chain:
\begin{align*}
    % s_{t_k-m} \xrightarrow{\theta_{k-d_m}} a_{t_k-m} \xrightarrow{\hat{\mathcal{P}}} s_{t_k-m+1} \xrightarrow{\theta_{k-d_m}} \widetilde{a}_{t_k-m+1} \cdots \widetilde{s}_{t_k} \xrightarrow{\theta_{k-d_m}} \widetilde{a}_{t_k} \xrightarrow{\hat{\mathcal{P}}} \widetilde{s}_{t_k+1}.
    \hat{s}_{t-m} \xrightarrow{\theta_{k-d_m}} \hat{a}_{t-m} \xrightarrow{\hat{\mathcal{P}}} \hat{s}_{t-m+1} \xrightarrow{\theta_{k-d_m}} \widetilde{a}_{t-m+1} \cdots \widetilde{s}_{t-1} \xrightarrow{\theta_{k-d_m}} \widetilde{a}_{t-1}\xrightarrow{\hat{\mathcal{P}}} \widetilde{s}_t \xrightarrow{\theta_{k-d_m}} \widetilde{a}_{t} \xrightarrow{\hat{\mathcal{P}}} \widetilde{s}_{t+1}.
\end{align*}
First we have
\begin{align*}\label{eq:lemma6-tmp0}
    &\left\langle \nabla J_\lambda(\theta_k), \delta(\hat{x}_{(k)},\theta_k)\psi_{\theta_{k-\tau_k}}(\hat{s}_{(k)},\hat{a}_{(k)}) - \nabla J(\theta_k) \right\rangle \\
    % &= \left\langle \nabla J_\lambda(\theta_k), \delta(\hat{x}_{(k)},\theta_k)\psi_{\theta_{k-\tau_k}}(\hat{s}_{(k)},\hat{a}_{(k)}) - \nabla J_\lambda(\theta_k) \right\rangle - \left\langle \nabla J_\lambda(\theta_k), \delta(\hat{x}_{(k)},\theta_k)\psi_{\theta_{k-d_m}}(\hat{s}_{(k)},\hat{a}_{(k)}) - \nabla J_\lambda(\theta_k) \right\rangle \\
    % &~~~~+ \left\langle \nabla J_\lambda(\theta_k), \delta(\hat{x}_{(k)},\theta_k)\psi_{\theta_{k-d_m}}(s,a) - \nabla J_\lambda(\theta_k) \right\rangle \\
    &= \left\langle \nabla J_\lambda(\theta_k), \delta(\hat{x}_{(k)},\theta_k)\left(\psi_{\theta_{k-\tau_k}}(\hat{s}_{(k)},\hat{a}_{(k)}) - \psi_{\theta_{k-d_m}}(\hat{s}_{(k)},\hat{a}_{(k)})\right) \right\rangle \\
    &~~~~ + \left\langle \nabla J_\lambda(\theta_k), \delta(\hat{x}_{(k)},\theta_k)\psi_{\theta_{k-d_m}} (\hat{s}_{(k)},\hat{a}_{(k)}) - \nabla J(\theta_k)\right\rangle. \numberthis \\
\end{align*}
We bound the first term in \eqref{eq:lemma6-tmp0} as
\begin{align*}\label{eq:lemma6-tmp1}
    &\left\langle \nabla J_\lambda(\theta_k), \delta(\hat{x}_{(k)},\theta_k)\left(\psi_{\theta_{k-\tau_k}}(\hat{s}_{(k)},\hat{a}_{(k)}) - \psi_{\theta_{k-d_m}}(\hat{s}_{(k)},\hat{a}_{(k)})\right) \right\rangle \\
    &\geq -\left\| \nabla J_\lambda(\theta_k) \right\|_2 \|\delta(\hat{x}_{(k)},\theta_k)\|_2 \|\psi_{\theta_{k-\tau_k}}(\hat{s}_{(k)},\hat{a}_{(k)}) - \psi_{\theta_{k-d_m}}(\hat{s}_{(k)},\hat{a}_{(k)})\|_2 \\
    &\geq -L_V C_\delta \|\psi_{\theta_{k-\tau_k}}(\hat{s}_{(k)},\hat{a}_{(k)}) - \psi_{\theta_{k-d_m}}(\hat{s}_{(k)},\hat{a}_{(k)})\|_2 \\
    &\geq -L_V C_\delta L_\psi \|\theta_{k-\tau_k} - \theta_{k-d_m}\|_2, \numberthis
\end{align*}
where the last inequality follows Assumption \ref{assumption:omega}.
Substituting \eqref{eq:lemma6-tmp1} into \eqref{eq:lemma6-tmp0} gives
\begin{align*}\label{eq:lemma6-tmp2}
    &\left\langle \nabla J_\lambda(\theta_k), \delta(\hat{x}_{(k)},\theta_k)\psi_{\theta_{k-\tau_k}}(\hat{s}_{(k)},\hat{a}_{(k)}) - \nabla J_\lambda(\theta_k) \right\rangle \\
    &\geq -L_V C_\delta L_\psi \|\theta_{k-\tau_k} - \theta_{k-d_m}\|_2 + \left\langle \nabla J_\lambda(\theta_k), \delta(\hat{x}_{(k)},\theta_k)\psi_{\theta_{k-d_m}} (\hat{s}_{(k)},\hat{a}_{(k)}) - \nabla J_\lambda(\theta_k) \right\rangle. \numberthis
\end{align*}
Then we start to bound the second term in \eqref{eq:lemma6-tmp2}. For brevity, we define
\begin{align*}
    \Delta_3(x,\theta) \coloneqq \left\langle \nabla J_\lambda(\theta), \delta(x,\theta)\psi_{\theta_{k-d_m}} (s,a) - \nabla J(\theta) \right\rangle.
\end{align*}
Throughout the following proof, we use $\theta$, $\theta'$, $x$ as shorthand notations of $\theta_k$, $\theta_{k-d_m}$, $x_t$ respectively. Define $\widetilde{x}\coloneqq (\widetilde{s}_{t-d_m},\widetilde{a}_{t-d_m},\widetilde{s}'_{t-d_m})$ where $\widetilde{s}_{t-d_m},\widetilde{a}_{t-d_m}$ are state and action on the auxiliary Markov chain, and $\widetilde{s}'_{t-d_m}\sim \mathcal{P}(\cdot|\widetilde{s}_{t-d_m},\widetilde{a}_{t-d_m})$ is a virtual sample.

We decompose $\Delta_3(x,\theta)$ as
\begin{align*}
    \Delta_3(x,\theta)
    &= \underbracket{\Delta_3(x,\theta)-\Delta_3(x,\theta')}_{I_1} + \underbracket{\Delta_3(x,\theta')-\Delta_3(\widetilde{x},\theta')}_{I_2} + \underbracket{\Delta_3(\widetilde{x},\theta')}_{I_3}.
\end{align*}
We first bound $I_1$ as
\begin{align*}\label{eq:I_1-lemma6}
    |I_1|&=|\Delta_3(x,\theta)-\Delta_3(x,\theta')| \\
    &= \left| \left\langle \nabla J_\lambda(\theta), \delta(x,\theta)\psi_{\theta'}(s,a) \right\rangle -\big\langle \nabla J_\lambda (\theta),\nabla J(\theta) \big\rangle - \left\langle \nabla J_\lambda(\theta'), \delta(x,\theta')\psi_{\theta'}(s,a) \right\rangle + \big\langle \nabla J_\lambda (\theta'),\nabla J(\theta') \big\rangle \right|\\
    &\leq \left| \left\langle \nabla J_\lambda(\theta), \delta(x,\theta)\psi_{\theta'}(s,a) \right\rangle - \left\langle \nabla J_\lambda(\theta'), \delta(x,\theta')\psi_{\theta'}(s,a) \right\rangle \right| +\left| \big\langle \nabla J_\lambda (\theta),\nabla J(\theta) \big\rangle -\big\langle \nabla J_\lambda (\theta'),\nabla J(\theta') \big\rangle \right| \\
    &\leq \left| \left\langle \nabla J_\lambda(\theta), \delta(x,\theta)\psi_{\theta'}(s,a) \right\rangle - \left\langle \nabla J_\lambda(\theta'), \delta(x,\theta')\psi_{\theta'}(s,a) \right\rangle \right| + 2L_V L_{\lambda} \|\theta-\theta'\|_2, \numberthis
\end{align*}
where the last inequality uses the fact that $\left\langle \nabla J_\lambda(\theta),\nabla J(\theta) \right\rangle$ is $2L_V L_\lambda$-lipschitz continuous.
We bound the first term in (\ref{eq:I_1-lemma6}) as
\begin{align*}
    &\left| \left\langle \nabla J_\lambda(\theta), \delta(x,\theta)\psi_{\theta'}(s,a) \right\rangle - \left\langle \nabla J_\lambda(\theta'), \delta(x,\theta')\psi_{\theta'}(s,a) \right\rangle \right| \\
    &\leq \left| \left\langle \nabla J_\lambda(\theta), \delta(x,\theta)\psi_{\theta'}(s,a) \right\rangle - \left\langle \nabla J_\lambda(\theta), \delta(x,\theta')\psi_{\theta'}(s,a) \right\rangle \right| \\
    &~~~~+ \left| \left\langle \nabla J_\lambda(\theta), \delta(x,\theta')\psi_{\theta'}(s,a) \right\rangle - \left\langle \nabla J_\lambda(\theta'), \delta(x,\theta')\psi_{\theta'}(s,a) \right\rangle \right| \\
    &= \left| \left\langle \nabla J_\lambda(\theta), \left(\delta(x,\theta)-\delta(x,\theta')\right)\psi_{\theta'}(s,a) \right\rangle\right|
    + \left| \left\langle \nabla J_\lambda(\theta) - \nabla J_\lambda(\theta'), \delta(x,\theta')\psi_{\theta'}(s,a) \right\rangle \right| \\
    &\leq L_V C_\psi\left| \delta(x,\theta)-\delta(x,\theta')\right|
    + C_p\|\nabla J_\lambda(\theta) - \nabla J_\lambda(\theta')\|_2 \\
    &= L_V C_\psi\left| \gamma(V_{\pi_\theta}(s')-V_{\pi_{\theta'}}(s')) + V_{\pi_{\theta'}}(s)-V_{\pi_{\theta}}(s)\right|
    + C_p\|\nabla J_\lambda(\theta) - \nabla J_\lambda(\theta')\|_2 \\
    &\leq L_V C_\psi\left(\gamma\left| V_{\pi_\theta}(s')-V_{\pi_{\theta'}}(s')\right| +\left| V_{\pi_{\theta'}}(s)-V_{\pi_{\theta}}(s)\right|\right)
    + C_p\|\nabla J_\lambda(\theta) - \nabla J_\lambda(\theta')\|_2 \\
    &\leq L_V C_\psi\left(\gamma L_V \|\theta-\theta'\|_2 + L_V \|\theta'-\theta\|\right)
    +C_p L_{\lambda} \|\theta-\theta'\|_2 \\
    &= \left(C_p L_{\lambda} +(1+\gamma)L_V^2 C_\psi \right) \|\theta-\theta'\|_2. \numberthis
\end{align*}
Substituting the above inequality into (\ref{eq:I_1-lemma6}) gives the lower bound of $I_1$:
\begin{align*}
    I_1 &\geq  -\left(C_p L_{\lambda} +(1+\gamma)L_V^2 C_\psi + 2L_V L_{\lambda} \right) \|\theta-\theta'\|_2.
\end{align*}
First we bound $I_2$ as
\begin{align*}\label{eq:I_2-1}
    \E[I_2 |\theta',\hat{s}_{t-m+1}]
    &= \E\left[\Delta_3(x,\theta')-\Delta_3(\widetilde{x},\theta') |\theta',\hat{s}_{t-m+1}\right] \\
    &\geq -\big|\E\left[\Delta_3(x,\theta') |\theta',\hat{s}_{t-m+1}\right]-\E\left[\Delta_3(\widetilde{x},\theta') |\theta',\hat{s}_{t-m+1}\right] \big| \\
    % &\geq -\left|\E\left\langle \nabla J_\lambda(\theta'),\delta(x,\theta')\psi_{\theta'}(s,a) \right\rangle - \E\left\langle \nabla J_\lambda(\theta'),\delta(\widetilde{x},\theta')\psi_{\theta'}(\widetilde{s},\widetilde{a}) \right\rangle \right|\\
    &\geq -\sup_{x}|\Delta_3(x,\theta')| \left\| \mathbb{P}(x \in \cdot | \theta',\hat{s}_{t-m+1})- \mathbb{P}(\widetilde{x} \in \cdot | \theta',\hat{s}_{t-m+1}) \right\|_{TV} \\
    &\geq -2 L_V(C_p + L_V) d_{TV}\left(\mathbb{P}(x \in \cdot | \theta',\hat{s}_{t-m+1}), \mathbb{P}(\widetilde{x} \in \cdot | \theta',\hat{s}_{t-m+1})\right) \\
    &\geq -L_V(C_p + L_V) |\mathcal{A}| L_\pi \sum_{i=\tau_k}^{d_m} \E\left[\|\theta_{k-i}-\theta_{k-d_m}\|_2 |\theta',\hat{s}_{t-m+1}\right], \numberthis
\end{align*}
where the second inequality is due to the definition of TV norm, the last inequality is due to (\ref{eq:auxiliary-markov-chain-4}) in Lemma \ref{lemma:auxiliary-chain}, and the second last inequality follows the fact that
\begin{align*}\label{eq:Delta_3bound}
    |\Delta_3(x,\theta')| \leq \|\nabla J_\lambda(\theta)\|_2 \left(\|\delta(x,\theta)\psi_{\theta_{k-d_m}}(s,a)\|_2 + \|\nabla J(\theta)\|_2 \right) \leq L_V(C_p + L_V). \numberthis
\end{align*}
Taking total expectation on both sides of \eqref{eq:I_2-1} yields
\begin{align*}
    \E[I_2] \geq -L_V(C_p + L_V) |\mathcal{A}| L_\pi \sum_{i=\tau_k}^{d_m} \E\|\theta_{k-i}-\theta_{k-d_m}\|_2. \numberthis
\end{align*}

Define $\overline{x} \coloneqq (\overline{s},\overline{a},\overline{s}')$, where $\overline{s} \sim d_{\theta'}$, $\overline{a} \sim \pi_{\theta'}$ and $\overline{s}' \sim \mathcal{P}$. Then we can further decompose $I_3$ as
\begin{align*}\label{eq:I_3-decomp}
    \E[I_3 |\theta',\hat{s}_{t-m+1}] = \E\left[\Delta_3(\widetilde{x},\theta') - \Delta_3(\overline{x}, \theta') |\theta',\hat{s}_{t-m+1}\right] + \E[\Delta_3(\overline{x},\theta') | \theta',\hat{s}_{t-m+1}].\numberthis
\end{align*}
The first term in \eqref{eq:I_3-decomp} can be bounded as
\begin{align*}\label{eq:EI_3-1-}
    & \E\left[\Delta_3(\widetilde{x},\theta') - \Delta_3(\overline{x}, \theta') |\theta',\hat{s}_{t-m+1}\right] \\
    % &= \E\left\langle \nabla J_\lambda(\theta'),  \delta(\widetilde{x},\theta')\psi_{\theta'}(\widetilde{s},\widetilde{a}) - \delta(\overline{x},\theta')\psi_{\theta'}(\overline{s},\overline{a}) \right\rangle  \\
    &\geq -\big| \E\left[\Delta_3(\widetilde{x},\theta')|\theta',\hat{s}_{t-m+1}\right] -\E\left[\Delta_3(\overline{x}, \theta')|\theta',\hat{s}_{t-m+1}\right]  \big| \\
    &\geq -\sup_{x}\left| \Delta_3(x,\theta')\right| \left\| \mathbb{P}(\widetilde{x} \in \cdot| \theta',\hat{s}_{t-m+1}) - \mathbb{P}(\overline{x} \in \cdot| \theta',\hat{s}_{t-m+1}) \right\|_{TV} \\
    &\geq -2 L_V (C_p +L_V)  d_{TV}\left(\mathbb{P}(\widetilde{x} \in \cdot| \theta',\hat{s}_{t-m+1}), \mathbb{P}(\overline{x} \in \cdot| \theta',\hat{s}_{t-m+1})  \right) \\
    &= -2L_V (C_p +L_V) d_{TV}\left(\mathbb{P}(\widetilde{x} \in \cdot| \theta',\hat{s}_{t-m+1}), d_{\theta'} \otimes \pi_{\theta'} \otimes \mathcal{P} \right) \numberthis
\end{align*}
where the second inequality follows the definition of $TV$-norm, and the third one follows \eqref{eq:Delta_3bound}.

The auxiliary Markov chain with policy $\pi_{\theta'}$ starts from initial state $\hat{s}_{t-m+1}$, and $\widetilde{s}_t$ is the $(m-1)$th state on the chain. Following Lemma \ref{lemma:geom-s-a-s'}, we have:
\begin{align*}
    d_{TV} \left( \mathbb{P}(\widetilde{x} \in \cdot | \theta',\hat{s}_{t-m+1}), d_{\theta'} \otimes\pi_{\theta'} \otimes \mathcal{P} \right)
    &= d_{TV} \left( \mathbb{P}\left((\widetilde{s}_t,\widetilde{a}_t,\widetilde{s}_{t+1}) \in \cdot | \theta',\hat{s}_{t-m+1}\right), d_{\theta'} \otimes\pi_{\theta'} \otimes \mathcal{P} \right) \\
    &\leq  \kappa \rho^{m-1}.
\end{align*}
Substituting the last inequality into \eqref{eq:EI_3-1-} yields
\begin{align*}\label{eq:I_3-1}
    \E\left[\Delta_3(\widetilde{x},\theta') - \Delta_3(\overline{x}, \theta') |\theta',\hat{s}_{t-m+1}\right] &\geq -2L_V (C_p +L_V) \kappa \rho^{m-1}. \numberthis
\end{align*}

The second term in \eqref{eq:I_3-decomp} is simply
\begin{align*}\label{eq:I_3-2}
    \E[\Delta_3(\overline{x},\theta') | \theta',\hat{s}_{t-m+1}] 
    &= \E\left[\left\langle \nabla J_\lambda(\theta'), \delta(\overline{x},\theta')\psi_{\theta'} (\overline{s},\overline{a}) - \nabla J(\theta') \right\rangle | \theta',\hat{s}_{t-m+1}\right]
    =0. \numberthis
\end{align*}

Substituting \eqref{eq:I_3-1} and \eqref{eq:I_3-2} into \eqref{eq:I_3-decomp} yields
\begin{align*}
     \E[I_3 |\theta',\hat{s}_{t-m+1}] \geq -2L_V (C_p +L_V) \kappa \rho^{m-1}. \numberthis
\end{align*}

Taking total expectation on $\Delta_3(x,\theta)$ and collecting lower bounds of $I_1$, $I_2$, $I_3$ yield
\begin{align*}
    \E[\Delta_3(x,\theta)]
    &\geq  -\left(C_p L_{\lambda} +(1+\gamma)L_V^2 C_\psi + 2L_V L_{\lambda} \right)\E\|\theta_k-\theta_{k-d_m}\|_2 \\
    &~~~~- L_V (C_p+L_V) |\mathcal{A}| L_\pi \sum_{i=\tau_k}^{d_m} \E\|\theta_{k-i}-\theta_{k-d_m}\|_2 - 2L_V (C_p+L_V) \kappa \rho^{m-1}. \numberthis
\end{align*}
Taking total expectation on (\ref{eq:lemma6-tmp2}) and substituting the above inequality into it yield
\begin{align*}
    &\E\left\langle \nabla J_\lambda(\theta_k), \delta(\hat{x}_{(k)},\theta_k)\psi_{\theta_{k-\tau_k}}(\hat{s}_{(k)},\hat{a}_{(k)}) - \nabla J(\theta_k) \right\rangle \\
    &\geq \!-\!D_6 \E\|\theta_{k-\tau_k} \!-\! \theta_{k-d_m}\|_2 \!-\! D_7 \E\|\theta_k\!-\!\theta_{k-d_m}\|_2
    \!-\! D_8 \sum_{i=\tau_k}^{d_m} \E\|\theta_{k-i}-\theta_{k-d_m}\|_2 \!-\! D_9 \kappa \rho^{m-1}, \numberthis
\end{align*}
where $D_6 \coloneqq L_V C_\delta L_\psi$, $D_7 \coloneqq C_p L_{\lambda} +(1+\gamma)L_V^2 C_\psi + 2L_V L_{\lambda}$, $D_8 \coloneqq L_V (C_p+L_V) |\mathcal{A}| L_\pi $, $D_9 \coloneqq 2 L_V (C_p+L_V)$.
\end{proof}

\end{document}